\pdfoutput=1

\documentclass[10pt,twocolumn,letterpaper]{article}

\usepackage{cvpr}              

\usepackage{multirow}
\usepackage{array}
\usepackage{xspace}
\usepackage{tikz}
\usepackage{fix-cm}
\usetikzlibrary{arrows.meta,calc}
\definecolor{stInk}{HTML}{253B4B}
\definecolor{stBlue}{HTML}{3770A8}
\definecolor{stTeal}{HTML}{00907F}
\definecolor{stOchre}{HTML}{DA973B}
\definecolor{stPurple}{HTML}{8A68A5}

\definecolor{pendingorange}{RGB}{230,120,0}
\definecolor{pendingfill}{RGB}{255,240,220}
\definecolor{branchblue}{RGB}{0,90,160}
\definecolor{branchfill}{RGB}{225,238,250}

\iftoggle{cvprfinal}{%
  \newcommand{\pending}[1]{\errmessage{paper: PENDING text present in final copy -- resolve every pending macro before the camera-ready build}}%
  \newcommand{\pendinginline}[1]{\errmessage{paper: PENDING text present in final copy -- resolve every pendinginline macro before the camera-ready build}}%
  \newcommand{\ifbranch}[3]{\errmessage{paper: unresolved outcome branch in final copy -- resolve every ifbranch macro before the camera-ready build}}%
}{%
  \newcommand{\pending}[1]{\par\noindent\fcolorbox{pendingorange}{pendingfill}{\parbox{\dimexpr\linewidth-2\fboxsep-2\fboxrule\relax}{\small\textcolor{pendingorange}{\textbf{PENDING}}\ #1}}\par}%
  \newcommand{\pendinginline}[1]{\textcolor{pendingorange}{\textbf{[PENDING: }#1\textbf{]}}}%
  \newcommand{\ifbranch}[3]{\par\noindent\fcolorbox{branchblue}{branchfill}{\parbox{\dimexpr\linewidth-2\fboxsep-2\fboxrule\relax}{\small\textcolor{branchblue}{\textbf{IF #1:}}\ #2\\[2pt]\textcolor{branchblue}{\textbf{ELSE:}}\ #3}}\par}%
}

\makeatletter
\newcommand{\stres}[1]{\@ifundefined{stresult@#1}{\errmessage{Missing SceneTether result macro}}{\csname stresult@#1\endcsname}}
\makeatother
\expandafter\def\csname stresult@echo_psnr\endcsname{5.15}
\expandafter\def\csname stresult@echo_psnr_ci\endcsname{[4.24, 6.13]}
\expandafter\def\csname stresult@echo_psnr_n\endcsname{42}
\expandafter\def\csname stresult@echo_de\endcsname{-9.24}
\expandafter\def\csname stresult@echo_de_ci\endcsname{[-12.38, -6.33]}
\expandafter\def\csname stresult@echo_de_n\endcsname{42}
\expandafter\def\csname stresult@echo_flow\endcsname{1.50}
\expandafter\def\csname stresult@echo_flow_ci\endcsname{[1.27, 1.76]}
\expandafter\def\csname stresult@echo_flow_n\endcsname{38}
\expandafter\def\csname stresult@spatial_psnr\endcsname{5.88}
\expandafter\def\csname stresult@spatial_psnr_ci\endcsname{[4.90, 6.87]}
\expandafter\def\csname stresult@spatial_psnr_n\endcsname{42}
\expandafter\def\csname stresult@spatial_de\endcsname{-10.20}
\expandafter\def\csname stresult@spatial_de_ci\endcsname{[-12.80, -7.73]}
\expandafter\def\csname stresult@spatial_de_n\endcsname{42}
\expandafter\def\csname stresult@spatial_flow\endcsname{2.30}
\expandafter\def\csname stresult@spatial_flow_ci\endcsname{[1.85, 2.83]}
\expandafter\def\csname stresult@spatial_flow_n\endcsname{30}
\expandafter\def\csname stresult@parent_psnr\endcsname{6.27}
\expandafter\def\csname stresult@parent_psnr_ci\endcsname{[5.31, 7.31]}
\expandafter\def\csname stresult@parent_psnr_n\endcsname{42}
\expandafter\def\csname stresult@parent_de\endcsname{-4.66}
\expandafter\def\csname stresult@parent_de_ci\endcsname{[-7.45, -2.31]}
\expandafter\def\csname stresult@parent_de_n\endcsname{42}
\expandafter\def\csname stresult@parent_flow\endcsname{0.90}
\expandafter\def\csname stresult@parent_flow_ci\endcsname{[0.80, 0.99]}
\expandafter\def\csname stresult@parent_flow_n\endcsname{36}
\expandafter\def\csname stresult@coverage_psnr\endcsname{8.14}
\expandafter\def\csname stresult@coverage_psnr_ci\endcsname{[7.05, 9.38]}
\expandafter\def\csname stresult@coverage_psnr_n\endcsname{42}
\expandafter\def\csname stresult@coverage_de\endcsname{-10.51}
\expandafter\def\csname stresult@coverage_de_ci\endcsname{[-16.92, -5.30]}
\expandafter\def\csname stresult@coverage_de_n\endcsname{42}
\expandafter\def\csname stresult@coverage_flow\endcsname{0.93}
\expandafter\def\csname stresult@coverage_flow_ci\endcsname{[0.85, 1.03]}
\expandafter\def\csname stresult@coverage_flow_n\endcsname{36}
\expandafter\def\csname stresult@half_psnr\endcsname{3.01}
\expandafter\def\csname stresult@half_psnr_ci\endcsname{[2.51, 3.54]}
\expandafter\def\csname stresult@half_psnr_n\endcsname{42}
\expandafter\def\csname stresult@half_de\endcsname{0.03}
\expandafter\def\csname stresult@half_de_ci\endcsname{[-1.47, 1.20]}
\expandafter\def\csname stresult@half_de_n\endcsname{42}
\expandafter\def\csname stresult@half_flow\endcsname{1.01}
\expandafter\def\csname stresult@half_flow_ci\endcsname{[0.97, 1.05]}
\expandafter\def\csname stresult@half_flow_n\endcsname{37}

\expandafter\def\csname stresult@horizon_appearance_n\endcsname{42}
\expandafter\def\csname stresult@horizon_activity_n\endcsname{40}

\expandafter\def\csname stresult@h12_psnr\endcsname{7.31}
\expandafter\def\csname stresult@h12_psnr_ci\endcsname{[5.42, 9.25]}
\expandafter\def\csname stresult@h12_psnr_n\endcsname{10}
\expandafter\def\csname stresult@h12_de\endcsname{-13.09}
\expandafter\def\csname stresult@h12_de_ci\endcsname{[-17.50, -8.81]}
\expandafter\def\csname stresult@h12_de_n\endcsname{10}
\expandafter\def\csname stresult@h12_mcff\endcsname{1.02}
\expandafter\def\csname stresult@h12_mcff_ci\endcsname{[0.21, 1.99]}
\expandafter\def\csname stresult@h12_mcff_n\endcsname{10}

\expandafter\def\csname stresult@t1_st_psnr\endcsname{22.37}
\expandafter\def\csname stresult@t1_st_color\endcsname{5.57}
\expandafter\def\csname stresult@t1_psnr_gain_min\endcsname{4.31}
\expandafter\def\csname stresult@t1_psnr_gain_max\endcsname{7.98}
\expandafter\def\csname stresult@t1_n_refs\endcsname{9}
\expandafter\def\csname stresult@cov_st_240\endcsname{93.2\%}
\expandafter\def\csname stresult@cov_sp_60\endcsname{74.4\%}
\expandafter\def\csname stresult@cov_sp_240\endcsname{38.5\%}

\expandafter\def\csname stresult@comp_flow\endcsname{1.05}
\expandafter\def\csname stresult@comp_flow_ci\endcsname{[0.92, 1.18]}
\expandafter\def\csname stresult@comp_flow_n\endcsname{38}
\expandafter\def\csname stresult@comp_psnr\endcsname{+0.20}
\expandafter\def\csname stresult@comp_psnr_ci\endcsname{[-1.15, 1.54]}
\expandafter\def\csname stresult@comp_psnr_n\endcsname{42}
\expandafter\def\csname stresult@comp_still_flicker\endcsname{0.23}
\expandafter\def\csname stresult@st_still_flicker\endcsname{3.79}

\expandafter\def\csname stresult@cfpp1_host_fps\endcsname{10.5}
\expandafter\def\csname stresult@cfpp1_host_fpswall\endcsname{7.3}
\expandafter\def\csname stresult@cfpp1_host_rtf\endcsname{2.2}
\expandafter\def\csname stresult@cfpp1_overhead_pct\endcsname{-1}
\expandafter\def\csname stresult@cfpp1_st_fps\endcsname{10.6}
\expandafter\def\csname stresult@cfpp1_st_fpswall\endcsname{7.4}
\expandafter\def\csname stresult@cfpp1_st_rtf\endcsname{2.2}
\expandafter\def\csname stresult@cfpp2_host_fps\endcsname{7.1}
\expandafter\def\csname stresult@cfpp2_host_fpswall\endcsname{5.5}
\expandafter\def\csname stresult@cfpp2_host_rtf\endcsname{2.9}
\expandafter\def\csname stresult@cfpp2_overhead_pct\endcsname{1}
\expandafter\def\csname stresult@cfpp2_st_fps\endcsname{7.1}
\expandafter\def\csname stresult@cfpp2_st_fpswall\endcsname{5.5}
\expandafter\def\csname stresult@cfpp2_st_rtf\endcsname{2.9}
\expandafter\def\csname stresult@svi_fps\endcsname{0.087}
\expandafter\def\csname stresult@svi_rtf\endcsname{172}

\expandafter\def\csname stresult@holm_p1_artefacts\endcsname{0.022}
\expandafter\def\csname stresult@holm_p1_motion\endcsname{0.543}
\expandafter\def\csname stresult@holm_p1_scene\endcsname{0.022}
\expandafter\def\csname stresult@holm_p2_artefacts\endcsname{0.001}
\expandafter\def\csname stresult@holm_p2_motion\endcsname{0.001}
\expandafter\def\csname stresult@holm_p2_scene\endcsname{0.001}
\expandafter\def\csname stresult@hs_automated\endcsname{1}
\expandafter\def\csname stresult@hs_c1_raters\endcsname{6}
\expandafter\def\csname stresult@hs_c1_scene_all\endcsname{88.5}
\expandafter\def\csname stresult@hs_c1_scene_qc\endcsname{86.5}
\expandafter\def\csname stresult@hs_coprimary_shift\endcsname{3.4}
\expandafter\def\csname stresult@hs_playback\endcsname{23}
\expandafter\def\csname stresult@hs_snapshot_date\endcsname{November 6, 2026}

\expandafter\def\csname stresult@twostep_psnr\endcsname{5.39}
\expandafter\def\csname stresult@twostep_color\endcsname{9.98}
\expandafter\def\csname stresult@twostep_flow\endcsname{1.44}
\expandafter\def\csname stresult@twostep_n\endcsname{42}
\expandafter\def\csname stresult@onestep_psnr\endcsname{2.86}
\expandafter\def\csname stresult@onestep_color\endcsname{4.75}
\expandafter\def\csname stresult@onestep_flow\endcsname{1.66}
\expandafter\def\csname stresult@onestep_n\endcsname{42}

\expandafter\def\csname stresult@overlap_support_mean\endcsname{84}
\expandafter\def\csname stresult@overlap_support_median\endcsname{93}
\expandafter\def\csname stresult@overlap_dynamic_mean\endcsname{18}

\expandafter\def\csname stresult@abs_color_240\endcsname{40}
\expandafter\def\csname stresult@abs_color_60\endcsname{40}
\expandafter\def\csname stresult@abs_psnr_240\endcsname{29}
\expandafter\def\csname stresult@abs_psnr_60\endcsname{29}
\expandafter\def\csname stresult@abs_psnr_diff_60\endcsname{5.1}
\expandafter\def\csname stresult@abs_psnr_ref_60\endcsname{Static sink, $k=3$}
\expandafter\def\csname stresult@abs_psnr_diff_240\endcsname{5.0}
\expandafter\def\csname stresult@abs_psnr_ref_240\endcsname{Framewise Echo, $k=1$}

\expandafter\def\csname stresult@drift60_st\endcsname{4.95}
\expandafter\def\csname stresult@drift60_comp\endcsname{9.03}
\expandafter\def\csname stresult@drift60_worse\endcsname{37}
\expandafter\def\csname stresult@drift60_n\endcsname{48}
\expandafter\def\csname stresult@var60_st\endcsname{13.77}
\expandafter\def\csname stresult@var60_comp\endcsname{13.86}
\expandafter\def\csname stresult@drift240_st\endcsname{6.27}
\expandafter\def\csname stresult@drift240_comp\endcsname{11.39}
\expandafter\def\csname stresult@drift240_worse\endcsname{10}
\expandafter\def\csname stresult@drift240_n\endcsname{12}
\expandafter\def\csname stresult@var240_st\endcsname{13.76}
\expandafter\def\csname stresult@var240_comp\endcsname{12.02}

\expandafter\def\csname stresult@cov_ad_240\endcsname{92.7\%}
\expandafter\def\csname stresult@cov_host_240\endcsname{82.5\%}
\expandafter\def\csname stresult@hs_m_motion\endcsname{75.0}
\expandafter\def\csname stresult@hs_m_raters\endcsname{6}
\expandafter\def\csname stresult@hs_m_scene\endcsname{97.9}
\expandafter\def\csname stresult@hs_m_votes\endcsname{36}
\expandafter\def\csname stresult@sc_band_hi\endcsname{$-$0.17}
\expandafter\def\csname stresult@sc_band_lo\endcsname{$-$0.50}
\expandafter\def\csname stresult@sc_band_st\endcsname{5.95}
\expandafter\def\csname stresult@sc_band_uni\endcsname{5.62}
\expandafter\def\csname stresult@sc_drift_hi\endcsname{+0.10}
\expandafter\def\csname stresult@sc_drift_lo\endcsname{$-$1.17}
\expandafter\def\csname stresult@sc_drift_st\endcsname{4.95}
\expandafter\def\csname stresult@sc_drift_uni\endcsname{4.43}
\expandafter\def\csname stresult@sc_edge_hi\endcsname{+0.02}
\expandafter\def\csname stresult@sc_edge_lo\endcsname{+0.00}
\expandafter\def\csname stresult@sc_edge_st\endcsname{0.81}
\expandafter\def\csname stresult@sc_edge_uni\endcsname{0.83}
\expandafter\def\csname stresult@sc_n\endcsname{48}
\expandafter\def\csname stresult@sh_anch240_ad\endcsname{0.84}
\expandafter\def\csname stresult@sh_anch240_host\endcsname{1.29}
\expandafter\def\csname stresult@sh_anch240_st\endcsname{0.81}
\expandafter\def\csname stresult@sh_anch60_ad\endcsname{0.84}
\expandafter\def\csname stresult@sh_anch60_comp\endcsname{0.98}
\expandafter\def\csname stresult@sh_anch60_host\endcsname{0.92}
\expandafter\def\csname stresult@sh_anch60_n\endcsname{47}
\expandafter\def\csname stresult@sh_anch60_st\endcsname{0.81}
\expandafter\def\csname stresult@sh_anch60_un\endcsname{0.90}

\expandafter\def\csname stresult@nl_n\endcsname{8}
\expandafter\def\csname stresult@nl_nmot\endcsname{8}
\expandafter\def\csname stresult@nl_st240\endcsname{21.87}
\expandafter\def\csname stresult@nl_st60\endcsname{21.96}
\expandafter\def\csname stresult@nl_stcol240\endcsname{5.65}
\expandafter\def\csname stresult@nl_stmot240\endcsname{1.07}
\expandafter\def\csname stresult@nl_svi240\endcsname{11.55}
\expandafter\def\csname stresult@nl_svi60\endcsname{15.53}
\expandafter\def\csname stresult@nl_svicol240\endcsname{35.92}
\expandafter\def\csname stresult@nl_svimot240\endcsname{0.86}
\expandafter\def\csname stresult@nl_wins240\endcsname{8}
\expandafter\def\csname stresult@nl_wins60\endcsname{7}

\expandafter\def\csname stresult@ctl_noref_act\endcsname{2.08}
\expandafter\def\csname stresult@ctl_noref_dpsnr\endcsname{$-$0.07}
\expandafter\def\csname stresult@ctl_noref_hi\endcsname{+0.04}
\expandafter\def\csname stresult@ctl_noref_lo\endcsname{$-$0.19}
\expandafter\def\csname stresult@ctl_noref_ratio\endcsname{0.98}
\expandafter\def\csname stresult@ctl_noref_rhi\endcsname{1.05}
\expandafter\def\csname stresult@ctl_noref_rlo\endcsname{0.91}
\expandafter\def\csname stresult@ctl_perm_act\endcsname{0.99}
\expandafter\def\csname stresult@ctl_perm_dpsnr\endcsname{$-$0.51}
\expandafter\def\csname stresult@ctl_perm_hi\endcsname{+0.27}
\expandafter\def\csname stresult@ctl_perm_lo\endcsname{$-$1.35}
\expandafter\def\csname stresult@ctl_perm_ratio\endcsname{0.47}
\expandafter\def\csname stresult@ctl_perm_rhi\endcsname{0.62}
\expandafter\def\csname stresult@ctl_perm_rlo\endcsname{0.34}
\expandafter\def\csname stresult@ctl_sperm_act\endcsname{2.03}
\expandafter\def\csname stresult@ctl_sperm_dpsnr\endcsname{+0.54}
\expandafter\def\csname stresult@ctl_sperm_hi\endcsname{+0.73}
\expandafter\def\csname stresult@ctl_sperm_lo\endcsname{+0.38}
\expandafter\def\csname stresult@ctl_sperm_ratio\endcsname{0.95}
\expandafter\def\csname stresult@ctl_sperm_rhi\endcsname{1.03}
\expandafter\def\csname stresult@ctl_sperm_rlo\endcsname{0.88}
\expandafter\def\csname stresult@ctl_st_act\endcsname{2.13}
\expandafter\def\csname stresult@ctl_suni_act\endcsname{2.11}
\expandafter\def\csname stresult@ctl_suni_dpsnr\endcsname{+0.52}
\expandafter\def\csname stresult@ctl_suni_hi\endcsname{+0.78}
\expandafter\def\csname stresult@ctl_suni_lo\endcsname{+0.29}
\expandafter\def\csname stresult@ctl_suni_ratio\endcsname{0.99}
\expandafter\def\csname stresult@ctl_suni_rhi\endcsname{1.07}
\expandafter\def\csname stresult@ctl_suni_rlo\endcsname{0.92}
\expandafter\def\csname stresult@ctl_uni_act\endcsname{1.39}
\expandafter\def\csname stresult@ctl_uni_dpsnr\endcsname{$-$0.58}
\expandafter\def\csname stresult@ctl_uni_hi\endcsname{+0.28}
\expandafter\def\csname stresult@ctl_uni_lo\endcsname{$-$1.50}
\expandafter\def\csname stresult@ctl_uni_ratio\endcsname{0.65}
\expandafter\def\csname stresult@ctl_uni_rhi\endcsname{0.82}
\expandafter\def\csname stresult@ctl_uni_rlo\endcsname{0.52}

\expandafter\def\csname stresult@cf_act_host\endcsname{1.02}
\expandafter\def\csname stresult@cf_act_st\endcsname{1.97}
\expandafter\def\csname stresult@cf_dbg\endcsname{$-$0.001}
\expandafter\def\csname stresult@cf_dbg_hi\endcsname{+0.001}
\expandafter\def\csname stresult@cf_dbg_lo\endcsname{$-$0.004}
\expandafter\def\csname stresult@cf_dcol\endcsname{$-$10.0}
\expandafter\def\csname stresult@cf_dcol_hi\endcsname{$-$6.9}
\expandafter\def\csname stresult@cf_dcol_lo\endcsname{$-$13.3}
\expandafter\def\csname stresult@cf_dpsnr\endcsname{+5.39}
\expandafter\def\csname stresult@cf_dpsnr_hi\endcsname{+6.48}
\expandafter\def\csname stresult@cf_dpsnr_lo\endcsname{+4.38}
\expandafter\def\csname stresult@cf_dsubj\endcsname{$-$0.002}
\expandafter\def\csname stresult@cf_dsubj_hi\endcsname{+0.002}
\expandafter\def\csname stresult@cf_dsubj_lo\endcsname{$-$0.005}
\expandafter\def\csname stresult@cf_psnr_host\endcsname{16.97}
\expandafter\def\csname stresult@cf_psnr_st\endcsname{22.37}
\expandafter\def\csname stresult@cf_ratio\endcsname{1.93}
\expandafter\def\csname stresult@cf_ratio_hi\endcsname{2.76}
\expandafter\def\csname stresult@cf_ratio_lo\endcsname{1.35}
\expandafter\def\csname stresult@cf_wins\endcsname{41}
\expandafter\def\csname stresult@sf_act_host\endcsname{1.33}
\expandafter\def\csname stresult@sf_act_st\endcsname{0.43}
\expandafter\def\csname stresult@sf_dbg\endcsname{+0.011}
\expandafter\def\csname stresult@sf_dbg_hi\endcsname{+0.016}
\expandafter\def\csname stresult@sf_dbg_lo\endcsname{+0.006}
\expandafter\def\csname stresult@sf_dcol\endcsname{$-$24.8}
\expandafter\def\csname stresult@sf_dcol_hi\endcsname{$-$17.9}
\expandafter\def\csname stresult@sf_dcol_lo\endcsname{$-$31.9}
\expandafter\def\csname stresult@sf_dpsnr\endcsname{+4.69}
\expandafter\def\csname stresult@sf_dpsnr_hi\endcsname{+5.66}
\expandafter\def\csname stresult@sf_dpsnr_lo\endcsname{+3.65}
\expandafter\def\csname stresult@sf_dsubj\endcsname{+0.016}
\expandafter\def\csname stresult@sf_dsubj_hi\endcsname{+0.023}
\expandafter\def\csname stresult@sf_dsubj_lo\endcsname{+0.009}
\expandafter\def\csname stresult@sf_n\endcsname{42}
\expandafter\def\csname stresult@sf_nmot\endcsname{40}
\expandafter\def\csname stresult@sf_port_dpsnr_max\endcsname{4.9}
\expandafter\def\csname stresult@sf_port_dpsnr_min\endcsname{4.5}
\expandafter\def\csname stresult@sf_port_psnr_max\endcsname{11.9}
\expandafter\def\csname stresult@sf_port_psnr_min\endcsname{11.5}
\expandafter\def\csname stresult@sf_port_ratio_max\endcsname{0.31}
\expandafter\def\csname stresult@sf_port_ratio_min\endcsname{0.24}
\expandafter\def\csname stresult@sf_psnr_host\endcsname{11.68}
\expandafter\def\csname stresult@sf_psnr_st\endcsname{16.37}
\expandafter\def\csname stresult@sf_ratio\endcsname{0.32}
\expandafter\def\csname stresult@sf_ratio_hi\endcsname{0.62}
\expandafter\def\csname stresult@sf_ratio_lo\endcsname{0.15}
\expandafter\def\csname stresult@sf_wins\endcsname{38}

\expandafter\def\csname stresult@pp_algpin_dpsnr\endcsname{$-$1.65}
\expandafter\def\csname stresult@pp_algpin_dpsnr_hi\endcsname{$-$1.25}
\expandafter\def\csname stresult@pp_algpin_dpsnr_lo\endcsname{$-$2.06}
\expandafter\def\csname stresult@pp_algpin_ratio\endcsname{1.76}
\expandafter\def\csname stresult@pp_algpin_ratio_hi\endcsname{2.47}
\expandafter\def\csname stresult@pp_algpin_ratio_lo\endcsname{1.22}
\expandafter\def\csname stresult@pp_algpin_wins\endcsname{4}
\expandafter\def\csname stresult@pp_dympin_dpsnr\endcsname{$-$1.90}
\expandafter\def\csname stresult@pp_dympin_dpsnr_hi\endcsname{$-$1.22}
\expandafter\def\csname stresult@pp_dympin_dpsnr_lo\endcsname{$-$2.59}
\expandafter\def\csname stresult@pp_dympin_ratio\endcsname{1.04}
\expandafter\def\csname stresult@pp_dympin_ratio_hi\endcsname{1.51}
\expandafter\def\csname stresult@pp_dympin_ratio_lo\endcsname{0.69}
\expandafter\def\csname stresult@pp_dympin_wins\endcsname{7}
\expandafter\def\csname stresult@pp_n\endcsname{42}
\expandafter\def\csname stresult@pp_nmot\endcsname{40}
\expandafter\def\csname stresult@pp_st_algpin_dpsnr\endcsname{+8.14}
\expandafter\def\csname stresult@pp_st_algpin_dpsnr_hi\endcsname{+9.41}
\expandafter\def\csname stresult@pp_st_algpin_dpsnr_lo\endcsname{+6.97}
\expandafter\def\csname stresult@pp_st_algpin_ratio\endcsname{1.05}
\expandafter\def\csname stresult@pp_st_algpin_ratio_hi\endcsname{1.76}
\expandafter\def\csname stresult@pp_st_algpin_ratio_lo\endcsname{0.69}
\expandafter\def\csname stresult@pp_st_dympin_dpsnr\endcsname{+7.38}
\expandafter\def\csname stresult@pp_st_dympin_dpsnr_hi\endcsname{+8.62}
\expandafter\def\csname stresult@pp_st_dympin_dpsnr_lo\endcsname{+6.22}
\expandafter\def\csname stresult@pp_st_dympin_ratio\endcsname{1.94}
\expandafter\def\csname stresult@pp_st_dympin_ratio_hi\endcsname{3.18}
\expandafter\def\csname stresult@pp_st_dympin_ratio_lo\endcsname{1.24}

\expandafter\def\csname stresult@sr_arms\endcsname{7}
\expandafter\def\csname stresult@sr_eta010_dcol\endcsname{+0.53}
\expandafter\def\csname stresult@sr_eta010_dpsnr\endcsname{+0.25}
\expandafter\def\csname stresult@sr_eta010_dpsnr_hi\endcsname{+0.53}
\expandafter\def\csname stresult@sr_eta010_dpsnr_lo\endcsname{$-$0.00}
\expandafter\def\csname stresult@sr_eta010_ratio\endcsname{0.99}
\expandafter\def\csname stresult@sr_eta010_rhi\endcsname{1.11}
\expandafter\def\csname stresult@sr_eta010_rlo\endcsname{0.88}
\expandafter\def\csname stresult@sr_eta050_dcol\endcsname{$-$0.50}
\expandafter\def\csname stresult@sr_eta050_dpsnr\endcsname{$-$0.71}
\expandafter\def\csname stresult@sr_eta050_dpsnr_hi\endcsname{$-$0.36}
\expandafter\def\csname stresult@sr_eta050_dpsnr_lo\endcsname{$-$1.08}
\expandafter\def\csname stresult@sr_eta050_ratio\endcsname{0.95}
\expandafter\def\csname stresult@sr_eta050_rhi\endcsname{1.06}
\expandafter\def\csname stresult@sr_eta050_rlo\endcsname{0.86}
\expandafter\def\csname stresult@sr_eta100_dcol\endcsname{$-$0.71}
\expandafter\def\csname stresult@sr_eta100_dpsnr\endcsname{$-$2.73}
\expandafter\def\csname stresult@sr_eta100_dpsnr_hi\endcsname{$-$2.04}
\expandafter\def\csname stresult@sr_eta100_dpsnr_lo\endcsname{$-$3.44}
\expandafter\def\csname stresult@sr_eta100_ratio\endcsname{1.00}
\expandafter\def\csname stresult@sr_eta100_rhi\endcsname{1.15}
\expandafter\def\csname stresult@sr_eta100_rlo\endcsname{0.89}
\expandafter\def\csname stresult@sr_maprefresh_dcol\endcsname{+0.70}
\expandafter\def\csname stresult@sr_maprefresh_dpsnr\endcsname{+0.28}
\expandafter\def\csname stresult@sr_maprefresh_dpsnr_hi\endcsname{+0.97}
\expandafter\def\csname stresult@sr_maprefresh_dpsnr_lo\endcsname{$-$0.40}
\expandafter\def\csname stresult@sr_maprefresh_ratio\endcsname{1.13}
\expandafter\def\csname stresult@sr_maprefresh_rhi\endcsname{1.38}
\expandafter\def\csname stresult@sr_maprefresh_rlo\endcsname{0.94}
\expandafter\def\csname stresult@sr_ramp_dcol\endcsname{+0.01}
\expandafter\def\csname stresult@sr_ramp_dpsnr\endcsname{+0.13}
\expandafter\def\csname stresult@sr_ramp_dpsnr_hi\endcsname{+0.42}
\expandafter\def\csname stresult@sr_ramp_dpsnr_lo\endcsname{$-$0.11}
\expandafter\def\csname stresult@sr_ramp_ratio\endcsname{0.88}
\expandafter\def\csname stresult@sr_ramp_rhi\endcsname{0.98}
\expandafter\def\csname stresult@sr_ramp_rlo\endcsname{0.81}
\expandafter\def\csname stresult@sr_rho010_dcol\endcsname{+0.21}
\expandafter\def\csname stresult@sr_rho010_dpsnr\endcsname{$-$0.23}
\expandafter\def\csname stresult@sr_rho010_dpsnr_hi\endcsname{$-$0.08}
\expandafter\def\csname stresult@sr_rho010_dpsnr_lo\endcsname{$-$0.39}
\expandafter\def\csname stresult@sr_rho010_ratio\endcsname{1.02}
\expandafter\def\csname stresult@sr_rho010_rhi\endcsname{1.14}
\expandafter\def\csname stresult@sr_rho010_rlo\endcsname{0.92}
\expandafter\def\csname stresult@sr_rho050_dcol\endcsname{$-$0.03}
\expandafter\def\csname stresult@sr_rho050_dpsnr\endcsname{+0.15}
\expandafter\def\csname stresult@sr_rho050_dpsnr_hi\endcsname{+0.37}
\expandafter\def\csname stresult@sr_rho050_dpsnr_lo\endcsname{$-$0.06}
\expandafter\def\csname stresult@sr_rho050_ratio\endcsname{0.96}
\expandafter\def\csname stresult@sr_rho050_rhi\endcsname{1.11}
\expandafter\def\csname stresult@sr_rho050_rlo\endcsname{0.82}

\newcommand{\Ours}{SceneTether\xspace}
\newcommand{\OursLabel}{\textbf{SceneTether (ours)}\xspace}

\newcommand{\CFpp}{Causal Forcing++\xspace}
\definecolor{cvprblue}{rgb}{0.21,0.49,0.74}
\usepackage[pagebackref,breaklinks,colorlinks,allcolors=cvprblue]{hyperref}

\title{SceneTether: Activity-Gated Value Anchoring for Long-Horizon Autoregressive Image Animation}

\author{
Matiur Rahman Minar\\
Sogang University\\
Seoul, Korea\\
{\tt\small minar@sogang.ac.kr}
\and
Seunghun Oh\\
Sogang University\\
Seoul, Korea\\
{\tt\small gnsgus190@sogang.ac.kr}
\and
GangHyeon Jeong\\
Sogang University\\
Seoul, Korea\\
{\tt\small ganghyeon@sogang.ac.kr} 
\and
Unsang Park\\
Sogang University\\
Seoul, Korea\\
{\tt\small unsangpark@sogang.ac.kr}
}

\begin{document}
\twocolumn[{%
\renewcommand\twocolumn[1][]{#1}%
\maketitle
\vspace{-1.5em}
\begin{center}
 \captionsetup{type=figure}
 \raisebox{-90.85350pt}{%
  \rlap{\raisebox{157.626bp}[0pt][0pt]{\hspace*{116.093bp}\hyperlink{cite.wu2026echoforcing}{\rule{0pt}{6.846bp}\hspace*{8.966bp}}}}%
  \rlap{\raisebox{113.711bp}[0pt][0pt]{\hspace*{99.934bp}\hyperlink{cite.jeon2026dymos}{\rule{0pt}{6.801bp}\hspace*{8.967bp}}}}%
  \rlap{\raisebox{73.690bp}[0pt][0pt]{\hspace*{92.180bp}\hyperlink{cite.choi2025alg}{\rule{0pt}{6.846bp}\hspace*{5.479bp}}}}%
  \includegraphics{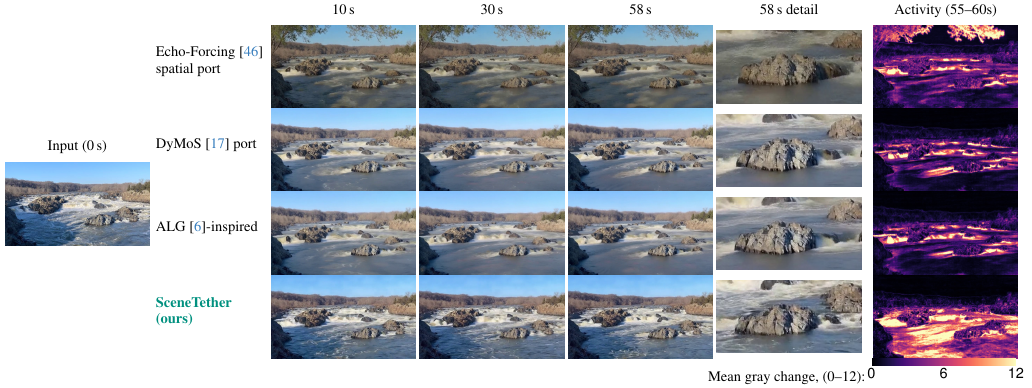}}
 \vspace{-5pt}
 \captionof{figure}{\textbf{Let the water change, not the scene.}
 One-minute animations from the same source image using the frozen Causal Forcing++~\cite{zhao2026causal_pp} backbone.
 Compared with the spatial Echo-Forcing port~\cite{wu2026echoforcing} and causal adaptations of DyMoS~\cite{jeon2026dymos} and ALG~\cite{choi2025alg}, \Ours retains the source rock and shoreline structures more closely at 58\,s while maintaining visible water activity.
 The rightmost column shows mean grayscale change over 55--60\,s on a shared scale, indicating temporal activity rather than motion naturalness.}
 \label{fig:fm_teaser}
\end{center}
\vspace{.2em}
}]

\begin{abstract}
Long-horizon image animation must keep the scene fixed while natural material e.g., water, smoke, or fire continues to move.
Autoregressive video generators read recent frames for temporal continuity, but this history also carries accumulated appearance errors, and sparsifying it disrupts motion.
We introduce \textbf{SceneTether}, a training-free attention-read policy that keeps where attention reads recent memory but changes what appearance that memory returns.
SceneTether keeps every key of the newest latent, blends its low-activity values toward co-located source-image values, and leaves high-activity values untouched so motion can evolve; an activity map estimated from the first generated latents and then frozen decides where to blend.
It adds no denoising evaluations and leaves the cache-update rule unchanged.
On a Causal Forcing++ generator, SceneTether raises source-support PSNR by 5.15\,dB over its framewise Echo host across two seeds with $1.5\times$ its late activity, and by 7.31\,dB over spatial Echo-Forcing on a preselected four-minute subset.
At equal recent-memory coverage, value anchoring alone adds 8.14\,dB.
A post-hoc pixel composite with the same map matches static fidelity but retains nearly twice the moving-region color drift, and viewers in a small blinded study preferred SceneTether for its natural scene and motion preservation.
Project page: \url{https://minar09.github.io/scenetether/}
\end{abstract}

\section{Introduction}
\label{fm:intro}

A river should keep flowing without its banks changing shape.
Long-horizon image animation requires this separation: given a source photograph and a caption, the generator must preserve the scene while water, smoke, or fire continues to evolve (\cref{fig:fm_teaser}).

Autoregressive video diffusion enables multi-minute generation by synthesizing frames or latent blocks sequentially, reusing history through a key--value (KV) cache~\cite{yin2025causvid,huang2025selfforcing,zhao2026causal_pp}.
Existing long-video memory policies primarily control which history remains available through anchors, compressed tokens, or attention reweighting~\cite{wu2026echoforcing,li2026packcache,zhang2025framepack}.
Fixed-camera image animation, however, introduces a persistent source constraint.
The input photograph specifies the scene's geometry and appearance, not merely its initial state: buildings, riverbanks, and viewpoint should remain stable, whereas individual foam patterns and flame shapes should keep changing.
A suitable memory policy must therefore preserve source-defined structure without repeatedly copying the initial state of moving materials.

The newest generated latent (four frames) is where this tension arises.
It supplies the recent context needed for temporal continuity, but also carries accumulated appearance errors into subsequent blocks. 
In development experiments, thinning recent spatial coverage reduces background drift but causes brief structural pop-ins, where scene elements abruptly reappear, whereas dense reads keep continuity but carry unwanted appearance changes.
This motivates separating \emph{where attention reads recent memory} from \emph{what appearance that memory returns}.

We introduce \Ours, a training-free, source-aligned attention-read policy that makes this separation explicit.
Rather than discarding recent tokens, \Ours keeps every token and key of the newest latent while interpolating its low-activity values toward co-located source values.
High-activity positions retain their recent values, allowing the generator to synthesize evolving material states.
A spatial activity map estimated once during warm-up governs this intervention throughout the rollout.
Operating entirely within attention, \Ours requires no model fine-tuning, output compositing, or additional denoising evaluations.

Our contributions are threefold:
\begin{itemize}[nosep,leftmargin=*]
  \item \textbf{Method.} A training-free read of recent memory that keeps complete newest-latent coverage and keys while anchoring low-activity values to the source image, with no extra denoising evaluations.
  \item \textbf{Mechanistic controls.} At equal newest-frame coverage, value anchoring alone adds 8.14\,dB source-support PSNR; spatial controls show that sparing high-activity positions, not the exact interior weighting, limits motion suppression; and a post-hoc pixel composite with the same map retains about $1.8\times$ the moving-region color drift.
  \item \textbf{Long-horizon evaluation.} A source-fixed protocol up to four minutes with region-resolved metrics, paired two-seed comparisons, registration-failure accounting, and blinded human judgments, against memory policies and image-to-video (I2V) guidance adaptations.
\end{itemize}

\section{Related work}
\label{fm:related}

\paragraph{Causal memory and feature editing.}
Autoregressive video diffusion generates frames or latent chunks sequentially~\cite{zhou2025magi,chen2024diffusion,ai2025magi1autoregressivevideogeneration}, with distillation and self-rollout training enabling few-step generation~\cite{yin2025causvid,huang2025selfforcing,zhu2026causal,zhao2026causal_pp}.
Long-horizon systems extend generation through rolling windows, attention sinks, error recycling, and noise or frequency scheduling~\cite{liu2025rolling,infinite-forcing,yang2025longlive,lu2025reward,helios,minar2026steadyforcing,li2025stable,qiu2024freenoise,lu2024freelong,cui2026selfforcingpp}.
Memory policies instead control which historical information attention retains.
Echo-Forcing maintains rotating anchors, compressed history, and recent frames~\cite{wu2026echoforcing}, and serves as the storage and update host for \Ours.
Deep Forcing combines deep sinks with compression~\cite{yi2026deepforcing}; Infinity-RoPE introduces block-relative coordinates and cache flushing~\cite{yesiltepe2025infinity}; and PackCache preserves conditioning-image tokens while budgeting older history by age~\cite{li2026packcache}.
Other methods condition on initial features~\cite{henschel2025streamingt2v}, compress history progressively or through learned representations~\cite{zhang2025framepack,wu2025packforce}, recompute selected historical states~\cite{ji2026forcingkv}, or adapt memory selection~\cite{luo2026futureforcing,mansour2026unstep}.

Closest to our operation, TetherCache aligns recalled historical KV statistics with trusted memory~\cite{meng2026tethercache}, while Custom Forcing modulates subject-reference values according to identity drift~\cite{ok2026customforcing}.
Attention-based image and video editing methods similarly inject source features or reuse inversion attention maps~\cite{cao2023masactrl,qi2023fatezero}.
Unlike these methods, which correct recalled history or reference tokens, \Ours modifies the \emph{newest generated latent} at read time, interpolating its low-activity values toward co-located source values while keeping all of its keys and spatial coverage.

\paragraph{Source-conditioned animation.}
Image-animation methods model motion through Eulerian fields~\cite{holynski2021animating,mahapatra2023text2cinemagraph}, controllable fluid elements~\cite{mahapatra2022controllable}, or learned spectral motion priors~\cite{Li_2024_gen_image_dynamics}.
Video diffusion models instead synthesize image-conditioned dynamics directly~\cite{xing2024dynamicrafter,blattmann2023svd,yang2025cogvideox,wan2025,kong2024hunyuanvideo,hacohen2024ltxvideo}, including approaches based on motion residuals~\cite{ma2025cinemo}.
\Ours operates within a causal diffusion generator, allowing new material states to be synthesized while anchoring low-activity appearance inside attention rather than compositing source pixels.

\paragraph{Motion-aware attention and I2V guidance.}
Ada-VE guides cross-frame attention using motion priors~\cite{mahmud2024adave}, while FlowV2V transfers motion through flow-conditioned generation~\cite{wang2025flowv2v}.
MotionCache reuses computation in static regions~\cite{xu2026motioncache}, and DiffTrack extracts temporal correspondences from diffusion-transformer attention~\cite{nam2025difftrack}.
Separately, image-to-video (I2V) models can suppress motion when reference conditioning dominates, a phenomenon studied as conditional image leakage~\cite{zhao2024cil}.
ALG mitigates this effect by filtering early image conditioning~\cite{choi2025alg}, and DyMoS attenuates excessive attention to reference-frame keys~\cite{jeon2026dymos}.
Whereas these methods weaken source conditioning to restore motion, \Ours strengthens it only where activity is low, by modifying the values of generated recent memory without additional denoising evaluations.

\paragraph{Evaluating fidelity and dynamics.}
VBench and VBench++ assess video consistency, motion, and visual quality~\cite{huang2023vbench,huang2024vbench++}, while other benchmarks study motion dynamics, human alignment, and intrinsic faithfulness~\cite{liao2024devil,liu2025dynamici2v,eval_crafter,he2024videoscore,Zheng2025VBench20AV}.
For fixed-camera scenes, WorldScore distinguishes static and moving regions~\cite{duan2025worldscore}, and SNF-Bench extends region-resolved evaluation to long natural-flow videos~\cite{snfbench2026}.
Our source-fixed protocol (\cref{fm:protocol}) uses shared evaluation masks, matched timestamps, and explicit registration-failure accounting.
We report source fidelity and regional activity separately, since motion magnitude alone cannot establish naturalness.

\section{\texorpdfstring{\Ours}{SceneTether}}
\label{fm:method}

\begin{figure*}[t]
 \centering
 \includegraphics{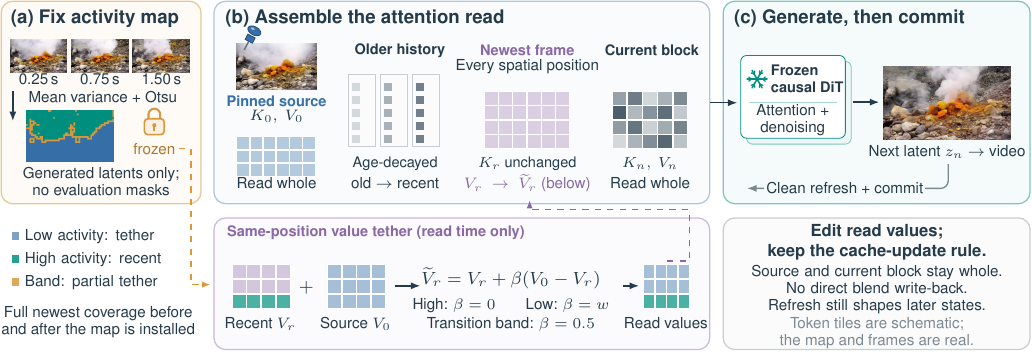}
 \par\vspace{12pt}
 \caption{\textbf{\Ours: retaining recent context, anchoring scene appearance.}
 \textbf{(a)} A spatial activity map is estimated from the first 1.5\,s of generated latents and frozen.
 \textbf{(b)} Attention retains all newest-frame keys, preserves high-activity values, and blends low-activity values toward co-located source values; the pinned source and age-decayed history remain accessible.
 \textbf{(c)} Only temporary read values change; the refresh/write rule is unchanged, but the states it commits depend on tethered reads.}
 \label{fig:fm_method}
\end{figure*}

Given a source image $I$ and caption $y$, a causal video diffusion model generates latent blocks conditioned on its committed key--value (KV) history.
\Ours updates the temporary KV set assembled for each self-attention read, leaving model weights, stored KV entries, cache eviction, denoising schedules, and text conditioning unchanged (\cref{fig:fm_method}).
Unlike output compositing, this intervention acts within the generator before subsequent states are generated.

\subsection{Separating coverage from appearance}
\label{fm:diagnosis}

Recent memory supplies spatial and temporal context for continuous motion, but can also propagate accumulated appearance errors.
Sparsifying recent tokens may remove context needed for continuity, whereas reading them densely can reinforce drifted background appearance.
We hypothesize that newest-frame keys remain useful for attention addressing, that is, deciding where queries read, even when their low-activity values have drifted.
Rather than discarding these tokens, \Ours retains their complete coverage and keys while selectively anchoring the values they return.
This does not assume that keys represent motion exclusively or that values encode only appearance; their roles remain coupled through attention.

\subsection{Backbone model and memory host}
\label{fm:method_backbone}

We use the frozen framewise two-step \CFpp student~\cite{zhao2026causal_pp} from the Wan2.1 family~\cite{wan2025,zhu2026causal}.
Its causal VAE encodes $I$ independently into a clean latent $z_0\in\mathbb{R}^{C\times H\times W}$; a $2\times2$ spatial patchifier yields $T=HW/4$ tokens ($C=16$, $T=1560$ at $480\times832$).
A clean $t=0$ prefill commits source keys and values $(K_0,V_0)$, which remain pinned throughout generation.
Each step $n\ge1$ generates one latent $z_n$ (four decoded frames) using the student's original denoising and clean cache-refresh schedule.

We inherit Echo-Forcing's memory host~\cite{wu2026echoforcing}, together with source pinning~\cite{li2026packcache} and Block-Relativistic RoPE~\cite{yesiltepe2025infinity,minar2026steadyforcing}.
These components provide a common source reference and bounded historical context; our contribution changes how the newest generated history is read, not how that history is stored or evicted.

\subsection{Estimating activity once}
\label{fm:method_map}

After the first six generated latents $z_1,\ldots,z_6$ (1.5\,s, excluding $z_0$ to avoid conditioning transients), we compute temporal population variance and average it across latent channels and each $2\times2$ patch $\mathcal P(p)$ associated with token $p$:
\begin{equation}
\begin{split}
v(p)&=\frac{1}{4C}\sum_{u\in\mathcal P(p)}\sum_{c=1}^{C}
\operatorname{Var}_{1\le j\le6}z_j(c,u),\\
L(p)&=\log_{10}\big(v(p)+10^{-12}\big).
\end{split}
\label{eq:fm_var}
\end{equation}
Here $u$ indexes latent-grid locations, $c$ channels, and $p\in\{1,\ldots,T\}$ transformer tokens.
The logarithm compresses the variance range before Otsu's threshold $\tau$~\cite{otsu1979threshold} defines a binary high-activity mask $M(p)=\mathbf{1}[L(p)>\tau]$.

$M$ and its tether weights are computed only from committed generated latents, never from future frames or evaluation masks, and stay frozen for the rest of the rollout.
Freezing prevents subsequent appearance drift from redefining which positions are anchored, although it cannot capture motion that begins after warm-up.
The mask is an activity proxy, not a semantic segmentation of moving objects.

For a transition band at static--dynamic boundaries, let $D(M)$ be the one-token, eight-neighbor dilation of $M$.
The static-interior anchoring weight is
\begin{equation}
w(p)=\operatorname{clip}\!\left(\frac{\tau-L(p)}{\eta},0,1\right)
\mathbf{1}[p\notin D(M)],
\label{eq:fm_map}
\end{equation}
with $\eta=0.3$.
If Otsu's split is degenerate or non-finite, we set $w=0$, so all non-moving positions receive the fallback strength $\lambda_b$ (\cref{eq:fm_beta}).

\subsection{Reading the newest frame with a source tether}
\label{fm:method_read}

During warm-up ($n\le6$), reads are untethered.
From step $n=7$, at every layer, head, and denoising or clean-refresh evaluation, the newest committed latent $z_{n-1}$ supplies full-grid keys and values $K_r,V_r\in\mathbb{R}^{T\times d_h}$.
They are paired with source values $V_0$ from the same layer and head ($d_h=128$; Supp.~Sec.~A).
Source values are computed once during prefill and retrieved at matching spatial indices throughout the rollout.

We preserve $K_r$ and its block-relativistic RoPE coordinates while replacing the temporary read values with
\begin{equation}
\widetilde V_r(p)=V_r(p)+\beta(p)\big[V_0(p)-V_r(p)\big],
\label{eq:fm_tether}
\end{equation}
where
\begin{equation}
\beta(p)=\begin{cases}
0, & M(p)=1,\\
\lambda w(p), & M(p)=0,\ w(p)>0,\\
\lambda_b, & M(p)=0,\ w(p)=0,
\end{cases}
\label{eq:fm_beta}
\end{equation}
with $\lambda=1$ and transition-band/fallback strength $\lambda_b=0.5$.

High-activity positions retain recent values, static-interior positions interpolate toward source values, and the non-moving transition band receives $\lambda_b$.
This partial tether applies a fixed strength near activity boundaries without removing their recent spatial context.
Because $0\le\beta(p)\le1$, each modified value is a convex interpolation of its source and recent counterparts; this local property does not bound the autoregressive rollout.
Interior weights can fall below $\lambda_b$, so $\beta$ can jump at the band's inner edge.
Spatial-rule controls in \cref{fm:controls} distinguish the benefit of leaving high-activity positions untethered from that of graded interior weights.

The source values $V_0(p)$ are gathered at the same token index $p$, without pixel compositing, key relocation, or feature-similarity retrieval.
Holding queries, keys, and the read set fixed, let $A_r\in\mathbb{R}^{T\times T}$ denote the columns of the current queries' softmax attention (over the full read set) that correspond to $K_r$, and $B=\operatorname{diag}(\beta)$.
Before output projection, the exact per-head attention shift is
\begin{equation}
\Delta O=A_rB(V_0-V_r).
\label{eq:st_output_change}
\end{equation}
Thus, queries retain access to the newest frame's complete spatial context while receiving source-aligned values at selected positions.
Dynamic queries can still attend to anchored background values; unchanged active-token values therefore do not guarantee unchanged motion.

The $t=0$ clean refresh applies the same tethered read.
Although $\widetilde V_r$ is never written directly to the persistent cache, it influences the hidden states from which the host commits subsequent $(K_n,V_n)$ entries.
Consequently, future memory states can change even though the cache-update rule remains unchanged.

\subsection{Older history and cost}
\label{fm:method_unchanged}\label{fm:method_cost}\label{fm:method_alg}

At steady state, each attention read includes three complete $T$-token grids: the pinned source $z_0$, newest latent $z_{n-1}$, and current block $z_n$.
Older generated latents of ages $d\in\{2,3,4,5\}$ retain the top key-norm-ranked
\begin{equation}
b_d=\operatorname{round}(T\rho^d),\qquad 2\le d\le 5,
\label{eq:fm_budget}
\end{equation}
tokens, with $\rho=0.3$ and respective budgets $140,42,13,4$.
Excluding Echo's rotating anchors and compressed entries, the steady-state attention read contains
\begin{equation}
|\mathcal R|=3T+\sum_{d=2}^{5}b_d=4879
\label{eq:fm_cost}
\end{equation}
tokens per layer, independent of rollout duration or activity-map area.

This is an attention-read count rather than allocated KV-cache memory.
The source and newest latents remain fully available, while only older history is thinned.
\Ours adds no denoising evaluations and incurs only a small measured gather-and-blend overhead (\cref{fm:cost}).
The method does not guarantee identity preservation, physically correct dynamics, or artifact-free generation; these properties require empirical evaluation.

\begin{table*}[t]
 \centering
 \scriptsize
 \setlength{\tabcolsep}{2pt}
 \caption{\textbf{Quantitative comparison.}
 Results on the frozen two-step framewise Causal Forcing++~\cite{zhao2026causal_pp} generator (48 sources, seed 42; $k$ denotes latent block size).
 Each panel scores the late window of rollouts of the stated length.
 \Ours achieves the highest source-support PSNR and lowest color error across all three horizons, alongside active late-stage material motion.
\textbf{Bold}/\underline{underline}: best/second-best per horizon, excluding activity and dynamic degree.
 Eligibility rules and adapted-method settings are described in Sec.~\ref{fm:protocol} and Supp.~Tab.~24.}
 \label{tab:fm_main}
 \label{tab:st_main_horizons}
 \begin{tabular}{@{}>{\raggedright\arraybackslash}p{.225\textwidth}*{9}{>{\centering\arraybackslash}p{.078\textwidth}}@{}}\toprule
 & \multicolumn{3}{c}{SNF-Bench} & \multicolumn{2}{c}{VBench-I2V} & \multicolumn{4}{c}{VBench-Long}\\
\cmidrule(lr){2-4}\cmidrule(lr){5-6}\cmidrule(lr){7-10}
Method & \shortstack{Grayscale\\peak signal-\\to-noise ratio\\(dB) $\uparrow$} & \shortstack{Mean-Lab\\color error\\$\downarrow$} & \shortstack{Late material\\activity\\(pixels/pair)} & \shortstack{Subject\\consistency\\$\uparrow$} & \shortstack{Background\\consistency\\$\uparrow$} & \shortstack{Motion\\smoothness\\$\uparrow$} & \shortstack{Dynamic\\degree} & \shortstack{Aesthetic\\quality\\$\uparrow$} & \shortstack{Imaging\\quality\\$\uparrow$}\\\midrule
\multicolumn{10}{l}{\textbf{60 seconds}}\\
Framewise Echo, $k=1$ & 16.97 & 15.54 & 1.02 & \textbf{0.981} & \textbf{0.971} & \underline{0.990} & 0.219 & 0.538 & 0.674\\
Static sink, $k=3$ & 17.29 & 13.72 & 0.69 & 0.969 & 0.962 & 0.989 & 0.082 & 0.528 & \textbf{0.699}\\
Age-decay read, $k=1$ & \underline{18.06} & \underline{7.35} & 1.16 & 0.952 & 0.948 & 0.985 & 0.154 & \underline{0.544} & 0.649\\
Echo-Forcing port, $k=3$ & 16.88 & 15.08 & 0.50 & 0.970 & 0.961 & \textbf{0.991} & 0.050 & 0.539 & \underline{0.697}\\
ALG-inspired, $k=1$ & 15.87 & 19.96 & 1.07 & 0.977 & \underline{0.969} & \underline{0.990} & 0.249 & 0.537 & 0.663\\
DyMoS port, $k=1$ & 16.89 & 16.95 & 0.97 & \underline{0.979} & \underline{0.969} & \underline{0.990} & 0.228 & 0.539 & 0.671\\
UnStep port, $k=3$ & 16.36 & 16.14 & 0.55 & 0.966 & 0.960 & \textbf{0.991} & 0.065 & 0.537 & 0.692\\
PackCache-style, $k=1$ & 16.48 & 9.23 & 1.21 & 0.963 & 0.956 & 0.986 & 0.249 & \underline{0.544} & 0.663\\
FramePack-style, $k=1$ & 14.39 & 16.17 & 1.97 & \underline{0.979} & \underline{0.969} & 0.989 & 0.394 & \underline{0.544} & 0.684\\
TetherCache port, $k=1$ & 14.38 & 23.47 & 1.32 & 0.978 & \underline{0.969} & \underline{0.990} & 0.276 & 0.519 & 0.669\\
\textbf{SceneTether (ours)} & \textbf{22.37} & \textbf{5.57} & 1.97 & \underline{0.979} & \underline{0.969} & 0.989 & 0.366 & \textbf{0.547} & 0.684\\
\midrule
\multicolumn{10}{l}{\textbf{120 seconds}}\\
Framewise Echo, $k=1$ & 17.11 & 15.31 & 0.77 & 0.914 & 0.916 & \underline{0.990} & 0.218 & 0.523 & 0.657\\
Static sink, $k=3$ & 16.91 & 14.33 & 0.73 & 0.883 & 0.893 & 0.988 & 0.120 & 0.494 & \underline{0.689}\\
Age-decay read, $k=1$ & \underline{18.12} & \underline{7.01} & 1.11 & 0.922 & 0.925 & 0.984 & 0.167 & \underline{0.543} & 0.649\\
Echo-Forcing port, $k=3$ & 16.37 & 17.52 & 0.46 & 0.890 & 0.898 & \textbf{0.991} & 0.042 & 0.523 & 0.668\\
ALG-inspired, $k=1$ & 15.69 & 20.69 & 0.97 & 0.894 & 0.894 & 0.988 & 0.234 & 0.510 & 0.636\\
DyMoS port, $k=1$ & 16.85 & 16.40 & 0.82 & 0.911 & 0.909 & 0.989 & 0.182 & 0.527 & 0.654\\
UnStep port, $k=3$ & 16.38 & 15.42 & 0.63 & 0.899 & 0.904 & \underline{0.990} & 0.058 & 0.535 & \textbf{0.691}\\
PackCache-style, $k=1$ & 16.56 & 9.41 & 1.21 & \underline{0.934} & \underline{0.933} & 0.986 & 0.250 & \textbf{0.544} & 0.662\\
FramePack-style, $k=1$ & 14.52 & 15.67 & 2.06 & 0.927 & 0.930 & 0.989 & 0.395 & 0.530 & 0.685\\
TetherCache port, $k=1$ & 14.04 & 26.50 & 1.36 & 0.859 & 0.875 & \underline{0.990} & 0.228 & 0.466 & 0.646\\
\textbf{SceneTether (ours)} & \textbf{22.46} & \textbf{5.53} & 2.09 & \textbf{0.954} & \textbf{0.947} & 0.989 & 0.360 & \textbf{0.544} & 0.685\\
\midrule
\multicolumn{10}{l}{\textbf{240 seconds}}\\
Framewise Echo, $k=1$ & 17.48 & 14.98 & 0.84 & 0.892 & 0.900 & 0.987 & 0.200 & 0.511 & 0.639\\
Static sink, $k=3$ & 17.07 & 15.07 & 0.76 & 0.860 & 0.876 & 0.986 & 0.238 & 0.486 & \textbf{0.693}\\
Age-decay read, $k=1$ & \underline{18.22} & \underline{6.99} & 1.03 & 0.926 & 0.926 & 0.985 & 0.165 & \underline{0.544} & 0.649\\
Echo-Forcing port, $k=3$ & 15.85 & 19.71 & 0.35 & 0.831 & 0.852 & \textbf{0.990} & 0.042 & 0.497 & 0.618\\
ALG-inspired, $k=1$ & 15.71 & 20.17 & 1.62 & 0.862 & 0.871 & 0.985 & 0.253 & 0.492 & 0.633\\
DyMoS port, $k=1$ & 16.58 & 17.84 & 1.15 & 0.888 & 0.892 & 0.987 & 0.213 & 0.516 & 0.639\\
UnStep port, $k=3$ & 16.47 & 16.03 & 0.60 & 0.895 & 0.901 & \textbf{0.990} & 0.072 & 0.532 & \underline{0.692}\\
PackCache-style, $k=1$ & 16.41 & 9.46 & 1.22 & \underline{0.933} & \underline{0.933} & 0.986 & 0.247 & 0.543 & 0.661\\
FramePack-style, $k=1$ & 14.24 & 18.77 & 2.19 & 0.918 & 0.925 & 0.988 & 0.386 & 0.525 & 0.689\\
TetherCache port, $k=1$ & 13.35 & 29.35 & 1.05 & 0.794 & 0.818 & \textbf{0.990} & 0.251 & 0.428 & 0.624\\
\textbf{SceneTether (ours)} & \textbf{22.47} & \textbf{5.69} & 1.91 & \textbf{0.953} & \textbf{0.948} & \underline{0.989} & 0.344 & \textbf{0.545} & 0.684\\
\bottomrule\end{tabular}

\end{table*}

\begin{figure*}[t]
 \centering
 \raisebox{-52.22684pt}{\includegraphics{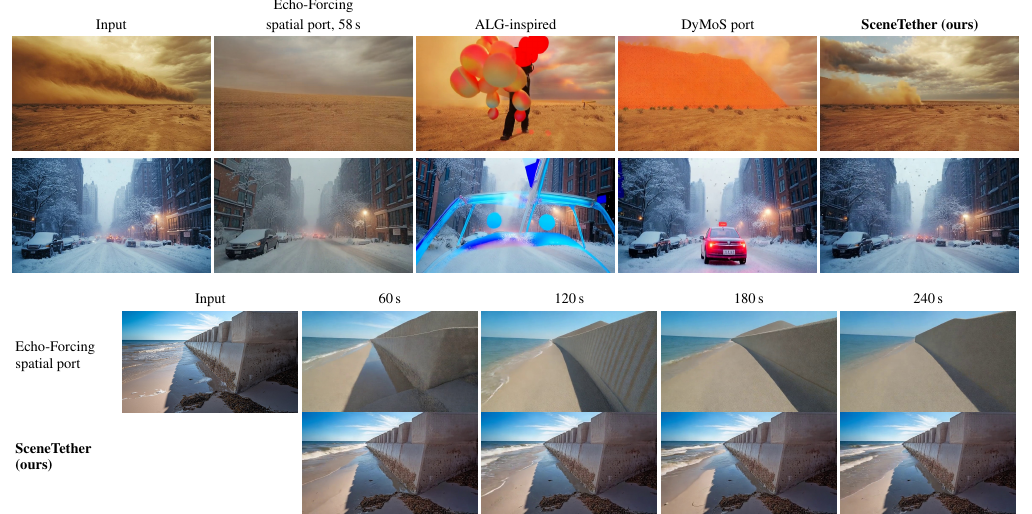}}
 \caption{\textbf{Qualitative comparisons.}
 \textit{Top:} Selected dust and snowfall examples at 58\,s, where \Ours retains the source palette and scene structure more closely than the adapted baselines.
 \textit{Bottom:} Frames from a single 240\,s breaking-wave rollout, showing greater preservation of the seawall under \Ours as spatial Echo-Forcing progressively drifts.
 More samples and failures are in supp.}
 \label{fig:st_main_comparison}
\end{figure*}

\begin{figure*}[t]
 \centering
 \raisebox{-54.68709pt}{\includegraphics{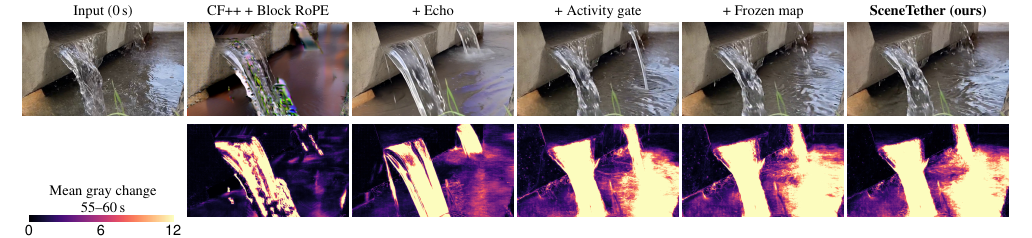}}
 \caption{\textbf{Visual progression across cumulative construction.}
 Activity-gated reading restores measured water activity, map freezing reduces background changes, and source-value tethering improves the appearance of the concrete structure while retaining active spillway motion.
 Bottom row shows mean frame-to-frame grayscale change over 55--60\,s on a shared 0--12 scale.}
 \label{fig:st_main_ablation}
\end{figure*}

\begin{figure*}[t]
 \centering
 \includegraphics[width=.75\textwidth]{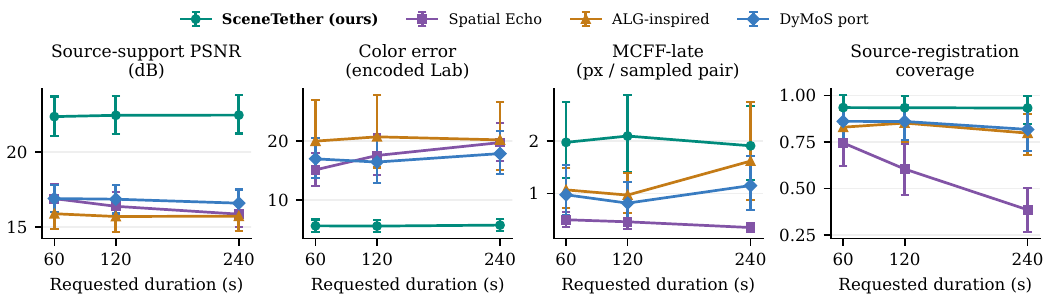}
 \caption{\textbf{Performance across multi-minute horizons.}
 Late-window results from separately generated 60, 120, and 240\,s rollouts, evaluated against the original source image (\cref{tab:st_main_horizons}).
 Points show scene means with descriptive 95\% bootstrap intervals (seed 42).
 \Ours maintains higher source-support fidelity at every endpoint, while the adapted methods exhibit appearance drift, activity change, and source-registration loss.
 Within-rollout trajectories, including the framewise host and age-decay read, appear in Supp.~Fig.~4 and Supp.~Tab.~15.}
 \label{fig:st_horizon}
\end{figure*}

\section{Source-fixed evaluation protocol}
\label{fm:protocol}

\paragraph{Sources and regions.}
We extend SNF-Bench~\cite{snfbench2026} with per-source evaluation masks for seven development, 48 benchmark, and 24 separately collected confirmation scenes.
Static and dynamic supports are shared across methods and never used in generation.
These evaluation masks are distinct from \Ours's frozen activity map, which is estimated solely from generated latents.
The benchmark has 42 appearance-, 40 activity-, and 45 geometry-eligible sources, and registration failures are tracked at matched timestamps.
Masks were initialized by text-prompted segmentation; the authors visually checked benchmark masks, whereas confirmation masks received AI-only review. They are shared evaluation supports, not independently annotated ground truth.

\paragraph{Metrics.}
Source-fidelity metrics compare raw outputs with the source image, not earlier generated frames, using late-window samples at matched scene times.
Static-support metrics are grayscale PSNR, mean-Lab error (Euclidean distance between spatial-mean OpenCV Lab vectors; not pixelwise CIEDE2000, so local shifts can cancel), and registration coverage, the fraction of sampled late-window frames that register to the source (alignment success, not accuracy).
Late dynamic activity is Motion-Compensated Foreground Flow (MCFF-late): similarity-compensated RAFT flow~\cite{teed2020raft} at 288-pixel height and 8\,Hz, in pixels per sampled pair.
Paired 60\,s contrasts use registered native-resolution Farneb\"ack flow~\cite{farneback2003} on one adjacent-frame pair per second over 40--60\,s, excluding invalid fits; the two estimators differ in resolution, window, and population and are not directly comparable.
Because flow amplitude does not establish naturalness, we also report individual VBench-I2V and VBench-Long dimensions~\cite{huang2023vbench,huang2024vbench++} on the same 48 sources.

\paragraph{Matched comparisons and uncertainty.}
Main comparisons share the pretrained \CFpp checkpoint, source, caption, horizon, and latent-indexed noise.
Causal ports and their inference adaptations are listed in Supp.~Tab.~24 (\eg, UnStep uses low-rank projection approximations~\cite{mansour2026unstep}, and the ALG-inspired adaptation~\cite{choi2025alg} blurs early-step source K/V because the distilled student has no CFG branch); native large-model pipelines keep their own samplers and are compared separately (Supp.~Tabs.~2 and~31).
\Cref{tab:fm_main} reports seed 42; paired Echo contrasts nest seeds 42 and 1042 within scenes, with descriptive, unadjusted 95\% bootstrap intervals that resample whole scenes.

\section{Experiments}
\label{fm:experiments}\label{fm:main}

\begin{figure*}[t]
 \centering
 \raisebox{-29.15993pt}{\includegraphics{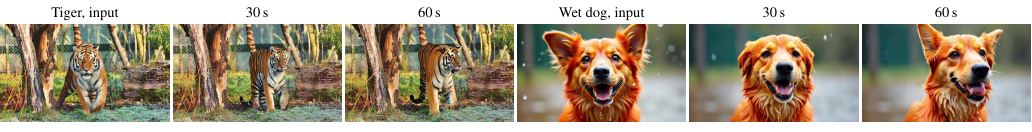}}
 \caption{\textbf{Out-of-domain exploration.} Selected 60\,s articulated-subject examples from VBench-I2V and DyMoS retain background structure with \Ours's default settings; further successes and failures are in Supp.~Figs.~10--11.}
 \label{fig:fm_oos}
\end{figure*}

\begin{figure}[t]
 \centering
 \raisebox{-27.52777pt}{\includegraphics{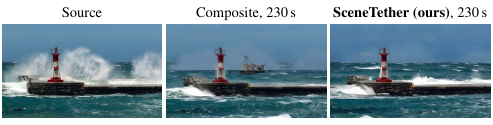}}
 \caption{\textbf{In-attention tethering vs.\ pixel compositing.} Post-hoc composite (same frozen map) keeps moving-region artifacts, such as boundary ghosts, from the untethered rollout, which \Ours avoids (Supp.~Figs.~20 and~24).}
 \label{fig:fm_composite}
\end{figure}

\paragraph{Comparisons with adapted methods.}
\label{fm:backbones}
On the same frozen two-step \CFpp generator, \Ours achieves the highest source-support PSNR and lowest mean-Lab color error among evaluated memory policies and causal guidance ports at 60, 120, and 240\,s (\cref{tab:fm_main}).
Paired over two seeds at 60\,s, \Ours improves PSNR by \stres{spatial_psnr}\,dB over spatial Echo-Forcing and by 5.15\,dB over its framewise Echo host, with $1.50\times$ the host's registered late activity (Supp.~Tab.~5).
Against age-decay memory, the seed-42 advantage is about 4.3\,dB at every horizon.
VBench results are mixed: framewise Echo slightly leads I2V consistency at 60\,s, whereas \Ours leads both subject and background consistency at 120 and 240\,s and leads or ties aesthetic quality throughout. Other policies sometimes score higher on imaging quality. 
At 240\,s, the FramePack-style policy retains higher late activity (2.19 vs.\ 1.91 pixels/pair) but substantially lower source-support PSNR (14.24 vs.\ 22.47\,dB), illustrating the fidelity--activity trade-off.
Qualitative results appear in \cref{fig:st_main_comparison}, with failure cases, including object hallucinations, in Supp.~Fig.~9.

\paragraph{System ablation.}
\label{fm:controls}
\Cref{tab:st_main_ablation} (upper) and \cref{fig:st_main_ablation} add block-relativistic RoPE, Echo memory, activity-gated reads, map freezing, and source-value tethering in order.
Because some stages also change source pinning, history selection, or coverage, this is cumulative construction, not single-factor decomposition.
The equal-coverage and spatial-rule controls below separately test the contribution of value anchoring and importance of leaving high-activity regions untethered.

\begin{table}[t]
 \centering\scriptsize\setlength{\tabcolsep}{1.1pt}\renewcommand{\arraystretch}{0.9}
 \caption{\textbf{Construction and isolated controls at 60\,s} (seed 42). \textit{Upper:} cumulative stages. \textit{Lower:} equal-newest-coverage controls that separate value tethering from coverage, and a post-hoc composite. Two-seed paired estimates and intervals are in Supp.~Tabs.~5 and~8.}
 \label{tab:st_main_ablation}
 \begin{tabular}{@{}>{\raggedright\arraybackslash}p{.36\linewidth}*{6}{>{\centering\arraybackslash}p{.09\linewidth}}@{}}\toprule
Stage & \shortstack{PSNR\\$\uparrow$} & \shortstack{Color\\$\downarrow$} & Act. & \shortstack{I2V\\Subj.\\$\uparrow$} & Dyn. & \shortstack{Image\\$\uparrow$}\\\midrule
CF++ + Block RoPE & 15.70 & 20.39 & 2.08 & 0.975 & 0.264 & 0.655\\
+ Echo & \underline{16.97} & 15.54 & 1.02 & \textbf{0.981} & 0.219 & \underline{0.674}\\
+ Activity-gated read & 15.45 & 11.84 & 2.38 & 0.972 & 0.416 & 0.667\\
+ Frozen activity map & 16.42 & \underline{11.24} & 2.36 & 0.971 & 0.412 & 0.664\\
\shortstack[l]{+ Value tether\\\textbf{SceneTether (ours)}} & \textbf{22.37} & \textbf{5.57} & 1.97 & \underline{0.979} & 0.366 & \textbf{0.684}\\
\bottomrule\end{tabular}

 \par\smallskip
 \begin{tabular}{@{}>{\raggedright\arraybackslash}p{.36\linewidth}*{6}{>{\centering\arraybackslash}p{.09\linewidth}}@{}}\toprule
\multicolumn{7}{l}{\textbf{Equal newest-frame coverage}}\\
Full coverage, no tether & 14.32 & 15.66 & 2.33 & \underline{0.978} & 0.372 & \underline{0.682}\\
Half tether & 19.44 & 5.71 & 1.87 & 0.977 & 0.383 & 0.672\\
\textbf{SceneTether (ours)} & \underline{22.37} & \underline{5.57} & 1.97 & \textbf{0.979} & 0.366 & \textbf{0.684}\\
Composite (post-hoc) & \textbf{22.57} & \textbf{5.46} & 1.66 & 0.977 & 0.267 & 0.665\\
\bottomrule\end{tabular}

\end{table}

\paragraph{Isolating value anchoring and spatial gating.}
With 100\% newest-frame coverage held fixed, tethering raises PSNR by \stres{coverage_psnr}\,dB over the untethered read; its two-seed flow ratio \stres{coverage_flow} has an interval spanning one, establishing neither motion loss nor equivalence (Supp.~Tabs.~5--6).
Full-strength tethering also exceeds half-strength tethering in fidelity.
Applying uniform or permuted weights to the whole frame, including $M$, reduces activity to \stres{ctl_uni_ratio}$\times$ and \stres{ctl_perm_ratio}$\times$ that of \Ours.
By contrast, keeping $\beta=0$ on active positions $M$ while flattening or permuting only the remaining weights yields inconclusive flow differences (ratios \stres{ctl_suni_ratio}, \stres{ctl_sperm_ratio}) and slightly higher PSNR (\stres{ctl_suni_dpsnr}, \stres{ctl_sperm_dpsnr}\,dB).
Thus sparing $M$, not the exact interior weight grading, limits motion suppression (Supp.~Tab.~25).

\paragraph{Against post-hoc compositing.}
Blending source pixels into the untethered rollout using the identical frozen map (Supp.~Sec.~C) changes source-support PSNR by $+0.20$\,dB at 60\,s and $+2.45$\,dB at 240\,s relative to \Ours; both confidence intervals include zero (Supp.~Tabs.~8 and~26).
Static PSNR alone therefore cannot distinguish the approaches.
However, the composite largely retains drift within active regions: moving-region mean-Lab drift is \stres{drift60_comp} versus \stres{drift60_st} for \Ours at 60\,s (worse on \stres{drift60_worse}/\stres{drift60_n} sources) and \stres{drift240_comp} versus \stres{drift240_st} at 240\,s (\stres{drift240_worse}/\stres{drift240_n}; Supp.~Tab.~28).
It also reduces variation in anchored regions (\stres{comp_still_flicker} vs.\ \stres{st_still_flicker} gray levels), while selected examples show boundary ghosts or retained inventions (\cref{fig:fm_composite}; Supp.~Fig.~24).
In matched-encoding full-video comparisons, raters preferred \Ours over the composite (\cref{tab:st_main_human}).

\paragraph{Long-horizon and fresh-source results.}
\label{fm:horizon}\label{fm:fresh}
On the twelve preselected four-minute sources (\stres{h12_psnr_n} appearance-eligible), \Ours gains \stres{h12_psnr}\,dB over spatial Echo-Forcing at 240\,s with more measured late activity (Supp.~Tab.~14).
Across 45 geometry-eligible benchmark sources, source-registration coverage at 240\,s is \stres{cov_st_240} for \Ours and 92.7\% for age-decay memory; spatial Echo-Forcing falls from \stres{cov_sp_60} at 60\,s to \stres{cov_sp_240} at 240\,s (\cref{fig:st_horizon}; Supp.~Tab.~15).
On 24 separately collected confirmation sources, a two-seed evaluation yields a 4.40\,dB gain over spatial Echo-Forcing at 60\,s (21 appearance-eligible; Supp.~Tab.~16). Their masks were AI-reviewed, not independently human-annotated.

\paragraph{Human evaluation.}
\label{fm:human}
Blinded viewers judged scene preservation, motion naturalness, and artifacts alongside the source image (\cref{tab:st_main_human}; Supp.~Sec.~F).
Preferences favor \Ours over age-decay memory (74.8\%--76.7\%) and over the matched-encoding composite (75.0\%--97.9\%) in full one-minute videos.
The matched-encoding comparison involved six raters and 36 votes, providing encouraging but limited perceptual evidence.
Against spatial Echo-Forcing, preferences are 66.5\%, 54.9\%, and 66.5\% for scene preservation, motion naturalness, and fewer artifacts, but all three 95\% intervals include 50\%, so this comparison is inconclusive (all-response analysis and study in Supp.~Sec.~F).
The four-minute module assesses scene preservation, repetition, and discontinuities rather than overall motion naturalness.

\begin{table}[t]
 \centering\scriptsize\setlength{\tabcolsep}{3pt}\renewcommand{\arraystretch}{0.92}
 \caption{\textbf{Blinded human evaluation.} \Ours preference (ties counted as half). Late-window, one-minute, and four-minute modules assess their stated criteria separately; one-minute sources are equally weighted. Full intervals and per-module counts in Supp.~Sec.~F.}
 \label{tab:st_main_human}
 \begin{tabular}{@{}lrrrr@{}}\toprule
\multicolumn{5}{c}{\textbf{Preference for SceneTether (ours), \%}}\\
Reference & \shortstack{Scene\\preserv.} & \shortstack{Motion\\natural.} & \shortstack{Fewer\\artifacts} & \shortstack{Raters/\\votes}\\\midrule
\multicolumn{5}{l}{\textit{Baseline variants, last 10\,s clips}}\\
\shortstack[l]{Echo-Forcing\\spatial port} & 66.5 & 54.9 & 66.5 & 14/112\\
Age-decay read & 76.7 & 75.2 & 74.8 & 14/105\\
\midrule
\multicolumn{5}{l}{\textit{Guidance module, last 10\,s clips}}\\
ALG-inspired & 80.4 & 71.4 & 67.9 & 7/28\\
DyMoS port & 67.9 & 58.9 & 57.1 & 7/28\\
\midrule
\multicolumn{5}{l}{\textit{Full one-minute videos}}\\
Composite (original) & 88.5 & 80.6 & 87.2 & 7/42\\
Composite (matched) & 97.9 & 75.0 & 91.7 & 6/36\\
\midrule
\multicolumn{5}{l}{\textit{Full four-minute videos}}\\
Reference & Scene & \shortstack{Less\\repetition} & \shortstack{Fewer\\discontinuities} & \shortstack{Raters/\\votes}\\
\shortstack[l]{Echo-Forcing\\spatial port} & 92.2 & 89.1 & 93.8 & 8/32\\
\bottomrule\end{tabular}
\end{table}

\paragraph{Efficiency and transfer.}
\label{fm:cost}
On one exclusive A100, the two-step \CFpp host runs at \stres{cfpp2_host_fps}\,FPS (\stres{cfpp2_host_fpswall}\,FPS including VAE decoding), versus \stres{cfpp2_st_fps}\,FPS (\stres{cfpp2_st_fpswall}\,FPS) for \Ours, approximately \stres{cfpp2_overhead_pct}\% extra wall time and no extra denoising evaluations (Supp.~Tab.~23).
Both rates are below the 16\,FPS output rate, so neither system runs in real time.
With identical settings, tethering improves the one-step \CFpp student's PSNR by \stres{onestep_psnr}\,dB at 60\,s (Supp.~Tabs.~22 and~27), but on Self-Forcing~\cite{huang2025selfforcing} it retains only \stres{sf_ratio}$\times$ of the host's activity despite better static fidelity (Supp.~Tab.~32).
On articulated subjects, results are mixed; failures occur when motion begins after warm-up and falls outside the frozen map (\cref{fig:fm_oos}; Supp.~Figs.~10--11).

\section{Conclusion}
\label{fm:conclusion}
\label{fm:limitations}
\label{fm:failures}
\label{fm:future}

We introduced \Ours, a training-free attention-read policy for long-horizon, fixed-camera image animation. By retaining complete newest-frame coverage while anchoring low-activity values to the source image, \Ours improves scene fidelity and color stability across multi-minute \CFpp rollouts without additional denoising evaluations, and unlike post-hoc pixel compositing, limits moving-region color drift. These results support separating the spatial context provided by recent memory from the appearance it carries.

\paragraph{Limitations and future work.}
Our frozen activity map can mistake illumination changes for motion or suppress dynamics that emerge after warm-up. Source-value anchoring can also propagate initial encoding errors and soften image details (Supp.~Tab.~29). Although \Ours gains are established on the \CFpp students, exploratory transfers to Self-Forcing~\cite{huang2025selfforcing} and MAGI-1~\cite{ai2025magi1autoregressivevideogeneration} reveal backbone-dependent limitations. Due to our fixed-camera formulation, we did not handle camera movement or scene cuts, and optical-flow activity alone cannot establish motion naturalness. Extending \Ours to broader video backbones and developing more perceptually grounded long-horizon motion evaluation remain promising directions.

\clearpage\section*{LLM Usage Statement}
OpenAI ChatGPT/Codex and Anthropic Claude Code assisted with literature discovery, code and test development, experiment scheduling, execution, diagram preparation, the review and repair of the automatically segmented evaluation-region masks (\cref{fm:protocol}), and manuscript revising and editing. The authors remain responsible for verifying all claims, citations, implementations, and reported results.

{
    \small
    \bibliographystyle{ieeenat_fullname}
    \bibliography{main}
}

\backrefsetup{disable}
\clearpage
\setcounter{page}{1}
\maketitlesupplementary
\setcounter{section}{0}
\setcounter{table}{0}
\setcounter{figure}{0}
\setcounter{equation}{0}
\renewcommand{\thesection}{\Alph{section}}
\renewcommand{\theHsection}{S\arabic{section}}
\renewcommand{\theHtable}{S\arabic{table}}
\renewcommand{\theHfigure}{S\arabic{figure}}
\renewcommand{\theHequation}{S\arabic{equation}}

\section{Implementation and reproducibility}
\label{fm:suppl_method}

\Ours implements the source-value tether of Sec.~3.4 in the main paper. The \emph{sparse frozen read} that appears in component and development studies is an earlier diagnostic variant that reads only a map-selected subset of newest-frame tokens instead of tethering their values; it is a control, not \Ours.
The equal-coverage untethered control matches the source, caption, checkpoint, sampler, video encoding, and memory host of \Ours and disables only value tethering. \Ours applies no color correction, super-resolution, or output compositing.
For every reported run, we record SHA-256 hashes of the model checkpoint, inference code, and outputs together with the exact command-line arguments.

\begin{figure}[ht]
 \centering
 \fbox{\parbox{.95\linewidth}{\raggedright\small
 \textbf{Algorithm 1: SceneTether}\\[3pt]
 Encode the source image; commit its K/V at $t=0$ and pin the source entry. Set $\beta=0$ until the activity map is installed.\\
 For each new latent block:\\
 \hspace*{1em}If six generated latents exist and no activity map is installed:\\
 \hspace*{2em}estimate log-variance activity and anchoring (taper) weights; freeze both.\\
 \hspace*{1em}For each denoising step and transformer layer:\\
 \hspace*{2em}read pinned source and current block in full;\\
 \hspace*{2em}when generated history exists, read its newest latent at every position with unchanged keys;\\
 \hspace*{2em}interpolate newest-frame values toward co-located source values\\
 \hspace*{3em}using Eq.~(4) of the main paper ($\beta=0$ before map installation and on high-activity tokens);\\
 \hspace*{2em}thin older history by its age budget; compute self-attention on this bank.\\
 \hspace*{1em}At $t=0$, refresh clean K/V using the same read policy and commit through the host update rule.\\
 Decode generated latents with the unmodified causal VAE.
 }}
 \caption{\textbf{Inference overview.} \Ours changes the temporary values assembled for attention reads, not the persistent cache-write rule. During the first six generated latents, the newest history remains untethered.}
\end{figure}

\paragraph{Timing.}
At 16 output frames per second, each generated latent advances scene time by $0.25\,\text{s}$.
\Ours installs the activity map immediately after committing the first six generated latents ($z_1,\dots,z_6$), before block~7 begins.
The temporal variance excludes $z_0$: it is the independently encoded clean source image, not a generated latent, so including it would add a conditioning transient to the variance. Otsu's threshold $\tau$ is then applied to the pooled log-variance $L(p)$.
We freeze this initial map and its anchoring (taper) weights $w(p)$ for the remainder of the rollout; although diagnostic runs may log subsequent maps, inference never uses them.

\paragraph{Anchoring (taper) weights and exceptional partitions.}
We set the log-variance ramp width to $\eta=0.3$ and compute the one-token boundary dilation $D(M)$ over an eight-neighbor spatial stencil.
If temporal variances are non-finite or $\tau$ assigns every token to the same class, the implementation sets the interior weight $w(p)=0$ everywhere; low-activity tokens then receive the fallback band strength $\lambda_b=0.5$, while high-activity tokens keep $\beta=0$.
This fallback keeps execution well defined on degenerate inputs; it does not claim a meaningful spatial partition.
The transition band $D(M)\setminus M$ is not the uncertain \emph{selection} band of the sparse frozen read: that variant pruned tokens, whereas \Ours keeps all $T$ spatial positions and assigns $\lambda_b=0.5$ to non-moving tokens with $w(p)=0$.

\paragraph{Map statistics and band strength.}
Across the 48 benchmark sources, frozen maps classify $40.0\%$ of tokens as high activity ($8.7\text{--}75.5\%$), assign positive interior weights to $42.1\%$, and place $17.9\%$ ($7.1\text{--}42.7\%$) in the transition band, yielding a mean tether strength $\beta$ of $0.449$ ($0.192\text{--}0.835$).
On every benchmark source, the $w(p)=0$ non-moving set equals the one-token ring $D(M)\setminus M$, with zero degenerate partitions.
We pre-specified $\lambda_b=0.5\lambda$, evaluating coupled arms $(\lambda,\lambda_b)=(0.5,0.25)$ and $(1,0.5)$ before varying $\lambda_b$ independently (\cref{tab:extra_sensitivity}). This choice is a heuristic rather than an empirically optimal boundary weight.
Because interior weights fall below $0.5$ on $4.2\%$ of tokens on average, $\beta(p)$ is non-monotonic across the $D(M)$ boundary (Eq.~(4) of the main paper).

\paragraph{Read versus write.}
Because source and newest-frame entries share an identical $30\times52$ token grid, \Ours gathers source values at matching spatial indices $p$ and interpolates only the temporary read tensor.
Within a read, the operation does not overwrite stored values, change the key-norm ranking of older tokens, relocate keys, or alter spatial coordinates and RoPE phases.
Across reads, however, tethered denoising changes the generated latent, and the $t=0$ refresh uses the same tethered read; the hidden states that produce subsequent $(K_n,V_n)$ entries therefore differ from an untethered rollout even though the write rule never stores $\widetilde V_r$ directly.
At $\beta(p)=0$ the read returns that entry's current value bit for bit, but this value is not the one an independent untethered rollout would hold, and queries at high-activity positions still attend to anchored values elsewhere (Eq.~(5) of the main paper).

\paragraph{Source values.}
We resize the input image to $480\times832$ pixels, normalize intensities to $[-1,1]$, and encode it with the Wan2.1 VAE into a single clean latent $z_0\in\mathbb{R}^{16\times60\times104}$, which the $1\times2\times2$ patchifier maps to $T=1560$ tokens on a $30\times52$ grid.
Before generating block~1, one clean forward pass processes $z_0$ at $t=0$ with an empty cache and the caption embedding as cross-attention context, so self-attention contains only source tokens; cross-attention still uses the caption.
At each layer $l$, this prefill commits raw values $V_0^{l}=W_v^{l}h_0^{l}$ (12 heads of dimension 128, without positional encoding) into Echo's pinned first slot.
We compute $V_0^{l}$ once per rollout; every subsequent tethered pass at layer $l$ gathers $V_0^{l}(p)$ directly by spatial index $p$ within the same head, using neither raw RGB pixels nor external image features.

\paragraph{Calls per block.}
Each framewise block generates one latent (four video frames).
Block~1 runs the student's four-step warm start (nominal timesteps $1000, 750, 500, 250$; warped to $1000, 937.5, 833.3, 625$ under shift-5) plus one clean refresh pass at $t=0$ (five passes total).
Each subsequent block runs two denoising steps (nominal $1000, 500$; warped to $1000, 833.3$) plus one $t=0$ refresh pass (three passes total; \cref{tab:st_calls}).
Including source prefill, a $60\,\text{s}$ rollout (240 generated latents, 961 decoded frames including the source) executes $1+5+239\times3=723$ complete transformer forward passes, each traversing 30 layers; \Ours adds none.
Value tethering operates across all 30 layers during both denoising steps and the refresh pass for latents 7--240 (702 of 723 passes), leaving source prefill and the 20 warm-up passes (latents 1--6) untethered.

\begin{table}[t]
 \centering\small\setlength{\tabcolsep}{3pt}
 \caption{\textbf{Transformer passes per block} for a $60\,\text{s}$ rollout (240 generated latents). Nominal denoising timesteps warp to $1000, 937.5, 833.3, 625$ in block~1 and $1000, 833.3$ thereafter under shift-5. Every generated block concludes with a $t=0$ refresh pass that commits its K/V cache, preceded by a single $t=0$ source-prefill pass.}
 \label{tab:st_calls}
 \begin{tabular}{@{}lcccc@{}}\toprule
 Block & Denoising $t$ & Refresh & Passes & Tethered\\\midrule
 Source prefill & -- & -- & 1 & 0\\
 Latent 1 & 1000, 750, 500, 250 & $t=0$ & 5 & 0\\
 Latents 2--6 & 1000, 500 & $t=0$ & $5\times3$ & 0\\
 Latents 7--240 & 1000, 500 & $t=0$ & $234\times3$ & $234\times3$\\\midrule
 Total & & & 723 & 702\\
 \bottomrule\end{tabular}
\end{table}

\paragraph{Inherited components.}
We use the public CF++ framewise two-step EMA checkpoint without weight modifications.
Echo's cache writes, token compression, and eviction rules remain intact, with the source entry pinned across all blocks.
At steady state, Echo's recent store holds five history latents: attention reads the newest latent ($d=1$) in full and thins ages $d\in\{2,\dots,5\}$ via Eq.~(6) of the main paper.
While the early anchor pool is still filling, an age-6 latent can briefly contribute one token before eviction.
Thinning the attention read does not shrink allocated GPU cache memory.

\paragraph{Randomness and reproducibility.}
We set each scene's seed as the base seed plus its fixed prompt index, indexing stochastic re-noising strictly by latent and timestep rather than global RNG consumption.
Every paired baseline, including the spatial Echo-Forcing port, uses this latent-indexed noise; it couples the random inputs but does not make different block sizes or read policies identical. No table mixes in videos generated under a different noise-indexing scheme.

\section{Evaluation protocol and reporting}
\label{fm:suppl_protocol}\label{fm:suppl_stats}

\paragraph{Native short-clip context.}
\Cref{tab:st_native5} contextualizes published I2V backbones and guidance methods over $0\text{--}5\,\text{s}$ on our seven development photographs (automatically generated masks).
Native models keep their own weights, conditioning branches, and samplers; the Self-Forcing and UnStep rows use image-prefix deployment, and the \Ours row uses the CF++ student.
For SNF-Bench metrics only, we resample $15$ and $24\,\text{fps}$ outputs onto the shared $16\,\text{fps}$ grid and include the conditioned initial frame in whole-window means.
Among these seven-scene native references, SVI 2.0 Pro achieves the highest $5\,\text{s}$ source-support fidelity, while \Ours shows competitive short-clip fidelity with measured activity. Multi-minute stability is evaluated separately at $60\text{--}240\,\text{s}$. Precipitation scenes are excluded from quantitative appearance metrics.

\begin{table*}[t]
 \centering\scriptsize\setlength{\tabcolsep}{2pt}
 \begin{tabular}{@{}lrrrrr|rr|rrrr@{}}\toprule
 & & \multicolumn{4}{c|}{SNF-Bench (0--5 s)} & \multicolumn{2}{c|}{VBench-I2V} & \multicolumn{4}{c}{VBench-Long (short input)}\\
\cmidrule(lr){3-6}\cmidrule(lr){7-8}\cmidrule(lr){9-12}
System & Size & PSNR $\uparrow$ & Color $\downarrow$ & Activity & Fail $\downarrow$ & Subj. $\uparrow$ & Bg. $\uparrow$ & Smooth. $\uparrow$ & Dyn. & Aesth. $\uparrow$ & Image $\uparrow$\\\midrule
Wan2.1 + ALG & 14B & 17.41 & 13.32 & 68.90 & \underline{1} & 0.965 & 0.959 & 0.976 & 0.286 & \textbf{0.625} & \underline{0.661}\\
Wan2.1 I2V & 14B & 20.34 & 10.43 & 62.28 & \textbf{0} & 0.976 & 0.964 & 0.977 & 0.476 & 0.605 & 0.650\\
Wan2.2 + DyMoS & A14B & 15.67 & 9.54 & 84.01 & \underline{1} & 0.978 & 0.968 & 0.980 & 0.476 & \underline{0.616} & 0.653\\
Wan2.2 I2V & A14B & 17.79 & \textbf{7.94} & 65.16 & \textbf{0} & \textbf{0.982} & \textbf{0.972} & 0.984 & 0.429 & \underline{0.616} & 0.657\\
Self-Forcing & 1.3B & 17.09 & 12.17 & 25.93 & \underline{1} & 0.958 & 0.956 & \textbf{0.991} & 0.286 & 0.607 & 0.640\\
Self-Forcing + UnStep & 1.3B & 16.92 & 13.86 & 23.70 & \underline{1} & 0.959 & 0.956 & \textbf{0.991} & 0.190 & 0.603 & 0.652\\
SVI 2.0 Pro & A14B & \textbf{23.41} & 8.28 & 40.77 & \textbf{0} & \underline{0.981} & \underline{0.970} & \textbf{0.991} & 0.429 & 0.590 & 0.651\\
Helios-Distilled & 14B & 13.55 & 17.07 & 13.45 & \underline{1} & 0.928 & 0.928 & \underline{0.989} & 0.143 & 0.564 & \textbf{0.665}\\
\textbf{SceneTether (ours)} & 1.3B & \underline{21.61} & \underline{8.00} & 64.81 & \textbf{0} & 0.976 & 0.967 & 0.985 & 0.524 & 0.591 & 0.638\\
\bottomrule\end{tabular}
\par\vspace{5pt}
\begin{tabular}{@{}llrrrr|rrrrr@{}}\toprule
\multicolumn{11}{l}{\textit{Separate archived ten-source cohort: standard VBench, not I2V or Long}}\\
System & $n$ & \multicolumn{4}{c|}{SNF-Bench unavailable} & Bg. $\uparrow$ & Smooth. $\uparrow$ & Image $\uparrow$ & Flicker $\uparrow$ & Aesth. $\uparrow$\\\midrule
Wan2.1 I2V 14B & 10 & -- & -- & -- & -- & \underline{0.965} & \underline{0.982} & \underline{0.608} & \underline{0.974} & \textbf{0.678}\\
Wan2.2 I2V A14B & 10 & -- & -- & -- & -- & \textbf{0.969} & 0.981 & 0.591 & 0.971 & \underline{0.670}\\
LTX-Video 13B distilled & 10 & -- & -- & -- & -- & 0.951 & \textbf{0.992} & \textbf{0.613} & \textbf{0.985} & 0.624\\
\bottomrule\end{tabular}

 \caption{\textbf{Native five-second context and archived short-clip evaluations.}
 \textit{Top:} Seven shared development scenes ($0\text{--}5\,\text{s}$, automatic masks), grouped into SNF-Bench and VBench suites.
 Static PSNR and color error reference the source image; camera-compensated displacement (px/s, $n=5$ after excluding precipitation and one ambiguous plume support) carries no directional arrow, and failures count rollouts with $<50\%$ source-registration coverage.
 Rows keep their system-specific settings, including the prefix-I2V deployments noted in the text; this is system-level context, not controlled architectural isolation, and VBench-Long serves only as a short-clip diagnostic.
 \textit{Bottom:} Standard VBench scores on a separate ten-source archive (no I2V or Long dimensions inferred).
 \textbf{Bold} and \underline{underline} denote best and second-best directional values per cohort; `--' marks uncomputed entries.}
 \label{tab:st_native5}
 \label{tab:st_native5_archive}
\end{table*}

\paragraph{Archived native results and population boundaries.}
\Cref{tab:st_native5_archive} reports archived native Wan and LTX-Video evaluations on a separate ten-source set under standard VBench, providing historical context without ranking them against the seven-source \Ours row (chunkwise causal runs from that archive are omitted).
Throughout this supplement, we keep development, 48-source benchmark, 24-source fresh confirmation, and 12-source horizon-stress populations strictly separate.
For scored comparisons, we report metric-specific eligible populations, keep multiple seeds nested within each source, evaluate registration at matched timestamps, and apply the pre-declared abstentions (including precipitation and insufficient dynamic support) from the fixed source-mask lists.
Evaluation masks were initialized by text-prompted segmentation; we visually checked the benchmark masks, whereas the confirmation masks received AI-only review.

\paragraph{Statistical uncertainty.}
Paired confidence intervals resample whole source scenes while nesting seeds within each scene.
We interpret optical-flow magnitude as motion activity rather than physical realism. Intervals spanning the null ($0\,\text{dB}$ difference or $1.0\times$ ratio) are inconclusive, not evidence of equivalence or non-inferiority; any such claim would require an explicit margin and corresponding confidence bound.
The 24 fresh sources (\cref{st:suppl_fresh}) form a confirmation split that was not used in method development.

\paragraph{Interpreting intervals and human preferences.}
\providecommand{\stHumanSpatialMotionPoint}{54.9}
\providecommand{\stHumanSpatialMotionCI}{[38.1, 70.8]}
Main-paper tables show point estimates; supplementary tables add descriptive, unadjusted 95\% bootstrap intervals for the stated estimand (a scene-mean difference, a ratio of scene means, or a preference score). They are neither video-level min--max ranges nor posterior probabilities, and they do not provide simultaneous coverage across the full comparison family.
For example, the paired two-seed contrast against spatial Echo-Forcing yields $+\stres{spatial_psnr}\,\text{dB}$ with interval \stres{spatial_psnr_ci} (strictly positive), whereas the equal-coverage Farneb\"ack flow ratio is \stres{coverage_flow} with interval \stres{coverage_flow_ci} (spanning $1.0$, establishing neither motion loss nor equivalence).
In human evaluations, ties receive $0.5$ credit ($50\%$ neutral).
Against spatial Echo-Forcing, motion-naturalness preference reaches \stHumanSpatialMotionPoint\% (\stHumanSpatialMotionCI), which spans $50\%$ and remains inconclusive.
Human intervals use a crossed bootstrap that resamples both sources and raters, so ratings from the same source or rater are not treated as independent; these intervals remain descriptive for the sampled source and rater pools.
Source fidelity, flow amplitude, and human preference measure distinct constructs and are not combined into a single headline score.

\paragraph{Longer horizons.}
At every horizon, appearance metrics reference the original source image at matched scene timestamps, whereas flow compares generated frames and VBench scores the generated video. We keep the preselected 12-source $240\,\text{s}$ subset separate from secondary full-benchmark rollouts and exclude sparse-frozen-read development runs.
Whereas Fig.~5 of the main paper connects endpoints of separately generated $60$, $120$, and $240\,\text{s}$ rollouts, \cref{fig:st_within240} tracks intermediate one-minute windows within the same continuous $240\,\text{s}$ rollout.

\paragraph{Metric definitions.}
Static-support PSNR averages unwarped per-frame grayscale PSNR against the preprocessed source over the final $15\%$ of at most 160 uniformly sampled frames.
Color error computes the Euclidean distance between mean OpenCV uint8-encoded Lab vectors of the source and late window on the same support (distinct from pixelwise CIEDE2000; opposing local shifts can cancel).
Source-registration coverage is the fraction of sampled late-window frames whose similarity registration to the source succeeds; rollouts below $50\%$ coverage count as registration failures.
MCFF-late, the late-window Motion-Compensated Foreground Flow of SNF-Bench~\cite{snfbench2026}, averages similarity-compensated RAFT~\cite{teed2020raft} flow magnitude inside the dynamic mask at $288$-pixel height and $8\,\text{Hz}$ over the final $12\%$ of transitions (in evaluation pixels per pair, excluding failed similarity fits).
Paired $60\,\text{s}$ flow ratios instead compute native $832\times480$ Farneb\"ack~\cite{farneback2003} flow on adjacent frames of the $16\,\text{fps}$ video, sampling one pair per second over $40\text{--}60\,\text{s}$ and excluding failed inter-frame registrations; each ratio is a ratio of scene means, not a mean of per-video ratios, on its own valid paired population.

\begin{table}[t]
 \centering\small\setlength{\tabcolsep}{4pt}
 \caption{\textbf{Source registration through two minutes.} Mean fraction of late-window frames successfully registering to the original source across all 45 geometry-eligible scenes. Registration coverage records successful alignment, not error magnitude on unregistered frames, and uses a different eligible population from appearance and activity scores.}
 \label{tab:st_horizon_coverage}
 \begin{tabular}{@{}lrr@{}}\toprule
Method & 60\,s & 120\,s\\\midrule
Framewise Echo, $k=1$ & 87.8\% & 84.3\%\\
Spatial Echo, $k=3$ & 74.4\% & 60.4\%\\
Static sink, $k=3$ & 85.1\% & 82.5\%\\
Age-decay read & \underline{92.0}\% & \underline{92.9}\%\\
ALG-inspired, $k=1$ & 83.1\% & 85.1\%\\
DyMoS port, $k=1$ & 86.0\% & 86.0\%\\
UnStep port, $k=3$ & 80.4\% & 77.7\%\\
PackCache-style, $k=1$ & 91.3\% & 91.5\%\\
FramePack-style, $k=1$ & 84.6\% & 84.8\%\\
TetherCache port, $k=1$ & 79.8\% & 77.1\%\\
\textbf{SceneTether (ours)} & \textbf{93.5}\% & \textbf{93.4}\%\\
\bottomrule\end{tabular}

\end{table}

\paragraph{External suites and runtime profiling.}
We report VBench-I2V and VBench-Long dimensions individually, pairing filtered flicker with its valid video count; camera-tracking evaluations that ran out of memory are reported as uncomputed (`--'), not as abstentions.
Because batch scoring jobs share GPUs, we measure wall-clock throughput only on dedicated single-GPU runs (\cref{tab:st_cost}).

\begin{figure*}[t]
 \centering\input{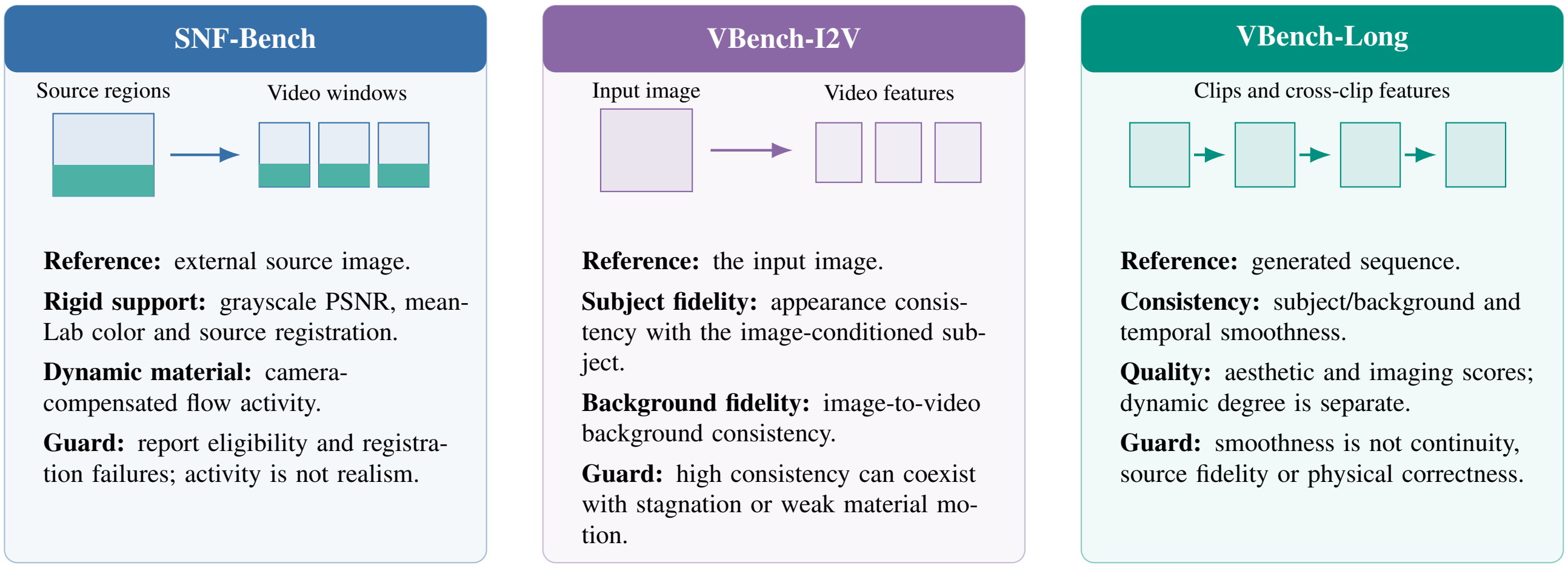}
 \caption{\textbf{Complementary evaluation suites for the scored benchmark rollouts.}
 Our source-fixed SNF-Bench v2 protocol~\cite{snfbench2026} reports static fidelity, registration coverage, and dynamic activity on source-defined regions; VBench-I2V measures input-image consistency; and VBench-Long assesses temporal smoothness, dynamics, and visual quality.
 Because high flow, dynamic degree, or smoothness cannot establish perceptual realism alone, we pair automatic metrics with blinded human evaluations.}
 \label{fig:st_metric_rubrics}
\end{figure*}

\section{One-minute benchmark results}
\label{st:completed}
Two-seed contrasts pair rollouts within each source; the two seeds are nested replicates, not additional independent scenes. \Ours's fidelity and color gains over Echo-Forcing hold for both seeds (\cref{tab:st_paired_completed}).
The equal-coverage control retains \Ours's exact Echo host, pinned source, $100\%$ newest-frame coverage, keys, RoPE coordinates, and pass count while setting $\lambda=\lambda_b=0$ to isolate value interpolation; its two-seed Farneb\"ack flow ratio spans $1.0$ (\cref{tab:st_activity_common}).
For reference, replacing the framewise Echo host ($16.97\,\text{dB}$ PSNR and $15.54$ color error in \cref{tab:st_full60}) with a naive framewise ($k=1$) static sink plus FIFO window (the unedited host of the TetherCache port in Sec.~\ref{fm:suppl_ablations}) drops $60\,\text{s}$ static PSNR to $16.23\,\text{dB}$ $[15.25, 17.24]$ and raises color error to $17.80$ at seed 42.

\begin{table*}[t]
 \centering\scriptsize\setlength{\tabcolsep}{2pt}
 \caption{\textbf{Complete $60\,\text{s}$ comparison including memory references} on the frozen two-step CF++ checkpoint (48 sources, seed 42; 42 appearance- and 40 activity-eligible scenes; all 48 for VBench). \textbf{Bold} and \underline{underline} denote best and second-best directional values; `--' marks uncomputed entries.}
 \label{tab:st_full60}
 \begin{tabular}{@{}>{\raggedright\arraybackslash}p{.225\textwidth}*{9}{>{\centering\arraybackslash}p{.078\textwidth}}@{}}\toprule
 & \multicolumn{3}{c}{SNF-Bench} & \multicolumn{2}{c}{VBench-I2V} & \multicolumn{4}{c}{VBench-Long}\\
\cmidrule(lr){2-4}\cmidrule(lr){5-6}\cmidrule(lr){7-10}
Method & \shortstack{Grayscale\\peak signal-\\to-noise ratio\\(dB) $\uparrow$} & \shortstack{Mean-Lab\\color error\\$\downarrow$} & \shortstack{Late material\\activity\\(pixels/pair)} & \shortstack{Subject\\consistency\\$\uparrow$} & \shortstack{Background\\consistency\\$\uparrow$} & \shortstack{Motion\\smoothness\\$\uparrow$} & \shortstack{Dynamic\\degree} & \shortstack{Aesthetic\\quality\\$\uparrow$} & \shortstack{Imaging\\quality\\$\uparrow$}\\\midrule
Framewise Echo, $k=1$ & 16.97 & 15.54 & 1.02 & \textbf{0.981} & \textbf{0.971} & \underline{0.990} & 0.219 & 0.538 & 0.674\\
Spatial Echo, $k=3$ & 16.88 & 15.08 & 0.50 & 0.970 & 0.961 & \textbf{0.991} & 0.050 & 0.539 & \underline{0.697}\\
Static sink, $k=3$ & 17.29 & 13.72 & 0.69 & 0.969 & 0.962 & 0.989 & 0.082 & 0.528 & \textbf{0.699}\\
Age-decay read & \underline{18.06} & \underline{7.35} & 1.16 & 0.952 & 0.948 & 0.985 & 0.154 & \underline{0.544} & 0.649\\
ALG-inspired, $k=1$ & 15.87 & 19.96 & 1.07 & 0.977 & \underline{0.969} & \underline{0.990} & 0.249 & 0.537 & 0.663\\
DyMoS port, $k=1$ & 16.89 & 16.95 & 0.97 & \underline{0.979} & \underline{0.969} & \underline{0.990} & 0.228 & 0.539 & 0.671\\
UnStep port, $k=3$ & 16.36 & 16.14 & 0.55 & 0.966 & 0.960 & \textbf{0.991} & 0.065 & 0.537 & 0.692\\
PackCache-style, $k=1$ & 16.48 & 9.23 & 1.21 & 0.963 & 0.956 & 0.986 & 0.249 & \underline{0.544} & 0.663\\
FramePack-style, $k=1$ & 14.39 & 16.17 & 1.97 & \underline{0.979} & \underline{0.969} & 0.989 & 0.394 & \underline{0.544} & 0.684\\
TetherCache port, $k=1$ & 14.38 & 23.47 & 1.32 & 0.978 & \underline{0.969} & \underline{0.990} & 0.276 & 0.519 & 0.669\\
\textbf{SceneTether (ours)} & \textbf{22.37} & \textbf{5.57} & 1.97 & \underline{0.979} & \underline{0.969} & 0.989 & 0.366 & \textbf{0.547} & 0.684\\
\bottomrule\end{tabular}

\end{table*}

\begin{table*}[t]
 \centering\small\setlength{\tabcolsep}{4pt}
 \caption{\textbf{Paired two-seed contrasts and equal-coverage controls at $60\,\text{s}$.} Differences report \Ours minus reference (PSNR, Lab color error) and ratios report \Ours over reference ($40\text{--}60\,\text{s}$ registered Farneb\"ack flow), with descriptive 95\% scene-bootstrap intervals nesting seeds 42 and 1042, except the age-decay row, which uses seed 42 only (42 appearance scenes; parenthetical counts show registration-qualified activity scenes). The bottom two rows match \Ours's source pin, $100\%$ newest-frame coverage, keys, history read, and pass count, differing only in tether strength; they are the isolating controls.}
 \label{tab:st_paired_completed}
 \begin{tabular}{llrrr}\toprule
Reference & Seeds & $\Delta$PSNR (dB) & $\Delta$Color & Flow ratio ($n$)\\\midrule
Echo, $k=1$ & 42, 1042 & $+5.15$ [$+4.24$, $+6.13$] & $-9.24$ [$-12.38$, $-6.33$] & 1.50 [1.27, 1.76] (38)\\
Spatial Echo, $k=3$ & 42, 1042 & $+5.88$ [$+4.90$, $+6.87$] & $-10.20$ [$-12.80$, $-7.73$] & 2.30 [1.85, 2.83] (30)\\
Age-decay read & 42 & $+4.30$ [$+3.34$, $+5.33$] & $-1.78$ [$-2.98$, $-0.55$] & 1.16 [0.98, 1.34] (36)\\
Sparse frozen read & 42, 1042 & $+6.27$ [$+5.31$, $+7.31$] & $-4.66$ [$-7.45$, $-2.31$] & 0.90 [0.80, 0.99] (36)\\
Full coverage, no tether & 42, 1042 & $+8.14$ [$+7.05$, $+9.38$] & $-10.51$ [$-16.92$, $-5.30$] & 0.93 [0.85, 1.03] (36)\\
Half tether & 42, 1042 & $+3.01$ [$+2.51$, $+3.54$] & $+0.03$ [$-1.47$, $+1.20$] & 1.01 [0.97, 1.05] (37)\\
\bottomrule\end{tabular}
\end{table*}

\paragraph{Activity on a common population.}
Restricting evaluation to scenes valid under both optical-flow estimators across both methods and seeds (\cref{tab:st_activity_common}) shows higher late-stage activity for \Ours over framewise Echo (RAFT $2.23\times$, Farneb\"ack $1.50\times$; 38 scenes) and spatial Echo ($4.53\times$, $2.30\times$; 30 scenes).
Against the equal-coverage control, every two-seed interval spans $1.0$ (Farneb\"ack point estimates $0.92\text{--}0.98$; RAFT $0.96$ own, $1.01$ common, $0.94$ source-registered).
Matching Farneb\"ack's temporal window to RAFT ($53\text{--}60\,\text{s}$) shifts $60\,\text{s}$ ratios by at most $0.09$ while dropping one scene against each Echo baseline, so the window boundary alone does not explain the magnitude difference between the estimators, which also differ in resolution, temporal sampling, and motion compensation.
Per seed across the 40 activity scenes, Farneb\"ack inter-frame compensation fails on $1\text{--}3$ rollouts for \Ours, framewise Echo, and the untethered control versus $7\text{--}8$ for spatial Echo (RAFT's self-fitted similarity warp records zero failures), while source registration (39 assessable scenes) fails on $3$ (\Ours), $4\text{--}5$ (framewise Echo), $7$ (untethered control), and $11\text{--}12$ (spatial Echo).

\begin{table*}[t]
\centering
\footnotesize
\setlength{\tabcolsep}{6pt}
\caption{\textbf{Activity ratios across estimator-specific and shared valid populations.} Values report ratios of scene means (\Ours over reference, nesting seeds with 10{,}000 scene-bootstrap draws as in \cref{tab:st_paired_completed}) for MCFF-late RAFT and registered Farneb\"ack flow. At $60\,\text{s}$, RAFT uses $288\text{p}$ at $8\,\text{Hz}$ over $52.9\text{--}60\,\text{s}$, and Farneb\"ack uses $832\times480$ adjacent-frame pairs sampled once per second over $40\text{--}60\,\text{s}$, with a window-matched $53\text{--}60\,\text{s}$ point check; the $120$ and $240\,\text{s}$ rows score the final minute of separately generated rollouts with the same final-window offsets (RAFT $112.9\text{--}120$ and $232.9\text{--}240\,\text{s}$; Farneb\"ack $100\text{--}120$ and $220\text{--}240\,\text{s}$; check $113\text{--}120$ and $233\text{--}240\,\text{s}$). \textit{Own:} each estimator's valid scenes; \textit{Common:} scenes valid for both estimators, methods, and seeds; \textit{+\,src.:} additionally source-registered ($\ge50\%$ coverage for both methods) over the Farneb\"ack window. \textit{Bottom:} per-seed counts of failed rollouts ($<50\%$ valid samples) at $60\,\text{s}$ across 40 activity scenes (39 assessable for source alignment); parentheses give the fraction of all late-window samples that register, pooled over eligible scenes and both seeds (57 RAFT transitions, 20 Farneb\"ack pairs, or 161 source frames per rollout).}
\label{tab:st_activity_common}
\begin{tabular*}{\textwidth}{@{\extracolsep{\fill}}lrllr@{}}
\toprule
Population & $n$ & RAFT & Farn.\ endpoint & Window check \\
\midrule
\multicolumn{5}{@{}l}{\textit{vs.\ Framewise Echo, $k=1$, seeds 42, 1042}} \\
Own & 40/38 & 2.17 [1.58, 3.02] & 1.50 [1.27, 1.76] & -- \\
Common & 38 & 2.23 [1.57, 3.17] & 1.50 [1.27, 1.76] & 1.51 (37) \\
+\,src. & 34 & 2.27 [1.64, 3.22] & 1.48 [1.24, 1.79] & -- \\
\multicolumn{5}{@{}l}{\textit{vs.\ Spatial Echo, $k=3$, seeds 42, 1042}} \\
Own & 40/30 & 3.81 [2.39, 5.89] & 2.30 [1.85, 2.83] & -- \\
Common & 30 & 4.53 [3.30, 6.22] & 2.30 [1.85, 2.83] & 2.39 (29) \\
+\,src. & 25 & 4.39 [3.04, 6.27] & 2.26 [1.78, 2.90] & -- \\
\multicolumn{5}{@{}l}{\textit{vs.\ Full coverage, no tether, seeds 42, 1042}} \\
Own & 40/36 & 0.96 [0.79, 1.24] & 0.93 [0.85, 1.03] & -- \\
Common & 36 & 1.01 [0.79, 1.38] & 0.93 [0.85, 1.03] & 0.98 \\
+\,src. & 31 & 0.94 [0.76, 1.17] & 0.92 [0.83, 1.01] & -- \\
\multicolumn{5}{@{}l}{\textit{vs.\ Framewise Echo, $k=1$, 120\,s, seed 42}} \\
Own & 40/35 & 2.70 [1.82, 3.88] & 1.41 [1.19, 1.74] & -- \\
Common & 35 & 2.51 [1.62, 3.80] & 1.41 [1.19, 1.74] & 1.41 (34) \\
+\,src. & 32 & 2.62 [1.59, 4.11] & 1.48 [1.25, 1.82] & -- \\
\multicolumn{5}{@{}l}{\textit{vs.\ Spatial Echo, $k=3$, 120\,s, seed 42}} \\
Own & 40/26 & 4.56 [2.91, 6.95] & 2.38 [1.83, 3.00] & -- \\
Common & 26 & 4.79 [2.65, 8.42] & 2.38 [1.83, 3.00] & 2.36 (24) \\
+\,src. & 20 & 5.35 [2.76, 9.40] & 2.17 [1.62, 2.87] & -- \\
\multicolumn{5}{@{}l}{\textit{vs.\ Spatial Echo, $k=3$, 240\,s, seed 42}} \\
Own & 40/16 & 5.45 [3.68, 7.85] & 2.36 [1.77, 3.00] & -- \\
Common & 16 & 6.05 [3.46, 10.01] & 2.36 [1.77, 3.00] & 2.56 \\
+\,src. & 12 & 4.91 [2.79, 8.07] & 2.13 [1.48, 3.19] & -- \\
\midrule
\multicolumn{5}{@{}l}{\textit{Failures at 60\,s, seed 42 / 1042}} \\
\multicolumn{2}{@{}l}{Method} & RAFT inter-fr. & Farn.\ inter-fr. & Source \\
\multicolumn{2}{@{}l}{\Ours} & 0 / 0 (1.00) & 2 / 1 (0.95) & 3 / 3 (0.92) \\
\multicolumn{2}{@{}l}{Framewise Echo, $k=1$} & 0 / 0 (1.00) & 2 / 1 (0.95) & 5 / 4 (0.86) \\
\multicolumn{2}{@{}l}{Spatial Echo, $k=3$} & 0 / 0 (1.00) & 7 / 8 (0.83) & 11 / 12 (0.72) \\
\multicolumn{2}{@{}l}{Full coverage, no tether} & 0 / 0 (1.00) & 1 / 3 (0.93) & 7 / 7 (0.82) \\
\bottomrule
\end{tabular*}
\end{table*}

\begin{table*}[t]
 \centering\scriptsize\setlength{\tabcolsep}{2pt}
 \caption{\textbf{Source-copying sanity diagnostics} (48 sources, seed 42, $60\,\text{s}$). Synthetic source-repeat and rigid-support paste-back controls show that source fidelity can improve through copying without establishing plausible material dynamics, motivating separate reports of fidelity, activity, and perceptual quality (`--' marks uncomputed entries).}
 \label{tab:st_copying}
 \begin{tabular}{@{}>{\raggedright\arraybackslash}p{.225\textwidth}*{9}{>{\centering\arraybackslash}p{.078\textwidth}}@{}}\toprule
 & \multicolumn{3}{c}{SNF-Bench} & \multicolumn{2}{c}{VBench-I2V} & \multicolumn{4}{c}{VBench-Long}\\
\cmidrule(lr){2-4}\cmidrule(lr){5-6}\cmidrule(lr){7-10}
Method & \shortstack{Grayscale\\peak signal-\\to-noise ratio\\(dB) $\uparrow$} & \shortstack{Mean-Lab\\color error\\$\downarrow$} & \shortstack{Late material\\activity\\(pixels/pair)} & \shortstack{Subject\\consistency\\$\uparrow$} & \shortstack{Background\\consistency\\$\uparrow$} & \shortstack{Motion\\smoothness\\$\uparrow$} & \shortstack{Dynamic\\degree} & \shortstack{Aesthetic\\quality\\$\uparrow$} & \shortstack{Imaging\\quality\\$\uparrow$}\\\midrule
Repeated source & \textbf{41.65} & \textbf{2.53} & 0.04 & \textbf{0.996} & \textbf{0.983} & \textbf{0.998} & 0.000 & \textbf{0.632} & \textbf{0.696}\\
Source pasteback & \underline{28.30} & \underline{3.37} & 0.63 & \underline{0.983} & \underline{0.973} & \underline{0.993} & 0.134 & 0.539 & \underline{0.694}\\
Framewise Echo & 16.97 & 15.54 & 1.02 & 0.981 & 0.971 & 0.990 & 0.219 & 0.538 & 0.674\\
\textbf{SceneTether (ours)} & 22.37 & 5.57 & 1.97 & 0.979 & 0.969 & 0.989 & 0.366 & \underline{0.547} & 0.684\\
\bottomrule\end{tabular}

\end{table*}

\paragraph{Output composites.}
To test whether post-hoc pixel blending can reproduce the gains, we construct two output composites without evaluation masks, using Gaussian feathering with $\sigma=8\,\text{px}$ (\cref{eq:st_composite}).
The primary composite blends the source photo into the untethered equal-coverage rollout using that run's own frozen warm-up map---which is bitwise identical to \Ours's map across all 48 sources because tethering begins only after warm-up.
A secondary composite blends the FramePack-style rollout using an activity map estimated from its first decoded frames, varying both memory policy and map estimation.
At $60\,\text{s}$, the primary composite matches \Ours closely on summary source metrics (\cref{tab:st_composites}): $\Delta\text{PSNR}=\stres{comp_psnr}\,\text{dB}$ \stres{comp_psnr_ci} ($n=\stres{comp_psnr_n}$) and flow ratio \stres{comp_flow} \stres{comp_flow_ci} ($n=\stres{comp_flow_n}$); both intervals include the null, so static-support PSNR and flow amplitude alone do not separate in-attention anchoring from pixel pasting.
The approaches differ beyond these static scores: compositing reduces variation within strongly anchored regions (\stres{comp_still_flicker} vs.\ \stres{st_still_flicker} gray levels), attenuates mean-frame gradients in the transition band, and retains the untethered rollout's color drift in moving regions (\cref{tab:extra_composite_drift}). Reduced variation is not necessarily a perceptual failure, so we additionally compare encoded videos through blinded full-minute human judgments (\cref{tab:st_human_full_duration}).

\paragraph{Composite construction.}
Given an untethered decoded 8-bit rollout $G$ and its $30\times52$ token schedule $\beta$ (Eq.~(4) of the main paper), we upsample $\beta$ to $480\times832$ via $16\times16$ nearest-neighbor replication and smooth it with a reflected-border Gaussian kernel $\mathcal{G}_{\sigma=8}$ ($65\times65$):
\begin{equation}
 \begin{split}
 \alpha&=\mathcal{G}_{\sigma=8}*\operatorname{NN}_{\uparrow16}(\beta),\\
 \widehat I_n&=\begin{cases}
 G_n, & n\le24,\\
 \operatorname{clip}\big\lfloor G_n+\alpha\odot(S-G_n)+\tfrac12\big\rfloor, & n\ge25,
 \end{cases}
 \end{split}
 \label{eq:st_composite}
\end{equation}
where $S$ denotes the bilinearly resized $480\times832$ source image, clipping bounds intensities to $[0,255]$, and $n\in\{0,\dots,960\}$ indexes output frames (leaving warm-up frames $0\text{--}24$ unblended to match latent~7 onset; mean $\alpha=0.449$).
We encode $\widehat I$ via H.264 (\texttt{libx264}, CRF 18, preset \texttt{medium}, \texttt{yuv420p}, $16\,\text{fps}$, ${\sim}1.8\,\text{Mb/s}$), introducing one extra encoding pass over native generator outputs (${\sim}1.0\,\text{Mb/s}$), which we control for via lossless regeneration in Sec.~\ref{fm:suppl_human}.
For the FramePack-style composite, we area-downsample frames $0\text{--}24$ by $8\times$, pool channel-averaged temporal RGB variance onto $30\times52$, compute $\log_{10}(\cdot+10^{-6})$, select Otsu's $\tau$ over 197 quantiles ($1\text{st--}99\text{th}$ percentiles), and apply the same taper without a degenerate fallback (assigning $0$ rather than $0.5$ to exterior tokens with $L=\tau$).

\begin{table*}[t]
 \centering\scriptsize\setlength{\tabcolsep}{3.5pt}
 \caption{\textbf{Post-hoc output composites vs.\ \Ours at $60\,\text{s}$} (seed 42, 48 sources). $\Delta\text{PSNR}$ (row minus \Ours) and $40\text{--}60\,\text{s}$ flow ratios (row over \Ours) include 95\% scene-bootstrap intervals. Region columns evaluate $40\text{--}60\,\text{s}$ under \Ours's frozen map: ``flicker'' is the $1\,\text{s}$-window grayscale standard deviation, not an artifact count, and edge loss measures the mean-frame gradient drop below the source (8-bit units). The map-transfer row is diagnostic, not a standalone baseline.}
 \label{tab:st_composites}
 \begin{tabular}{@{}lccccccc@{}}\toprule
 & \multicolumn{4}{c}{Source endpoints} & \multicolumn{3}{c}{Regions of the \Ours map}\\\cmidrule(lr){2-5}\cmidrule(lr){6-8}
Method & PSNR $\uparrow$ & $\Delta$PSNR & Color $\downarrow$ & Act.\ ratio & \shortstack{Band\\flicker} & \shortstack{Still\\flicker} & \shortstack{Band\\edge loss}\\\midrule
\OursLabel & 22.37 & -- & 5.57 & -- & 5.95 & 3.79 & 17.1\\
Full coverage, no tether & 14.32 & -8.05 [-9.34, -6.87] & 15.66 & 1.14 [1.01, 1.30] & 10.00 & 8.03 & 24.2\\
Composite on untethered & 22.57 & +0.20 [-1.15, 1.54] & 5.46 & 1.05 [0.92, 1.18] & 5.13 & 0.23 & 20.0\\
Composite on FramePack-style port & 20.01 & -2.36 [-3.43, -1.26] & 7.02 & 0.89 [0.78, 1.02] & 6.35 & 2.58 & 23.2\\
\quad with \Ours map (diagnostic) & -- & -- & -- & -- & 5.21 & 0.23 & 19.8\\
\bottomrule\end{tabular}

\end{table*}

\begin{table}[t]
 \centering\scriptsize\setlength{\tabcolsep}{5pt}
 \caption{\textbf{VBench consistency dimensions at $60\,\text{s}$} across all 48 benchmark scenes (seed 42; `--' marks uncomputed suites), separating input-referenced VBench-I2V from sequence-level VBench-Long consistency.}
 \label{tab:st_vbench_consistency}
 \begin{tabular}{lrrrr}\toprule
Method & I2V subj. & I2V bg. & Long subj. & Long bg.\\\midrule
CF++ + Echo & \textbf{0.981} & \textbf{0.971} & 0.984 & \underline{0.968}\\
Spatial Echo, k=3 & 0.970 & 0.961 & \underline{0.985} & 0.967\\
Static sink, k=3 & 0.969 & 0.962 & 0.984 & 0.966\\
Age-decay read & 0.952 & 0.948 & 0.977 & 0.959\\
DyMoS port, k=1 & \underline{0.979} & \underline{0.969} & 0.983 & 0.967\\
ALG-inspired, k=1 & 0.977 & \underline{0.969} & 0.982 & 0.965\\
UnStep port, k=3 & 0.966 & 0.960 & 0.981 & 0.962\\
PackCache-style & 0.963 & 0.956 & 0.980 & 0.962\\
FramePack-style & \underline{0.979} & \underline{0.969} & 0.982 & 0.965\\
Full coverage, no tether & 0.978 & \underline{0.969} & 0.982 & 0.964\\
Frozen gated read & 0.971 & 0.963 & 0.979 & 0.958\\
\textbf{SceneTether (ours)} & \underline{0.979} & \underline{0.969} & \textbf{0.987} & \textbf{0.970}\\
\bottomrule\end{tabular}

\end{table}

\begin{table}[t]
 \centering\scriptsize\setlength{\tabcolsep}{5pt}
 \caption{\textbf{VBench-Long visual and temporal quality dimensions at $60\,\text{s}$} (same 48 rollouts as \cref{tab:st_vbench_consistency}). High smoothness or dynamic degree can coexist with frozen backgrounds or visual artifacts.}
 \label{tab:st_vbench_quality}
 \begin{tabular}{lrrrr}\toprule
Method & Smooth. & Dynamic & Aesthetic & Imaging\\\midrule
CF++ + Echo & \underline{0.990} & 0.219 & 0.538 & 0.674\\
Spatial Echo, k=3 & \textbf{0.991} & 0.050 & 0.539 & \underline{0.697}\\
Static sink, k=3 & 0.989 & 0.082 & 0.528 & \textbf{0.699}\\
Age-decay read & 0.985 & 0.154 & \underline{0.544} & 0.649\\
DyMoS port, k=1 & \underline{0.990} & 0.228 & 0.539 & 0.671\\
ALG-inspired, k=1 & \underline{0.990} & 0.249 & 0.537 & 0.663\\
UnStep port, k=3 & \textbf{0.991} & 0.065 & 0.537 & 0.692\\
PackCache-style & 0.986 & 0.249 & \underline{0.544} & 0.663\\
FramePack-style & 0.989 & 0.394 & \underline{0.544} & 0.684\\
Full coverage, no tether & 0.988 & 0.372 & 0.542 & 0.682\\
Frozen gated read & 0.987 & 0.412 & 0.537 & 0.664\\
\textbf{SceneTether (ours)} & 0.989 & 0.366 & \textbf{0.547} & 0.684\\
\bottomrule\end{tabular}

\end{table}

\begin{table}[t]
 \centering\scriptsize\setlength{\tabcolsep}{4pt}
 \caption{\textbf{Temporal flicker dimensions and valid scene counts} (out of 48 at $60\,\text{s}$, seed 42; `--' uncomputed). Because scene filtering alters the evaluated subset, filtered and unfiltered means are not directly comparable.}
 \label{tab:st_vbench_flicker}
 \begin{tabular}{lrr}\toprule
Method & Filtered (coverage) & Unfiltered (coverage)\\\midrule
CF++ + Echo & 0.989 (37/48) & \textbf{0.986} (48/48)\\
Spatial Echo, k=3 & 0.987 (46/48) & \textbf{0.986} (48/48)\\
Static sink, k=3 & 0.987 (46/48) & \textbf{0.986} (48/48)\\
Age-decay read & 0.981 (35/48) & 0.978 (48/48)\\
DyMoS port, k=1 & 0.989 (36/48) & \textbf{0.986} (48/48)\\
ALG-inspired, k=1 & 0.988 (38/48) & \underline{0.985} (48/48)\\
UnStep port, k=3 & 0.986 (45/48) & \underline{0.985} (48/48)\\
PackCache-style & 0.983 (35/48) & 0.979 (48/48)\\
FramePack-style & 0.986 (26/48) & 0.980 (48/48)\\
Full coverage, no tether & 0.985 (27/48) & 0.979 (48/48)\\
Frozen gated read & 0.983 (28/48) & 0.977 (48/48)\\
\textbf{SceneTether (ours)} & 0.988 (30/48) & 0.983 (48/48)\\
\bottomrule\end{tabular}

\end{table}

\begin{figure*}[t]
 \centering\includegraphics[width=\textwidth]{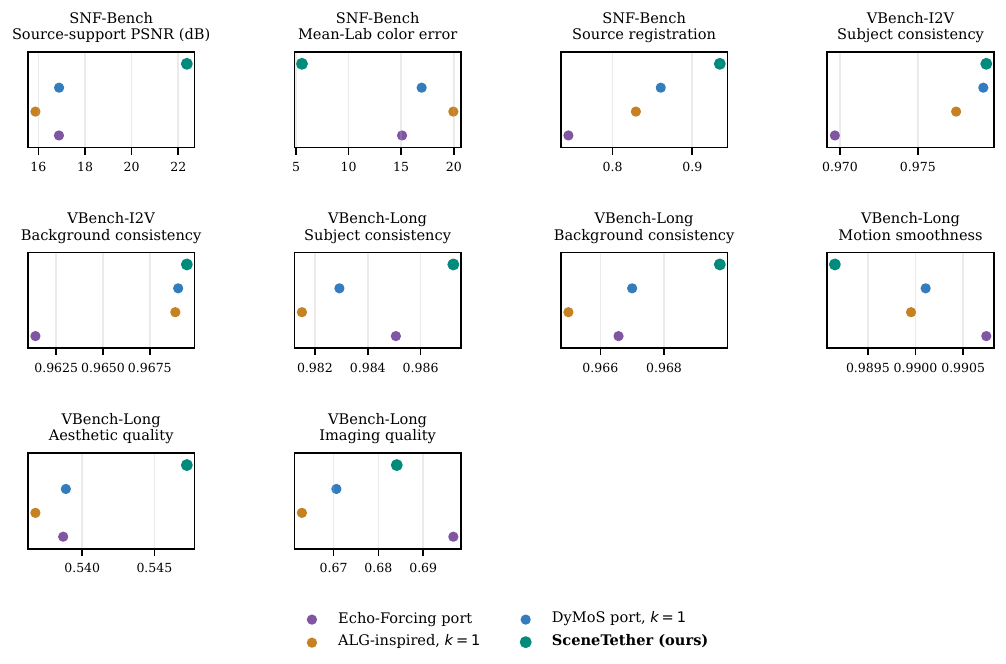}
 \caption{\textbf{Metric profiles across $60\,\text{s}$ benchmark outputs} (\cref{tab:st_full60}). Directional metrics are shown on their respective native scales; non-directional material activity and dynamic degree are omitted.}
 \label{fig:st_suite_profiles}
\end{figure*}

\section{Two- and four-minute comparisons}
\label{st:suppl_long}
\Cref{tab:st_long120,tab:st_long240} report $120\,\text{s}$ and $240\,\text{s}$ benchmark results, using the prescribed metric-specific late-window sampling within the final minute of each separately generated rollout. Source-referenced metrics always use the original input image; uncomputed entries (`--') are not imputed from shorter generations.
Highlighting ranks available directional scores strictly within each horizon's measured cohort.

\begin{table*}[t]
 \centering\scriptsize\setlength{\tabcolsep}{2pt}
 \caption{\textbf{$120\,\text{s}$ benchmark comparison} from separately generated two-minute rollouts on the frozen CF++ student (42 appearance-, 40 activity-, and 48 VBench-eligible scenes). Scores follow their metric-specific late-window definitions within the final minute; `--' denotes uncomputed entries, and adapted rows represent causal ports rather than official native pipelines.}
 \label{tab:st_long120}
 \begin{tabular}{@{}>{\raggedright\arraybackslash}p{.225\textwidth}*{9}{>{\centering\arraybackslash}p{.078\textwidth}}@{}}\toprule
 & \multicolumn{3}{c}{SNF-Bench} & \multicolumn{2}{c}{VBench-I2V} & \multicolumn{4}{c}{VBench-Long}\\
\cmidrule(lr){2-4}\cmidrule(lr){5-6}\cmidrule(lr){7-10}
Method & \shortstack{Grayscale\\peak signal-\\to-noise ratio\\(dB) $\uparrow$} & \shortstack{Mean-Lab\\color error\\$\downarrow$} & \shortstack{Late material\\activity\\(pixels/pair)} & \shortstack{Subject\\consistency\\$\uparrow$} & \shortstack{Background\\consistency\\$\uparrow$} & \shortstack{Motion\\smoothness\\$\uparrow$} & \shortstack{Dynamic\\degree} & \shortstack{Aesthetic\\quality\\$\uparrow$} & \shortstack{Imaging\\quality\\$\uparrow$}\\\midrule
Framewise Echo, $k=1$ & 17.11 & 15.31 & 0.77 & 0.914 & 0.916 & \underline{0.990} & 0.218 & 0.523 & 0.657\\
Spatial Echo, $k=3$ & 16.37 & 17.52 & 0.46 & 0.890 & 0.898 & \textbf{0.991} & 0.042 & 0.523 & 0.668\\
Static sink, $k=3$ & 16.91 & 14.33 & 0.73 & 0.883 & 0.893 & 0.988 & 0.120 & 0.494 & \underline{0.689}\\
Age-decay read & \underline{18.12} & \underline{7.01} & 1.11 & 0.922 & 0.925 & 0.984 & 0.167 & \underline{0.543} & 0.649\\
ALG-inspired, $k=1$ & 15.69 & 20.69 & 0.97 & 0.894 & 0.894 & 0.988 & 0.234 & 0.510 & 0.636\\
DyMoS port, $k=1$ & 16.85 & 16.40 & 0.82 & 0.911 & 0.909 & 0.989 & 0.182 & 0.527 & 0.654\\
UnStep port, $k=3$ & 16.38 & 15.42 & 0.63 & 0.899 & 0.904 & \underline{0.990} & 0.058 & 0.535 & \textbf{0.691}\\
PackCache-style, $k=1$ & 16.56 & 9.41 & 1.21 & \underline{0.934} & \underline{0.933} & 0.986 & 0.250 & \textbf{0.544} & 0.662\\
FramePack-style, $k=1$ & 14.52 & 15.67 & 2.06 & 0.927 & 0.930 & 0.989 & 0.395 & 0.530 & 0.685\\
TetherCache port, $k=1$ & 14.04 & 26.50 & 1.36 & 0.859 & 0.875 & \underline{0.990} & 0.228 & 0.466 & 0.646\\
\textbf{SceneTether (ours)} & \textbf{22.46} & \textbf{5.53} & 2.09 & \textbf{0.954} & \textbf{0.947} & 0.989 & 0.360 & \textbf{0.544} & 0.685\\
\bottomrule\end{tabular}

\end{table*}
\begin{table*}[t]
 \centering\scriptsize\setlength{\tabcolsep}{2pt}
 \caption{\textbf{$240\,\text{s}$ benchmark comparison} from separately generated four-minute rollouts (metric-specific late-window definitions as in \cref{tab:st_long120}). The full-benchmark analysis is secondary to the preselected 12-source primary endpoint (\cref{tab:st_h240_primary}).}
 \label{tab:st_long240}
 \begin{tabular}{@{}>{\raggedright\arraybackslash}p{.225\textwidth}*{9}{>{\centering\arraybackslash}p{.078\textwidth}}@{}}\toprule
 & \multicolumn{3}{c}{SNF-Bench} & \multicolumn{2}{c}{VBench-I2V} & \multicolumn{4}{c}{VBench-Long}\\
\cmidrule(lr){2-4}\cmidrule(lr){5-6}\cmidrule(lr){7-10}
Method & \shortstack{Grayscale\\peak signal-\\to-noise ratio\\(dB) $\uparrow$} & \shortstack{Mean-Lab\\color error\\$\downarrow$} & \shortstack{Late material\\activity\\(pixels/pair)} & \shortstack{Subject\\consistency\\$\uparrow$} & \shortstack{Background\\consistency\\$\uparrow$} & \shortstack{Motion\\smoothness\\$\uparrow$} & \shortstack{Dynamic\\degree} & \shortstack{Aesthetic\\quality\\$\uparrow$} & \shortstack{Imaging\\quality\\$\uparrow$}\\\midrule
Framewise Echo, $k=1$ & 17.48 & 14.98 & 0.84 & 0.892 & 0.900 & 0.987 & 0.200 & 0.511 & 0.639\\
Spatial Echo, $k=3$ & 15.85 & 19.71 & 0.35 & 0.831 & 0.852 & \textbf{0.990} & 0.042 & 0.497 & 0.618\\
Static sink, $k=3$ & 17.07 & 15.07 & 0.76 & 0.860 & 0.876 & 0.986 & 0.238 & 0.486 & \textbf{0.693}\\
Age-decay read & \underline{18.22} & \underline{6.99} & 1.03 & 0.926 & 0.926 & 0.985 & 0.165 & \underline{0.544} & 0.649\\
ALG-inspired, $k=1$ & 15.71 & 20.17 & 1.62 & 0.862 & 0.871 & 0.985 & 0.253 & 0.492 & 0.633\\
DyMoS port, $k=1$ & 16.58 & 17.84 & 1.15 & 0.888 & 0.892 & 0.987 & 0.213 & 0.516 & 0.639\\
UnStep port, $k=3$ & 16.47 & 16.03 & 0.60 & 0.895 & 0.901 & \textbf{0.990} & 0.072 & 0.532 & \underline{0.692}\\
PackCache-style, $k=1$ & 16.41 & 9.46 & 1.22 & \underline{0.933} & \underline{0.933} & 0.986 & 0.247 & 0.543 & 0.661\\
FramePack-style, $k=1$ & 14.24 & 18.77 & 2.19 & 0.918 & 0.925 & 0.988 & 0.386 & 0.525 & 0.689\\
TetherCache port, $k=1$ & 13.35 & 29.35 & 1.05 & 0.794 & 0.818 & \textbf{0.990} & 0.251 & 0.428 & 0.624\\
\textbf{SceneTether (ours)} & \textbf{22.47} & \textbf{5.69} & 1.91 & \textbf{0.953} & \textbf{0.948} & \underline{0.989} & 0.344 & \textbf{0.545} & 0.684\\
\bottomrule\end{tabular}

\end{table*}
\begin{table*}[t]
 \centering\scriptsize\setlength{\tabcolsep}{2pt}
 \caption{\textbf{Four-minute primary subset and supplementary references} (metric-specific late-window endpoints of $240\,\text{s}$ rollouts, seed 42). We locked the 12-source primary subset (10 appearance-eligible) on October 14 prior to $240\,\text{s}$ generation, designating spatial Echo-Forcing as the primary contrast and reporting all-48 means (42/40 eligible) and additional baselines as secondary context (paired differences report \Ours minus reference with 95\% scene-bootstrap intervals).}
 \label{tab:st_h240_primary}
 
\begin{tabular*}{\textwidth}{@{\extracolsep{\fill}}llrrrrrr@{}}
\toprule
Population & Method & PSNR $\uparrow$ & Color $\downarrow$ & Activity & $\Delta$PSNR [95\% CI] & $\Delta$Color [95\% CI] & $\Delta$Activity [95\% CI] \\
\midrule
Primary, frozen 12 & Spatial Echo, $k=3$ & 16.87 & 18.12 & 0.35 & +7.31 [+5.42, +9.25] & -13.09 [-17.50, -8.81] & +1.02 [+0.21, +1.99] \\
 & Framewise Echo, $k=1$ & 17.24 & 18.17 & 1.10 & +6.93 [+5.11, +8.75] & -13.14 [-22.28, -5.60] & +0.27 [-0.70, +1.45] \\
 & Static sink, $k=3$ & 17.47 & 15.12 & 0.48 & +6.71 [+5.22, +8.17] & -10.09 [-14.72, -5.86] & +0.89 [+0.05, +1.93] \\
 & Age-decay read & \underline{19.13} & \underline{7.71} & 0.93 & +5.05 [+3.41, +6.64] & -2.69 [-5.94, +0.72] & +0.43 [-0.01, +0.98] \\
 & ALG-inspired, $k=1$ & 16.58 & 24.42 & 2.66 & +7.60 [+5.92, +9.07] & -19.40 [-30.13, -10.71] & -1.30 [-4.35, +0.68] \\
 & DyMoS port, $k=1$ & 16.97 & 22.31 & 0.87 & +7.20 [+5.09, +9.22] & -17.28 [-27.89, -9.78] & +0.50 [-0.31, +1.43] \\
 & UnStep port, $k=3$ & 17.41 & 14.26 & 0.50 & +6.77 [+4.99, +8.62] & -9.23 [-14.46, -4.86] & +0.87 [+0.08, +1.86] \\
 & PackCache-style, $k=1$ & 17.09 & 12.83 & 0.73 & +7.09 [+5.30, +8.88] & -7.80 [-14.25, -2.93] & +0.64 [+0.06, +1.31] \\
 & FramePack-style, $k=1$ & 15.28 & 24.46 & 2.16 & +8.90 [+6.52, +11.39] & -19.44 [-36.57, -4.66] & -0.79 [-1.53, -0.10] \\
 & TetherCache port, $k=1$ & 13.51 & 33.32 & 1.15 & +10.67 [+8.80, +12.54] & -28.29 [-40.39, -17.64] & +0.22 [-0.73, +1.11] \\
 & \textbf{SceneTether (ours)} & \textbf{24.18} & \textbf{5.03} & 1.37 & -- & -- & -- \\
\midrule
Secondary, all 48 & Spatial Echo, $k=3$ & 15.85 & 19.71 & 0.35 & +6.62 [+5.66, +7.60] & -14.02 [-17.28, -11.06] & +1.56 [+0.92, +2.29] \\
 & Framewise Echo, $k=1$ & 17.48 & 14.98 & 0.84 & +4.99 [+4.05, +5.92] & -9.28 [-12.52, -6.30] & +1.07 [+0.36, +1.86] \\
 & Static sink, $k=3$ & 17.07 & 15.07 & 0.76 & +5.40 [+4.31, +6.62] & -9.38 [-13.52, -5.98] & +1.14 [+0.47, +1.91] \\
 & Age-decay read & \underline{18.22} & \underline{6.99} & 1.03 & +4.25 [+3.41, +5.17] & -1.30 [-2.58, -0.07] & +0.88 [+0.23, +1.59] \\
 & ALG-inspired, $k=1$ & 15.71 & 20.17 & 1.62 & +6.76 [+5.58, +8.07] & -14.48 [-21.19, -9.22] & +0.29 [-0.76, +1.24] \\
 & DyMoS port, $k=1$ & 16.58 & 17.84 & 1.15 & +5.89 [+4.80, +7.05] & -12.15 [-16.10, -8.58] & +0.76 [+0.04, +1.54] \\
 & UnStep port, $k=3$ & 16.47 & 16.03 & 0.60 & +6.00 [+4.96, +7.06] & -10.34 [-12.75, -7.97] & +1.31 [+0.65, +2.03] \\
 & PackCache-style, $k=1$ & 16.41 & 9.46 & 1.22 & +6.06 [+5.19, +6.94] & -3.77 [-6.00, -1.85] & +0.69 [+0.21, +1.26] \\
 & FramePack-style, $k=1$ & 14.24 & 18.77 & 2.19 & +8.22 [+7.00, +9.60] & -13.08 [-21.24, -6.48] & -0.28 [-0.98, +0.39] \\
 & TetherCache port, $k=1$ & 13.35 & 29.35 & 1.05 & +9.11 [+7.87, +10.39] & -23.66 [-30.46, -17.42] & +0.86 [+0.27, +1.54] \\
 & \textbf{SceneTether (ours)} & \textbf{22.47} & \textbf{5.69} & 1.91 & -- & -- & -- \\
\bottomrule
\end{tabular*}

\end{table*}
\begin{figure*}[t]
 \centering\includegraphics[width=.98\textwidth]{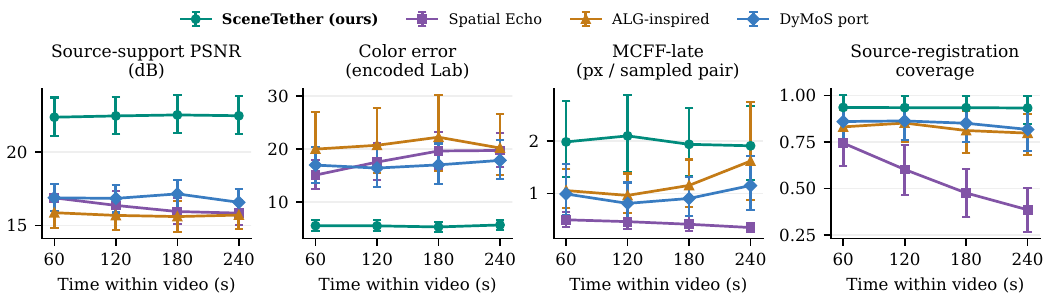}
 \caption{\textbf{Within-rollout trajectories across four consecutive one-minute windows} of continuous $240\,\text{s}$ generations (seed 42; means and 95\% scene-bootstrap intervals on shared eligible scenes; see \cref{tab:st_long240,tab:st_long_registration}). The plot shows the four methods in its legend; \cref{tab:st_long_registration} additionally reports the framewise Echo host and age-decay read, which lose far less registration coverage than the spatial Echo-Forcing $k=3$ port, so that port's decline is not inherent to this CF++ checkpoint.}
 \label{fig:st_within240}
\end{figure*}
\begin{table}[t]
 \centering\small\setlength{\tabcolsep}{4pt}
 \caption{\textbf{Within-rollout source registration through four minutes} across all 45 geometry-eligible scenes (including failures) for the $240\,\text{s}$ videos in \cref{fig:st_within240}.}
 \label{tab:st_long_registration}
 \begin{tabular}{@{}lrrrr@{}}\toprule
Method & 60\,s & 120\,s & 180\,s & 240\,s\\\midrule
Framewise Echo, $k=1$ & 87.8\% & 84.2\% & 83.0\% & 82.5\%\\
Spatial Echo, $k=3$ & 74.4\% & 60.2\% & 47.6\% & 38.5\%\\
Static sink, $k=3$ & 85.1\% & 82.6\% & 81.3\% & 84.0\%\\
Age-decay read & \underline{92.0}\% & \underline{92.9}\% & 92.4\% & \underline{92.7}\%\\
ALG-inspired, $k=1$ & 83.1\% & 85.1\% & 81.1\% & 79.7\%\\
DyMoS port, $k=1$ & 86.0\% & 86.3\% & 85.0\% & 81.7\%\\
UnStep port, $k=3$ & 80.4\% & 77.9\% & 79.6\% & 79.5\%\\
PackCache-style, $k=1$ & 91.3\% & 91.6\% & \underline{92.9}\% & \underline{92.7}\%\\
FramePack-style, $k=1$ & 84.6\% & 84.8\% & 85.0\% & 85.1\%\\
TetherCache port, $k=1$ & 79.8\% & 76.8\% & 74.0\% & 74.2\%\\
\textbf{SceneTether (ours)} & \textbf{93.5}\% & \textbf{93.4}\% & \textbf{93.4}\% & \textbf{93.2}\%\\
\bottomrule\end{tabular}

\end{table}

\section{Fresh-source confirmation}
\label{st:suppl_fresh}
To test out-of-split generalization on scenes unseen during method development, we froze 24 fresh sources on October 7, 2026, generated $60\,\text{s}$ rollouts across two nested seeds (42 and 1042), and evaluated them with automatic masks that received AI-only visual review (\cref{tab:st_fresh}).
Against spatial Echo-Forcing, the paired intervals favor \Ours in source PSNR, mean-Lab color error, and measured activity. Relative to the sparse frozen read, \Ours improves source fidelity and color stability but has lower flow amplitude; relative to age decay, it improves PSNR with unresolved color and activity differences. The collection yields 21 appearance-eligible sources, and each registered-flow comparison uses its own valid subset. AI visual review does not make these masks independent human annotations.

\begin{table*}[t]
 \centering\small\setlength{\tabcolsep}{4pt}
 \caption{\textbf{Confirmation on 24 sources not used in method development.} Differences are \Ours minus reference, with seeds 42 and 1042 nested in 95\% scene-bootstrap intervals. There are 21 appearance-eligible sources after three precipitation exclusions; registered-flow ratios use their own valid paired populations. Masks received AI-only visual review.}
 \label{tab:st_fresh}
 \begin{tabular}{@{}llrrr@{}}\toprule
Reference & Seeds & $\Delta$PSNR (dB) & $\Delta$Color & Flow ratio ($n$)\\\midrule
Spatial Echo, $k=3$ & 42, 1042 & $+4.40$ [$+2.72$, $+5.90$] & $-11.28$ [$-17.07$, $-5.84$] & 1.44 [1.13, 1.83] (12)\\
Sparse frozen read & 42, 1042 & $+5.03$ [$+3.92$, $+6.18$] & $-4.27$ [$-7.33$, $-1.83$] & 0.84 [0.69, 0.94] (14)\\
Age-decay read & 42, 1042 & $+3.06$ [$+1.54$, $+4.53$] & $+0.15$ [$-2.88$, $+3.89$] & 1.03 [0.88, 1.15] (13)\\
\bottomrule\end{tabular}
\end{table*}

\section{Human evaluation}
\label{fm:suppl_human}

\begin{figure*}[t]
 \centering
 \raisebox{-113.73082pt}{\includegraphics{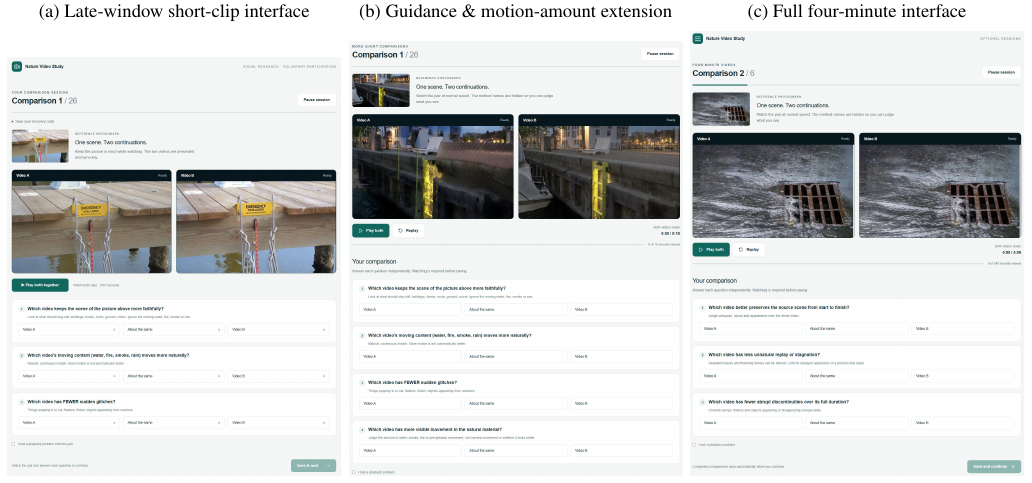}}
 \caption{\textbf{Blinded participant interfaces} shown at native aspect ratios with the source image, synchronized anonymized A/B video players, criterion-specific prompts, and explicit tie options. Panel~(b) adds an activity-amount question alongside motion naturalness, while Panel~(c) evaluates scene preservation, replay/stagnation, and discontinuities across full $240\,\text{s}$ videos.}
 \label{fig:st_study_interfaces}
\end{figure*}

\paragraph{Experimental setup.}
Blinded raters compared anonymized left/right video pairs alongside the original source photograph through a web interface that synchronized HTML5 playback and paused both players upon buffering or clock skew (\cref{fig:st_study_interfaces}).
Short-clip and one-minute stimuli come from a pool of 24 category-stratified benchmark scenes drawn with a fixed seed without consulting automatic scores; each module uses the subset reported in its table. The four-minute module uses the 12 preselected horizon scenes.
Ties receive half-credit ($50\%$ neutral), and $95\%$ confidence intervals resample source scenes and raters independently via a crossed bootstrap (\cref{tab:st_human_progress}); intervals are descriptive and unadjusted for multiplicity across pairs and criteria.

\paragraph{Study modules.}
In the $50\text{--}60\,\text{s}$ modules (\cref{tab:st_human_all}), viewers evaluated the final $10\,\text{s}$ of each rollout (encoded identically across methods) on scene preservation, motion naturalness, and fewer transient artifacts across memory baselines, guidance/coverage controls, and baseline-only pairs that check whether raters distinguish baselines without \Ours.
The one-minute modules show complete $60\,\text{s}$ rollouts and ask about scene preservation, motion naturalness, artifacts, and replay/stagnation, weighting observed sources equally after averaging within-source votes (\cref{tab:st_human_full_duration}). The four-minute module asks about scene preservation, replay/stagnation, and discontinuities, not overall motion naturalness (\cref{tab:st_human_all}). We do not pool questions or exposure durations across modules.
Each comparison trial yields one rating per question. Module totals in \cref{tab:st_human_progress} count comparison trials; for example, the 72 matched-encoding trials comprise 36 composite and 36 untethered-control comparisons. Because a participant can join several modules, module rater counts should not be summed.
Because the initial post-hoc pixel composite passed through one extra H.264 encode (CRF~18, $16\,\text{fps}$), a dedicated matched-encoding module repeated the $60\,\text{s}$ composite and untethered comparisons from lossless renders under identical single-pass encoding.

\paragraph{Participants.}
Participants were graduate researchers in AI, machine learning, and computer vision, and were therefore familiar with video generation and its typical artifacts.
Each participant used an anonymous rater code; we collected no names or contact details.
Before each module, participants read the task description, the expected duration, and the data to be logged (choices, optional notes, watch time), and gave explicit electronic consent.
Participation was voluntary and unpaid, could be stopped at any time, and showed only natural-scene photographs and animations.
We did not seek institutional review board (IRB) review or a formal exemption determination for this minimal-risk perceptual study; we followed the principles of informed consent, voluntary participation, anonymity, and data minimization.
We treat these results as supporting perceptual evidence, not as an independent user study.

\paragraph{Responses analyzed.}
We retain every human method-comparison response, without rater or response exclusions. All tables in this section and Tab.~3 of the main paper report this single analysis.
Identical-video attention-check trials, which compare a video with itself, and our one automated interface test are not method comparisons and therefore enter no table.

\begin{table}[t]
 \centering\scriptsize\setlength{\tabcolsep}{5pt}
 \caption{\textbf{Human-study participation by module.} Some participants completed multiple modules, so rater counts are module-specific. ``Comparisons'' counts comparison trials, not per-question responses.}
 \label{tab:st_human_progress}
 \begin{tabular}{@{}lrrr@{}}\toprule
Module & Raters & Sources & Comparisons\\\midrule
Baseline variants ($50\text{--}60\,\text{s}$) & 14 & 24 & 321\\
Guidance/coverage ($50\text{--}60\,\text{s}$) & 7 & 12 & 84\\
Four-minute ($0\text{--}240\,\text{s}$) & 8 & 12 & 32\\
Baseline validation ($50\text{--}60\,\text{s}$) & 7 & 16 & 168\\
One-minute references ($0\text{--}60\,\text{s}$) & 11 & 24 & 132\\
One-minute composite ($0\text{--}60\,\text{s}$) & 7 & 24 & 84\\
One-minute composite, matched ($0\text{--}60\,\text{s}$) & 6 & 24 & 72\\
\bottomrule\end{tabular}
\end{table}

\begin{table*}[t]
 \centering\small\setlength{\tabcolsep}{3pt}
 \caption{\textbf{Short-clip ($50\text{--}60\,\text{s}$) and full four-minute ($0\text{--}240\,\text{s}$) human evaluations.} Values report preference (\%) for Method~A over Method~B (ties $=0.5$, tie rates listed separately) with descriptive, unadjusted crossed source--rater $95\%$ bootstrap intervals. Baseline-only rows check that raters distinguish known-different baselines.}
 \label{tab:st_human_all}
 \begin{tabular}{@{}lllrrrr@{}}\toprule
Preference for A over B & Question & Preference (\%) [95\% CI] & Raters & Sources & Votes & Ties (\%)\\\midrule
\multicolumn{7}{l}{\textit{Baseline variants, 50--60\,s}}\\
\OursLabel vs. Spatial Echo & Scene preservation & 66.5 [48.7, 82.5] & 14 & 24 & 112 & 18.8\\
 & Motion naturalness & 54.9 [38.1, 70.8] & 14 & 24 & 112 & 27.7\\
 & Fewer artifacts & 66.5 [49.5, 81.7] & 14 & 24 & 112 & 22.3\\
\OursLabel vs. Age-decay read & Scene preservation & 76.7 [61.6, 89.9] & 14 & 24 & 105 & 25.7\\
 & Motion naturalness & 75.2 [61.9, 87.2] & 14 & 24 & 105 & 34.3\\
 & Fewer artifacts & 74.8 [59.4, 88.7] & 14 & 24 & 105 & 21.9\\
Static sink vs. Framewise Echo & Scene preservation & 29.7 [6.2, 57.8] & 4 & 8 & 32 & 21.9\\
 & Motion naturalness & 35.9 [14.1, 57.8] & 4 & 8 & 32 & 46.9\\
 & Fewer artifacts & 31.2 [9.4, 50.0] & 4 & 8 & 32 & 31.2\\
Echo, $k=6$ vs. Echo, $k=3$ & Scene preservation & 1.2 [0.0, 7.5] & 5 & 8 & 40 & 2.5\\
 & Motion naturalness & 3.8 [0.0, 12.5] & 5 & 8 & 40 & 7.5\\
 & Fewer artifacts & 1.2 [0.0, 7.5] & 5 & 8 & 40 & 2.5\\
Age-decay read vs. Spatial Echo & Scene preservation & 17.2 [1.6, 37.5] & 4 & 8 & 32 & 21.9\\
 & Motion naturalness & 14.1 [0.0, 34.4] & 4 & 8 & 32 & 15.6\\
 & Fewer artifacts & 7.8 [0.0, 18.8] & 4 & 8 & 32 & 15.6\\
\midrule
\multicolumn{7}{l}{\textit{Guidance/coverage, 50--60\,s}}\\
\OursLabel vs. ALG-inspired & Scene preservation & 80.4 [55.4, 98.1] & 7 & 12 & 28 & 17.9\\
 & Motion naturalness & 71.4 [52.0, 89.7] & 7 & 12 & 28 & 50.0\\
 & Fewer artifacts & 67.9 [41.7, 90.3] & 7 & 12 & 28 & 14.3\\
\OursLabel vs. DyMoS port & Scene preservation & 67.9 [36.5, 93.5] & 7 & 12 & 28 & 14.3\\
 & Motion naturalness & 58.9 [34.5, 84.6] & 7 & 12 & 28 & 39.3\\
 & Fewer artifacts & 57.1 [27.9, 86.2] & 7 & 12 & 28 & 21.4\\
\OursLabel vs. Full coverage, no tether & Scene preservation & 76.8 [55.0, 94.4] & 7 & 12 & 28 & 32.1\\
 & Motion naturalness & 60.7 [47.8, 78.6] & 7 & 12 & 28 & 71.4\\
 & Fewer artifacts & 67.9 [42.9, 88.7] & 7 & 12 & 28 & 42.9\\
\midrule
\multicolumn{7}{l}{\textit{Four-minute, 0--240\,s}}\\
\OursLabel vs. Spatial Echo & Scene preservation & 92.2 [75.0, 100.0] & 8 & 12 & 32 & 3.1\\
 & Less replay/stagnation & 89.1 [70.7, 100.0] & 8 & 12 & 32 & 9.4\\
 & Fewer discontinuities & 93.8 [76.5, 100.0] & 8 & 12 & 32 & 0.0\\
\midrule
\multicolumn{7}{l}{\textit{Baseline validation with activity question, 50--60\,s (separate 16-source module)}}\\
Static sink vs. Framewise Echo & Scene preservation & 27.3 [7.8, 48.4] & 4 & 16 & 64 & 4.7\\
 & Motion naturalness & 34.4 [18.8, 50.8] & 4 & 16 & 64 & 40.6\\
 & Fewer artifacts & 24.2 [7.0, 44.5] & 4 & 16 & 64 & 10.9\\
 & Activity amount & 39.1 [17.2, 60.9] & 4 & 16 & 64 & 25.0\\
Echo, $k=6$ vs. Echo, $k=3$ & Scene preservation & 0.9 [0.0, 5.1] & 7 & 16 & 56 & 1.8\\
 & Motion naturalness & 8.9 [1.0, 19.9] & 7 & 16 & 56 & 17.9\\
 & Fewer artifacts & 3.6 [0.0, 12.1] & 7 & 16 & 56 & 7.1\\
 & Activity amount & 4.5 [0.0, 12.9] & 7 & 16 & 56 & 8.9\\
Age-decay read vs. Spatial Echo & Scene preservation & 43.8 [21.8, 66.7] & 3 & 16 & 48 & 12.5\\
 & Motion naturalness & 36.5 [12.5, 61.5] & 3 & 16 & 48 & 22.9\\
 & Fewer artifacts & 17.7 [3.1, 37.5] & 3 & 16 & 48 & 10.4\\
 & Activity amount & 46.9 [19.8, 72.9] & 3 & 16 & 48 & 14.6\\
\bottomrule\end{tabular}
\end{table*}

\begin{table*}[t]
 \centering\small\setlength{\tabcolsep}{3pt}
 \caption{\textbf{Full one-minute ($0\text{--}60\,\text{s}$) human evaluations.} Preference (\%) for Method~A over Method~B (ties $=0.5$, sources weighted equally) with descriptive, unadjusted crossed source--rater $95\%$ bootstrap intervals. Vote counts apply to each pair and question.}
 \label{tab:st_human_full_duration}
 \begin{tabular}{@{}lllrrrr@{}}\toprule
Preference for A over B & Question & Preference (\%) [95\% CI] & Raters & Sources & Votes & Ties (\%)\\\midrule
\multicolumn{7}{l}{\textit{One minute: reference comparisons}}\\
\OursLabel vs. Framewise Echo & Scene preservation & 87.5 [70.0, 100.0] & 11 & 24 & 33 & 6.1\\
 & Motion naturalness & 78.1 [57.1, 96.4] & 11 & 24 & 33 & 15.2\\
 & Fewer artifacts & 77.1 [54.8, 96.7] & 11 & 24 & 33 & 9.1\\
 & Less replay/stagnation & 79.9 [62.0, 94.4] & 11 & 24 & 33 & 27.3\\
\OursLabel vs. Age-decay read & Scene preservation & 91.7 [72.7, 100.0] & 11 & 24 & 33 & 0.0\\
 & Motion naturalness & 93.8 [78.6, 100.0] & 11 & 24 & 33 & 9.1\\
 & Fewer artifacts & 95.8 [83.3, 100.0] & 11 & 24 & 33 & 3.0\\
 & Less replay/stagnation & 91.7 [76.4, 100.0] & 11 & 24 & 33 & 12.1\\
\OursLabel vs. ALG-inspired & Scene preservation & 87.2 [60.6, 100.0] & 11 & 24 & 33 & 9.1\\
 & Motion naturalness & 80.2 [53.3, 96.9] & 11 & 24 & 33 & 12.1\\
 & Fewer artifacts & 80.2 [53.3, 96.9] & 11 & 24 & 33 & 12.1\\
 & Less replay/stagnation & 86.5 [62.5, 100.0] & 11 & 24 & 33 & 18.2\\
\OursLabel vs. DyMoS port & Scene preservation & 81.9 [60.7, 97.1] & 11 & 24 & 33 & 15.2\\
 & Motion naturalness & 72.2 [51.2, 88.3] & 11 & 24 & 33 & 42.4\\
 & Fewer artifacts & 73.6 [45.3, 94.9] & 11 & 24 & 33 & 18.2\\
 & Less replay/stagnation & 75.3 [50.0, 92.2] & 11 & 24 & 33 & 33.3\\
\midrule
\multicolumn{7}{l}{\textit{One minute: composite comparisons}}\\
\OursLabel vs. Output composite & Scene preservation & 88.5 [70.8, 100.0] & 7 & 24 & 42 & 9.5\\
 & Motion naturalness & 80.6 [64.6, 92.4] & 7 & 24 & 42 & 40.5\\
 & Fewer artifacts & 87.2 [72.5, 97.5] & 7 & 24 & 42 & 19.0\\
 & Less replay/stagnation & 86.5 [71.4, 95.6] & 7 & 24 & 42 & 31.0\\
\OursLabel vs. Full coverage, no tether & Scene preservation & 67.0 [48.1, 87.9] & 7 & 24 & 42 & 16.7\\
 & Motion naturalness & 64.6 [47.7, 80.6] & 7 & 24 & 42 & 38.1\\
 & Fewer artifacts & 75.7 [57.5, 90.6] & 7 & 24 & 42 & 21.4\\
 & Less replay/stagnation & 66.7 [50.0, 83.3] & 7 & 24 & 42 & 33.3\\
\midrule
\multicolumn{7}{l}{\textit{One minute: composite comparisons, matched encoding}}\\
\OursLabel vs. Output composite & Scene preservation & 97.9 [89.5, 100.0] & 6 & 24 & 36 & 5.6\\
 & Motion naturalness & 75.0 [61.4, 88.2] & 6 & 24 & 36 & 44.4\\
 & Fewer artifacts & 91.7 [78.9, 100.0] & 6 & 24 & 36 & 19.4\\
 & Less replay/stagnation & 80.2 [66.2, 93.1] & 6 & 24 & 36 & 33.3\\
\OursLabel vs. Full coverage, no tether & Scene preservation & 78.1 [62.5, 94.5] & 6 & 24 & 36 & 22.2\\
 & Motion naturalness & 61.5 [47.4, 77.8] & 6 & 24 & 36 & 52.8\\
 & Fewer artifacts & 75.0 [62.7, 91.0] & 6 & 24 & 36 & 38.9\\
 & Less replay/stagnation & 62.5 [46.9, 80.3] & 6 & 24 & 36 & 44.4\\
\bottomrule\end{tabular}

\end{table*}

\section{Ablations, transfer, and limitations}
\label{fm:suppl_ablations}
\begin{table}[t]
 \centering\small\setlength{\tabcolsep}{2pt}
 \caption{\textbf{Source-registration failures at $60\,\text{s}$} across the 45 geometry-eligible benchmark scenes (seed 42), complementing the main-paper fidelity and activity tables.}
 \label{tab:st_all_reference_rows}
 \begin{tabular}{lr}\toprule
Method & Registration failures\\\midrule
CF++ + Echo & 5/45\\
Spatial Echo, k=3 & 11/45\\
Static sink, k=3 & 6/45\\
Age-decay read & \textbf{3}/45\\
DyMoS port, k=1 & 6/45\\
ALG-inspired, k=1 & 9/45\\
UnStep port, k=3 & 10/45\\
PackCache-style & \underline{4}/45\\
FramePack-style & 5/45\\
Full coverage, no tether & 7/45\\
Frozen gated read & 6/45\\
\textbf{SceneTether (ours)} & \textbf{3}/45\\
\bottomrule\end{tabular}

\end{table}
\begin{table}[t]
 \centering\small\setlength{\tabcolsep}{3pt}
 \caption{\textbf{Development-split observations} across the seven scenes that guided method design ($60\,\text{s}$, single seed; 7 appearance/event scenes, 5 water-eligible scenes).}
 \label{tab:fm_controls}
 \begin{tabular}{lrrr}\toprule
Development method & PSNR $\uparrow$ & Color $\downarrow$ & Events $\downarrow$\\\midrule
Frozen gated read & 17.03 & 5.60 & 45.5\\
Half tether & \underline{18.49} & \textbf{3.75} & \underline{41.9}\\
\textbf{SceneTether (ours)} & \textbf{21.46} & \underline{5.55} & \textbf{36.2}\\
\bottomrule\end{tabular}

\end{table}

\begin{figure*}[t]
 \centering
 \includegraphics[width=.97\textwidth]{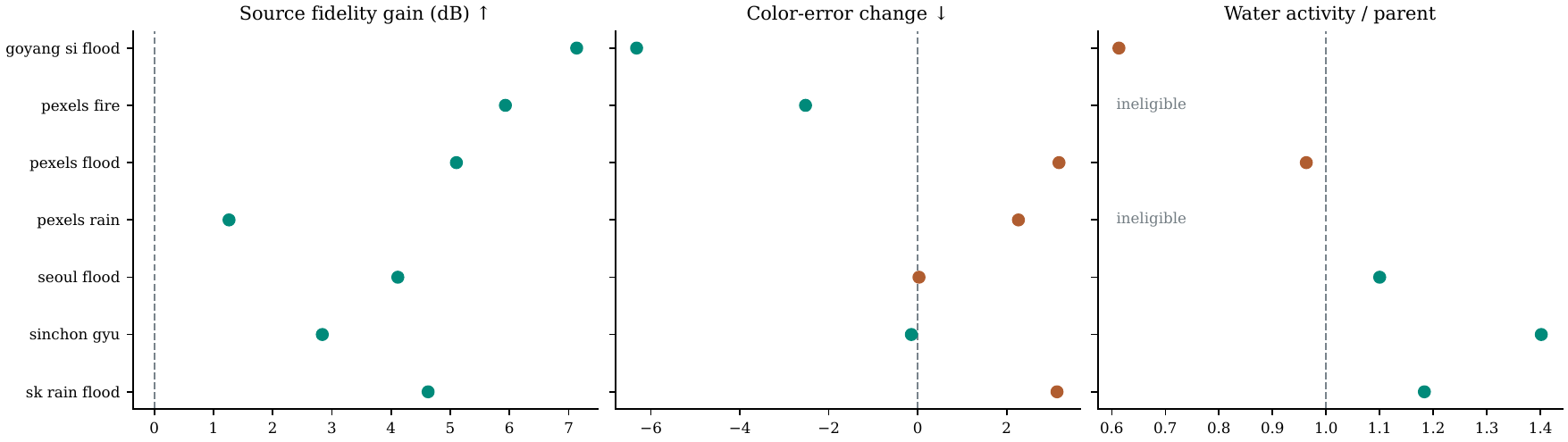}
 \caption{\textbf{Per-scene development trade-offs} comparing \Ours against the sparse frozen read across the seven development scenes (water flow abstains on fire and rain; development observations only).}
 \label{fig:st_dev_tradeoffs}
\end{figure*}

The two-seed equal-coverage controls (\cref{tab:st_paired_completed}) isolate value interpolation within the deployed read configuration, whereas the cumulative construction stages varied coverage and values jointly. Full tethering raises PSNR substantially over the untethered read with an inconclusive flow ratio of \stres{coverage_flow}, and exceeds half-strength tethering in PSNR with unresolved color and flow differences.

\paragraph{Causal backbones for long image animation.}
Transferring a recent-KV read policy requires a compatible causal generator with a usable initial-image prefix. Several released few-step systems---including CF++~\cite{zhao2026causal_pp}, Causal Forcing~\cite{zhu2026causal}, Self-Forcing~\cite{huang2025selfforcing}, Rolling Forcing~\cite{liu2025rolling}, Reward Forcing~\cite{lu2025reward}, Infinite Forcing~\cite{infinite-forcing}, LongLive~\cite{yang2025longlive}, and CausVid~\cite{yin2025causvid}---primarily target text-conditioned causal rollout; providing a cached first image does not by itself establish native I2V training or reliable long-horizon image animation.
In our tested public configurations, direct image-prefix adaptation was not uniformly available: LongLive lacked the required initial-latent interface, the tested Rolling Forcing image path failed at runtime, Reward Forcing overwrote the initial sink, and the inspected CausVid I2V release provided ODE initialization rather than a deployable few-step student. Self-Forcing produced severe artifacts at an unsupported $k=1$ setting (${\sim}9\,\text{dB}$ PSNR) but was more coherent at its trained $k=3$ block size; these screening outcomes are implementation-specific, not general claims that the original methods cannot generate video.
Conversely, native I2V models such as Wan2.1/2.2-I2V~\cite{wan2025}, CogVideoX-5B-I2V~\cite{yang2025cogvideox}, and LTX-Video~\cite{hacohen2024ltxvideo} denoise finite bidirectional clips; extending them by RGB clip continuation (\eg, SVI 2.0 Pro~\cite{li2025stable}) exposes no persistent newest-latent KV cache, so \Ours cannot be applied to them as a drop-in read policy.
Although SVI 2.0 Pro leads the seven-source $5\,\text{s}$ native-fidelity context (\cref{tab:st_native5}), on \stres{nl_n} shared long-horizon scenes its source PSNR changes from \stres{nl_svi60}\,\text{dB} at $60\,\text{s}$ to \stres{nl_svi240}\,\text{dB} at $240\,\text{s}$, versus \stres{nl_st60} and \stres{nl_st240}\,\text{dB} for \Ours. \Ours scores higher on \stres{nl_wins240}/\stres{nl_n} scenes at four minutes (\cref{tab:st_native_long}); throughput differs substantially (\cref{tab:st_cost}), but this is a system-level comparison across unmatched backbones, training, and samplers, not a controlled architectural ranking.
Helios~\cite{helios} recomputes compressed history without cached frame reads, and image-trained MAGI-1~\cite{ai2025magi1autoregressivevideogeneration} denoises six-latent chunks against a sliding four-chunk cache where an exploratory port fails to curb drift (\cref{fig:st_magi}).
Within the CF++ family, applying \Ours with unchanged settings to the one-step student gains \stres{onestep_psnr}\,\text{dB} alongside higher activity (\cref{tab:st_onestep}); we therefore scope positive transfer claims to the one- and two-step CF++ students (other backbones appear in \cref{fm:suppl_backbones}).

\begin{table}[t]
 \centering\scriptsize\setlength{\tabcolsep}{3pt}
 \caption{\textbf{Transfer within the CF++ student family at $60\,\text{s}$} (\Ours minus each student's framewise Echo host across 48 test scenes, seed 42; \stres{onestep_n} appearance scenes; registration-qualified activity counts in parentheses).}
 \label{tab:st_onestep}
 \begin{tabular}{lrrr}\toprule
Student & $\Delta$PSNR (dB) & $\Delta$Color & Flow ($n$)\\\midrule
Two-step & +5.39 [+4.40, +6.48] & $-$9.98 [$-$13.25, $-$6.93] & 1.44 [1.19, 1.74] (38)\\
One-step & +2.86 [+1.90, +3.85] & $-$4.75 [$-$8.89, $-$0.58] & 1.66 [1.26, 2.28] (37)\\
\bottomrule\end{tabular}

\end{table}

\paragraph{Speed by backbone.}
Exclusive single-A100 profiling (batch~1, $832\times480$, 961 frames; one warm-up verification run followed by three timed scenes) shows that backbone step count governs wall-clock speed while \Ours adds a small gather-and-blend overhead (\cref{tab:st_cost}): about \stres{cfpp2_overhead_pct}\% extra wall time on the two-step student, which is below the table's rounding.
Because the evaluated CF++ pipeline decodes latents after rollout rather than delivering frames continuously, we report throughput with and without VAE decoding; these are offline rates, not live delivery rates. SVI 2.0 Pro throughput is reported under its native pipeline; that comparison also changes the model, sampler, and continuation scheme, so it does not isolate denoising-step count.

\begin{table}[t]
 \centering\scriptsize\setlength{\tabcolsep}{3pt}
 \caption{\textbf{Inference throughput on one exclusive A100 GPU} (batch~1, $832\times480$). CF++ rows compare the framewise Echo host with and without \Ours ($16\,\text{fps}$); SVI 2.0 Pro ($15\,\text{fps}$) logs native bidirectional clip-continuation throughput including VAE passes.}
 \label{tab:st_cost}
 \begin{tabular}{@{}lcccc@{}}\toprule
Backbone (host) & \shortstack{Frames/s\\rollout} & \shortstack{Frames/s\\with decode} & \shortstack{Wall s per\\output s} & \shortstack{Peak\\GiB}\\\midrule
CF++ two-step, 1.3B & 7.1 & 5.5 & 2.9 & 27.5 \\
\quad + \OursLabel & 7.1 & 5.5 & 2.9 & 27.5 \\
CF++ one-step, 1.3B & 10.5 & 7.3 & 2.2 & 27.5 \\
\quad + \OursLabel & 10.6 & 7.4 & 2.2 & 27.5 \\
SVI 2.0 Pro, Wan2.2-A14B & -- & 0.087 & 172 & 68.7 \\
\quad + \OursLabel & \multicolumn{4}{c}{not implemented} \\
\bottomrule\end{tabular}

\end{table}

\paragraph{Adapted references.}
\Cref{tab:st_ports} documents the exact tensor interventions, target layers, denoising steps, hyperparameters, and untransferred native components for every reference row in Tab.~1 of the main paper (all sharing the CFG-free two-step CF++ student).
We selected the age-decay parameter $\rho=0.3$ and the spatial-index handling of the $k=3$ Echo port on the seven development scenes. Other port settings follow the stated published or pre-declared values; porting to CF++ does not reproduce the original model-specific pipelines.
Porting TetherCache~\cite{meng2026tethercache} onto its matching static-sink + FIFO host ($k=1$) reduces $60\,\text{s}$ PSNR from $16.23$ to $14.38\,\text{dB}$ and increases color error from $17.80$ to $23.47$.
Because framewise Echo's rotating anchors include $z_0$ in $68\%$ of blocks whereas \Ours pins $z_0$ in every block, we also re-ran the ALG-inspired and DyMoS ports with $z_0$ pinned in every block (\cref{tab:st_pinned_ports}): pinning shifts PSNR by \stres{pp_algpin_dpsnr}\,\text{dB} [\stres{pp_algpin_dpsnr_lo}, \stres{pp_algpin_dpsnr_hi}] and \stres{pp_dympin_dpsnr}\,\text{dB} [\stres{pp_dympin_dpsnr_lo}, \stres{pp_dympin_dpsnr_hi}], increases color error, and introduces severe foreground hallucinations (e.g., striped arches over lava), leaving \Ours ahead by \stres{pp_st_algpin_dpsnr} and \stres{pp_st_dympin_dpsnr}\,\text{dB}. Pinning consumes one anchor slot and therefore also changes the retained history.

\begin{table*}[t]
 \centering\small\setlength{\tabcolsep}{3pt}
 \caption{\textbf{Specification of causal memory references and adapted inference ports} in Tab.~1 of the main paper. Metric-eligibility rules are given in Sec.~\ref{fm:suppl_protocol}. All rows start from the frozen CFG-free CF++ two-step checkpoint, and only the UnStep port modifies projection matrices as specified below ($z_0$ prefill at $t=0$, 21-latent window except UnStep at 9, $T=1560$ tokens/latent, latent-indexed noise, seed 42, nominal steps $1000,750,500,250$ then $1000,500$ plus $t=0$ refresh). Without source pinning, Echo's rotating anchors expose $z_0$ in $68\%$ of reads; age-decay, PackCache-style, and FramePack-style reads omit rotating anchors and compressed tokens at steady state ($d$: latent age; $k$: latents/block).}
 \label{tab:st_ports}
 \begin{tabular}{@{}>{\raggedright\arraybackslash}p{.10\textwidth}>{\raggedright\arraybackslash}p{.21\textwidth}>{\raggedright\arraybackslash}p{.065\textwidth}>{\raggedright\arraybackslash}p{.11\textwidth}>{\raggedright\arraybackslash}p{.11\textwidth}>{\raggedright\arraybackslash}p{.14\textwidth}>{\raggedright\arraybackslash}p{.15\textwidth}@{}}
 \toprule
 Row (host) & Intervention & Layers & Evaluations & Settings & Parameter source & Untransferred native design \\
 \midrule
 \multicolumn{7}{@{}l}{\textit{Reference memories}}\\
 Framewise Echo, $k=1$ & Cache selection: 12 anchors rotating over first 18 latents, 3-latent evicted compressed budget ranked against mean early query, 6 recent latents (including current); full 21-latent read & all 30 & all & anchors 12, comp.\ 3, recent 6, pool 18, stride 3 & Host single-scene Echo-Forcing config & Scene-recall frames and difference-aware decay for multi-cut transitions \\
 Echo-Forcing port, $k=3$ & Rescaled $k=3$ entry counts (4 anchor and 2 recent 3-latent chunks, 3-latent compressed budget); compressed keys retain spatial indices & all 30 & all & matched 21-latent budget; spatial indices on & Rescaled host budget; index fix validated on 7 dev scenes & Same multi-scene modules as above \\
 Static sink, $k=3$ & Retains $z_0$ plus latents 1--2 as fixed sink with FIFO window and no EMA blending & all 30 & all & sink 3, window 21 & Base CF++ sink configuration & -- \\
 Age-decay read, $k=1$ & Pins $z_0$ on Echo; reads top $\mathrm{round}(1560\cdot0.3^{d})$ key-norm tokens at age $d$ ($468, 140, 42, 13, 4$) for K and V; $z_0$ and current block whole ($3787$ tokens) & all 30 & all & $\rho=0.3$, key-norm rank & Tuned on 7 dev scenes & Shares operator with PackCache-style row at steeper $\rho$ \\
 \bottomrule
 \end{tabular}
\end{table*}

\begin{table*}[t]
 \ContinuedFloat
 \centering\small\setlength{\tabcolsep}{3pt}
 \caption[]{\textbf{Specification of causal memory references and adapted inference ports} (continued).}
 \begin{tabular}{@{}>{\raggedright\arraybackslash}p{.10\textwidth}>{\raggedright\arraybackslash}p{.21\textwidth}>{\raggedright\arraybackslash}p{.065\textwidth}>{\raggedright\arraybackslash}p{.11\textwidth}>{\raggedright\arraybackslash}p{.11\textwidth}>{\raggedright\arraybackslash}p{.14\textwidth}>{\raggedright\arraybackslash}p{.15\textwidth}@{}}
 \toprule
 Row (host) & Intervention & Layers & Evaluations & Settings & Parameter source & Untransferred native design \\
 \midrule
 \multicolumn{7}{@{}l}{\textit{Adapted inference methods}}\\
 ALG-inspired, $k=1$ (framewise Echo) & Substitutes source K/V from a $5\times5$ Gaussian-blurred $z_0$ prefill pass without overwriting stored cache & all 30 & nominal $t\ge501$ ($1000,750$ in block~1; $1000$ later) & $\sigma=1$ latent px, cutoff $501$ & Pre-declared analogue of ALG's $2.5\times$ bilinear filter & Native ALG filters frame-concatenated conditioning inside CFG difference over first $10\text{--}20\%$ of 50 steps \\
 DyMoS port, $k=1$ (framewise Echo) & Subtracts $\gamma$ from pre-softmax logits $QK^\top/\sqrt{d_h}$ ($d_h=128$) of generated queries onto source keys & all 30 & first step per block (nominal $1000$); off at refresh & $\gamma=0.6$, uniform weight & Published Wan2.2 default ($\lambda=0.2$); dev checks retained default & Native bidirectional 81-frame Wan2.2-I2V-A14B (40 CFG steps, conditional-branch bias on steps 0--7) \\
 UnStep port, $k=3$ (static sink) & (i) 9-latent cache (sink 3 + 6) with 18 layers reading last 8 latents; (ii) noisy $t=0$ refresh $0.98\,x_0+0.02\,\epsilon$ and re-emission; (iii) rank-$1413/1536$ SVD on V and O projections & (i,ii) all; (iii) 8--29 & (i,iii) all; (ii) refresh & buffer 9, $\sigma^\ast=0.02$, rank $0.92$ & Released code and paper Tab.~1 defaults & Native Self-Forcing checkpoint, warped step $625$ (vs.\ $833.3$), and compiled custom kernels \\
 PackCache-style, $k=1$ (Echo, $z_0$ pinned) & Reads top $\mathrm{round}(1560\cdot0.5^{d})$ key-norm tokens at age $d$ ($780, 390, 195, 98, 49$) post-RoPE; $z_0$ and current block whole ($4632$ tokens) & all 30 & all & $\rho=0.5$, key-norm rank & Paper $\rho=0.5$; pre-declared norm ranking & Discrete masked-token decoding, text anchor, and kernel speedup (no official code) \\
 FramePack-style, $k=1$ (Echo, $z_0$ pinned) & Read-time average pooling of encoded K/V: $1\times$ on newest latent, $2\times2$ on next 2, $4\times4$ on next 16 (2 in steady state); $z_0$ and current block whole ($5668$ tokens) & all 30 & all & $1 / 2 / 16$ schedule, drop older & FramePack-F1 demo schedule & Learned multi-rate input projections and anti-drift training \\
 TetherCache port, $k=1$ (static sink, $k=1$) & Splits window into sink~3, 14-latent memory, and 4 recent latents; refresh selects memory via importance + diversity (GRAB) and aligns K/V moments to sink + memory (TAME) & all 30 & GRAB/TAME at refresh; read at all & $\alpha=0.35$, $\tau=0.6$, temp.\ $1$ & Paper $\tau=0.6$ and released code $\alpha=0.35$ & Native chunked Self-Forcing backbone and multi-latent block size \\
 \bottomrule
 \end{tabular}
\end{table*}

\paragraph{Distinction from closest feature edits and perceptual scope.}
Whereas TetherCache~\cite{meng2026tethercache} aligns moments of recalled distant K/V entries and Custom Forcing~\cite{ok2026customforcing} scales reference subject values via identity drift, \Ours leaves keys untouched while spatially interpolating newest-frame values toward co-located source values under a frozen warm-up map.
On the perceptual side, $10\,\text{s}$ late-window trials assess local quality after accumulated history; only the full-duration modules expose intervening events and longer replay cycles (\cref{tab:st_human_all,tab:st_human_full_duration}).

\section{Visual examples and reproducibility}
\label{fm:suppl_gallery}
\begin{figure*}[t]
 \centering\raisebox{-104.12573pt}{\includegraphics{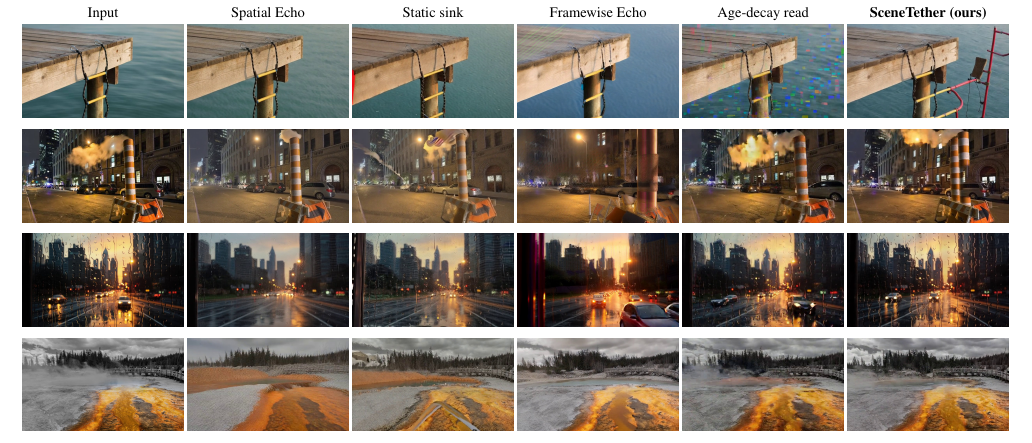}}
 \caption{\textbf{Stratified random draw from remaining benchmark scenes} at $58\,\text{s}$ (four categories sampled via seed 20261021 after excluding score-selected showcases; shared inputs and noise). \Ours hallucinates a red ladder beside the pier and a car later in the rain rollout.}
 \label{fig:st_random_gallery}
\end{figure*}

\begin{figure*}[t]
 \centering\raisebox{-64.13460pt}{\includegraphics{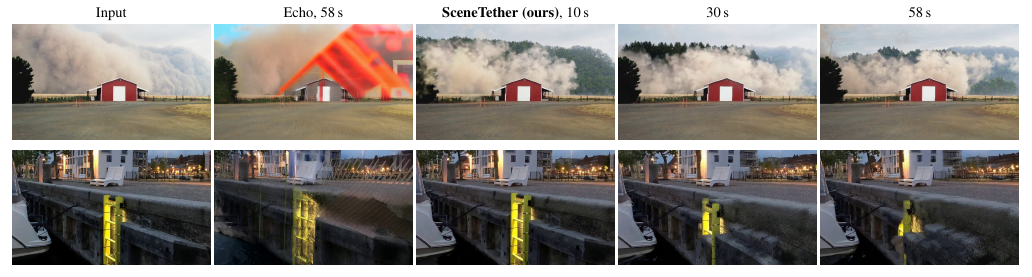}}
 \caption{\textbf{Lowest static-PSNR scenes} for \Ours at $60\,\text{s}$ (seed 42) among appearance-eligible benchmark sources not shown elsewhere, displayed at native resolution.}
 \label{fig:st_lower_fidelity}
\end{figure*}

\begin{figure*}[t]
 \centering\raisebox{-124.76920pt}{\includegraphics{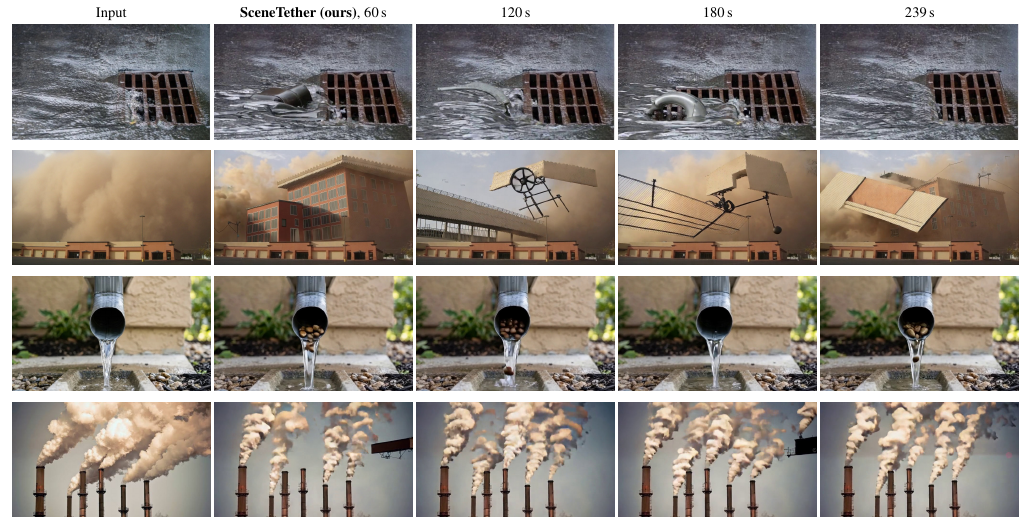}}
 \caption{\textbf{Foreground hallucinations under \Ours across $240\,\text{s}$ rollouts.} Four scenes excluded from showcase figures during visual screening ($60, 120, 180,$ and $239\,\text{s}$): a dark cylinder beside the drain, a kite-like artifact above the dust-covered building, pebbles inside the downspout, and an overhead gantry. Two of these scenes rank highest in $60\,\text{s}$ static PSNR margin, underscoring that static-support fidelity does not prevent foreground inventions.}
 \label{fig:st_failures}
\end{figure*}

To disclose showcase selection, we ranked the 42 appearance-eligible benchmark scenes by \Ours's $60\,\text{s}$ PSNR margin over the strongest displayed baseline (seed 42), visually audited leading candidates through $240\,\text{s}$, and documented visually detected object or text hallucinations in a separate failure gallery (\cref{fig:st_failures}) rather than treating selected showcases as representative samples.
We distributed remaining candidates across diverse natural media (rapids, streams, weirs, breaking waves, fire, smoke, steam, dust, storm clouds, lava) alongside visually selected precipitation examples, complementing them with a category-stratified random draw from the scenes not already selected (\cref{fig:st_random_gallery}) and the lowest-PSNR remaining scenes (\cref{fig:st_lower_fidelity}). Because the random draw excludes the score-selected showcases, it represents that remainder rather than the full benchmark.
All frames are native-resolution PNGs without post-processing, and we keep the rankings, contact sheets, and rejected scenes behind every figure.


\begin{figure}[t]
 \centering
 \raisebox{-198.42943pt}{\includegraphics{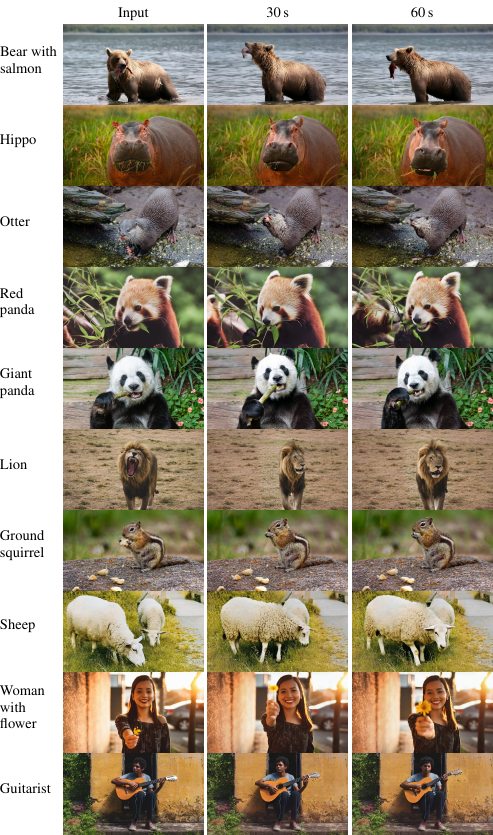}}
 \caption{\textbf{Exploratory out-of-scope rollouts} of \Ours with default settings across ten articulated VBench-I2V and DyMoS scenes ($60\,\text{s}$, seed 42, unscored qualitative stills).}
 \label{fig:st_oos_gallery}
\end{figure}

\begin{figure}[t]
 \centering
 \raisebox{-81.47177pt}{\includegraphics{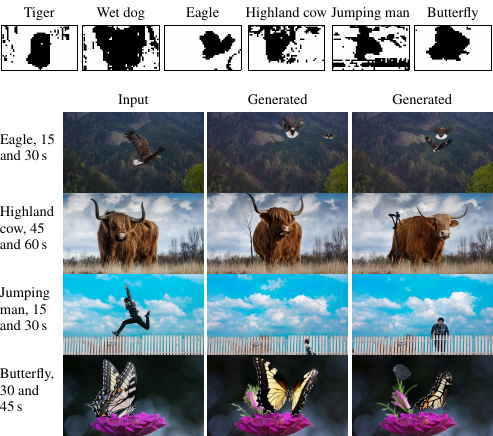}}
 \caption{\textbf{Out-of-scope activity maps and failure modes.} \textit{Top:} Frozen warm-up maps for the two main-paper scenes and four failures (black denotes untethered active regions $M(p)=1$). \textit{Bottom:} Subject duplication, spurious foreground objects, and subject disappearance across single seed-42 rollouts.}
 \label{fig:st_oos}
\end{figure}

\paragraph{Out-of-scope samples.}
To probe behavior on articulated subjects outside our natural-flow scope, we cropped 30 VBench-I2V images~\cite{huang2024vbench++} (including 10 featured in DyMoS) and two bundled DyMoS release frames~\cite{jeon2026dymos} to $832\times480$, authored benchmark-style fixed-camera prompts, and ran single unscored $60\,\text{s}$ rollouts at seed 42; we exclude depictions of children from the displayed examples.
From $15\,\text{s}$-interval contact sheets, we selected two examples for Fig.~6 of the main paper, ten for \cref{fig:st_oos_gallery}, and four failure cases (\cref{fig:st_oos}): a duplicated eagle, a spurious object beside the highland cow, a vanishing/reappearing jumper, and a hallucinated plant beside the butterfly.
Because high-activity tokens remain untethered ($\beta=0$), subjects moving during warm-up receive no direct value anchoring against structural drift or duplication; conversely, motion that begins after warm-up falls in tethered regions and can be suppressed.

\begin{figure*}[t]
 \centering\raisebox{-64.13460pt}{\includegraphics{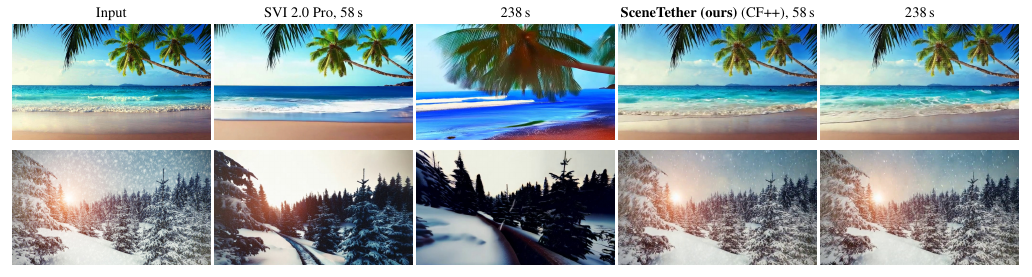}}
 \caption{\textbf{Qualitative system comparison with native SVI 2.0 Pro} (Wan2.2-I2V-A14B, 50 steps, CFG 4, $15\,\text{fps}$) alongside \Ours on the 1.3B two-step CF++ student near $58\,\text{s}$ and $238\,\text{s}$ on the two remaining shared rollout sources.}
 \label{fig:st_native_svi}
\end{figure*}
\begin{figure*}[t]
 \centering\raisebox{-124.76920pt}{\includegraphics{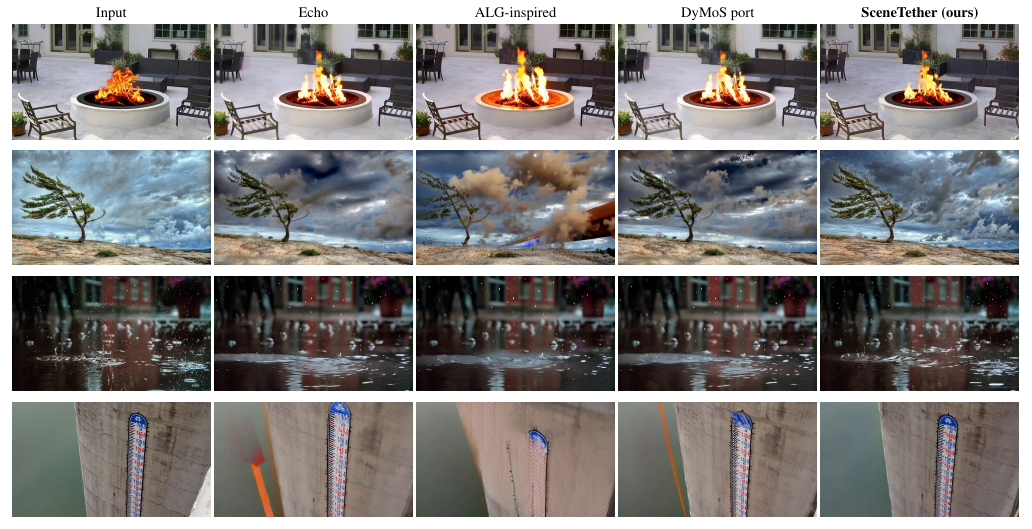}}
 \caption{\textbf{Additional $58\,\text{s}$ comparisons against causal guidance adaptations} (framewise Echo, ALG-inspired, and DyMoS ports under matched inputs and noise, selected via the disclosed showcase rule).}
 \label{fig:st_guidance_extra}
\end{figure*}

\begin{figure*}[t]
 \centering\raisebox{-64.26886pt}{\includegraphics{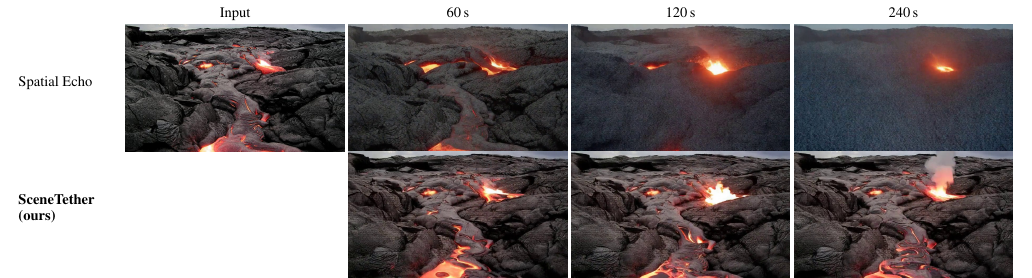}}
 \caption{\textbf{Continuous $240\,\text{s}$ lava-field rollout} comparing spatial Echo-Forcing and \Ours at matched timestamps (seed 42).}
 \label{fig:st_long0}
\end{figure*}

\begin{figure*}[t]
 \centering\raisebox{-64.26886pt}{\includegraphics{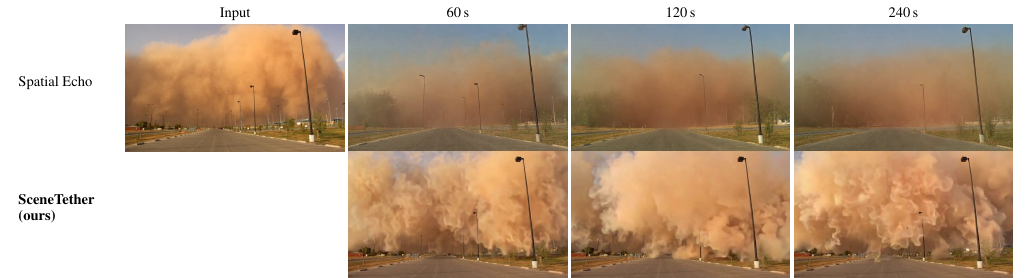}}
 \caption{\textbf{Continuous $240\,\text{s}$ dust-wall rollout} under the protocol of \cref{fig:st_long0}.}
 \label{fig:st_long1}
\end{figure*}

\begin{figure*}[t]
 \centering\raisebox{-64.26886pt}{\includegraphics{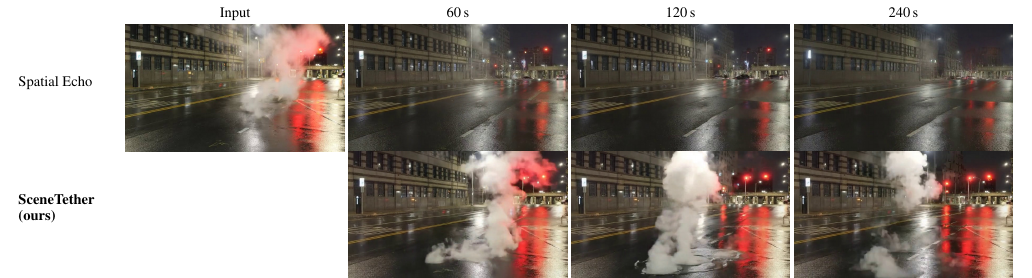}}
 \caption{\textbf{Continuous $240\,\text{s}$ street-steam rollout} under the protocol of \cref{fig:st_long0}.}
 \label{fig:st_long2}
\end{figure*}

\begin{figure*}[t]
 \centering\raisebox{-64.26886pt}{\includegraphics{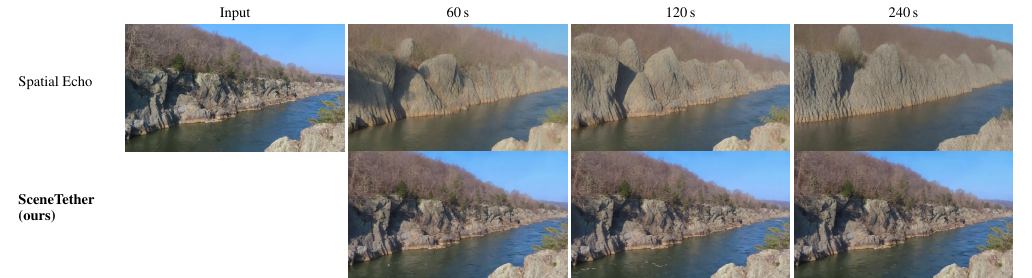}}
 \caption{\textbf{Continuous $240\,\text{s}$ river-gorge rollout} under the protocol of \cref{fig:st_long0}.}
 \label{fig:st_long3}
\end{figure*}

\begin{figure*}[t]
 \centering\raisebox{-64.26886pt}{\includegraphics{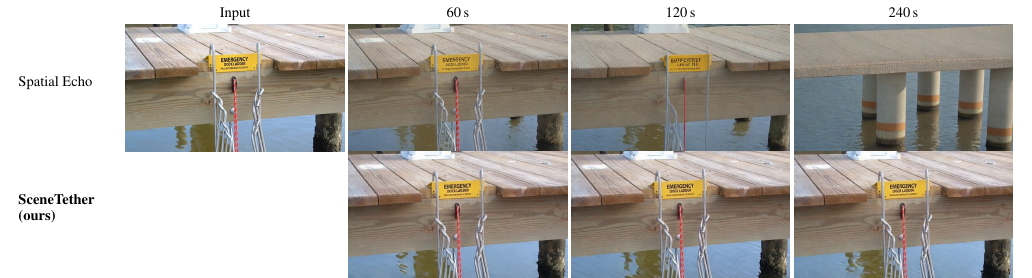}}
 \caption{\textbf{Continuous $240\,\text{s}$ harbor-water rollout} under the protocol of \cref{fig:st_long0}.}
 \label{fig:st_long4}
\end{figure*}

\begin{figure*}[t]
 \centering\raisebox{-64.26886pt}{\includegraphics{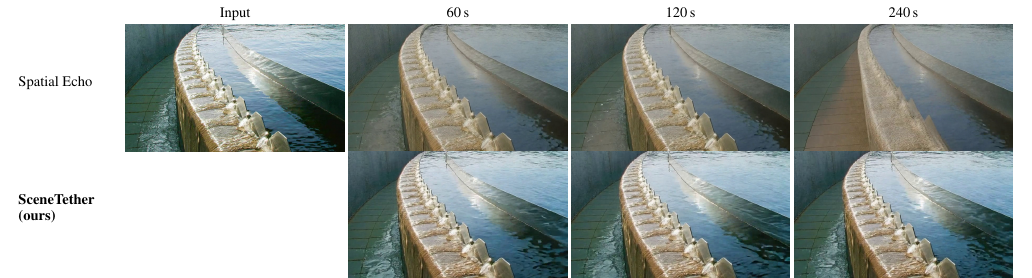}}
 \caption{\textbf{Continuous $240\,\text{s}$ weir rollout} under the protocol of \cref{fig:st_long0}.}
 \label{fig:st_long5}
\end{figure*}

\paragraph{Long-horizon gallery selection.}
\Cref{fig:st_long0,fig:st_long1,fig:st_long2,fig:st_long3,fig:st_long4,fig:st_long5} display six continuous $240\,\text{s}$ rollouts selected under the same rule without overlap with other figures (all embedded as unenhanced native-resolution PNGs).


\clearpage
\section{Additional controls and sensitivity}
\label{fm:suppl_controls}
\Cref{tab:extra_mechanism_controls} details our two-seed value-rule controls at $60\,\text{s}$.
Applying a uniform mean weight or permuting weights across the whole frame also anchors positions in $M$; these controls reduce measured activity with inconclusive PSNR shifts.
Keeping $\beta=0$ on $M$ while flattening or permuting only the remaining weights raises static PSNR, with flow-ratio intervals that span $1.0$ (no detected change, not demonstrated equivalence).
Inspecting the static-only mean-weight control at seed 42 (\stres{sc_n} scenes, regions of \cref{tab:st_sharpness}) reveals lower band variation (\stres{sc_band_uni} vs.\ \stres{sc_band_st} gray levels; paired difference [\stres{sc_band_lo}, \stres{sc_band_hi}]), comparable anchored per-frame edge retention (\stres{sc_edge_uni} vs.\ \stres{sc_edge_st}), comparable moving-region color drift (\stres{sc_drift_uni} vs.\ \stres{sc_drift_st}; [\stres{sc_drift_lo}, \stres{sc_drift_hi}]), and no visible mask-boundary seams at $58\,\text{s}$ on top-difference scenes.
These controls indicate that sparing $M$, not the interior weight grading, limits motion suppression, and that the graded interior weights are not essential. Because we locked \Ours before running these checks, we report the method unchanged.
\Cref{tab:extra_composite240} reports the $240\,\text{s}$ composite PSNR contrast.

\begin{table*}[t]
 \centering\small
 \caption{\textbf{Two-seed value-rule and refresh controls at $60\,\text{s}$} (nesting seeds 42 and 1042 across 42 appearance and 40 activity scenes; 10{,}000 scene-bootstrap draws). Whole-frame uniform and permuted controls redistribute weights across all positions, including $M$, whereas static-only variants preserve $\beta=0$ on $M$ and flatten or shuffle remaining weights; the refresh control disables tethering only during the $t=0$ cache-commit pass ($\Delta\text{PSNR}$: row minus \Ours; flow ratio: row over \Ours).}
 \label{tab:extra_mechanism_controls}
 \begin{tabular}{@{}lrrrrr@{}}\toprule
Method & PSNR $\uparrow$ & $\Delta$PSNR vs. ours [95\% CI] & Color $\downarrow$ & Activity & Activity ratio [95\% CI]\\\midrule
\OursLabel & 22.52 & -- & 5.78 & 2.13 & --\\
Uniform weights & 21.94 & -0.58 [-1.50, +0.28] & 5.30 & 1.39 & 0.65 [0.52, 0.82]\\
Permuted weights & 22.01 & -0.51 [-1.35, +0.27] & 5.13 & 0.99 & 0.47 [0.34, 0.62]\\
No refresh tether & 22.45 & -0.07 [-0.19, +0.04] & 5.97 & 2.08 & 0.98 [0.91, 1.05]\\
Static-only mean weight & 23.04 & +0.52 [+0.29, +0.78] & 5.62 & 2.11 & 0.99 [0.92, 1.07]\\
Static-only permuted & 23.06 & +0.54 [+0.38, +0.73] & 5.40 & 2.03 & 0.95 [0.88, 1.03]\\
\bottomrule\end{tabular}

\end{table*}
\begin{table*}[t]
 \centering\small
 \caption{\textbf{Four-minute post-hoc composite control} over the final $60\,\text{s}$ of the 12 pre-selected horizon scenes (seed 42; 10 appearance, 12 band-variation, and 11 anchored-variation scenes; $\Delta\text{PSNR}$ reports composite minus \Ours with 95\% scene-bootstrap intervals).}
 \label{tab:extra_composite240}
 \begin{tabular}{@{}lrrrrr@{}}\toprule
Method & PSNR $\uparrow$ & $\Delta$PSNR vs. ours [95\% CI] & Color $\downarrow$ & Band variation & Anchored variation\\\midrule
\OursLabel & 24.18 & -- & 5.03 & 5.78 & 3.75\\
Full coverage, no tether & 15.60 & -8.58 [-10.91, -6.35] & 21.46 & 9.09 & 7.64\\
Composite (post-hoc) & 26.63 & +2.45 [-0.06, +5.26] & 5.28 & 4.61 & 0.20\\
\bottomrule\end{tabular}

\end{table*}

\begin{table*}[t]
 \centering\small\setlength{\tabcolsep}{3pt}
 \caption{\textbf{Long-horizon transfer to the one-step CF++ student} at $120\,\text{s}$ and $240\,\text{s}$ (final $60\,\text{s}$, seed 42, 48 sources: 42 appearance, 40 activity, 48 VBench; \textbf{bold}/\underline{underline} mark top directional scores per panel).}
 \label{tab:extra_onestep_horizons}
 \begin{tabular}{@{}lrrrrrrrrr@{}}\toprule
 & \multicolumn{3}{c}{SNF-Bench} & \multicolumn{2}{c}{VBench-I2V} & \multicolumn{4}{c}{VBench-Long}\\
Method & PSNR $\uparrow$ & Color $\downarrow$ & Act. & Subject $\uparrow$ & Back. $\uparrow$ & Smooth $\uparrow$ & Dyn. & Aesth. $\uparrow$ & Image $\uparrow$\\\midrule
\multicolumn{10}{l}{\textit{120 seconds}}\\
Framewise Echo & \underline{18.15} & \underline{15.90} & 0.48 & \underline{0.930} & \underline{0.931} & \textbf{0.992} & 0.069 & \underline{0.538} & \textbf{0.680}\\
\OursLabel & \textbf{20.85} & \textbf{9.99} & 1.60 & \textbf{0.958} & \textbf{0.955} & \underline{0.990} & 0.259 & \textbf{0.548} & \underline{0.678}\\
\midrule
\multicolumn{10}{l}{\textit{240 seconds}}\\
Framewise Echo & \underline{18.22} & \underline{15.73} & 0.74 & \underline{0.925} & \underline{0.930} & \textbf{0.991} & 0.046 & \underline{0.535} & \underline{0.676}\\
\OursLabel & \textbf{20.73} & \textbf{10.47} & 1.62 & \textbf{0.959} & \textbf{0.954} & \underline{0.990} & 0.260 & \textbf{0.548} & \textbf{0.677}\\
\bottomrule\end{tabular}

\end{table*}

\begin{figure*}[t]
 \centering
 \raisebox{-158.28198pt}{\includegraphics{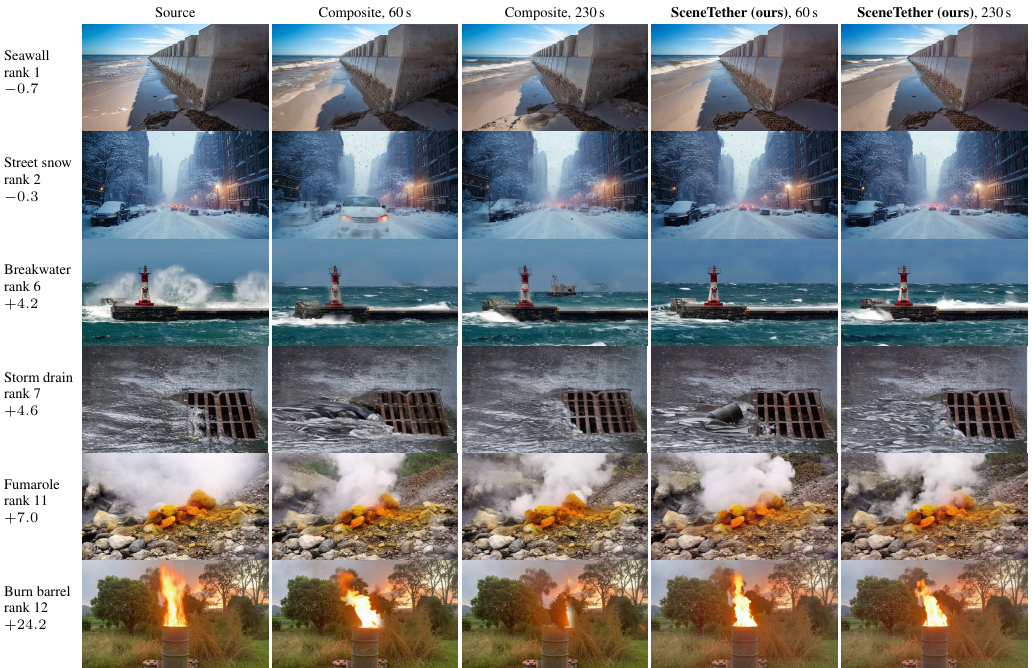}}
 \caption{\textbf{Rank-stratified comparisons of the post-hoc composite and \Ours at $60$ and $230\,\text{s}$.} We ranked the 12 horizon scenes by composite-minus-\Ours moving-region drift, from lowest to highest, and show ranks 1--2, the median pair, and ranks 11--12 (\cref{tab:extra_composite_drift}). The composite retains untethered hallucinations inside active regions (a dark sand patch, a street car, a horizon structure) and fades the burn-barrel flame at $230\,\text{s}$, whereas \Ours hallucinates a storm-drain object at $60\,\text{s}$.}
 \label{fig:st_composite_frames}
\end{figure*}

\paragraph{Moving-region drift of the output composite.}
Prior to generating $240\,\text{s}$ untethered and composite rollouts on the 12 horizon scenes, we pre-specified a moving-region color-drift test measuring the encoded-Lab distance between a region's late-window mean color (final $20\,\text{s}$) and its post-warmup baseline (frames $25\text{--}48$, about $1.6\text{--}3.0\,\text{s}$ at $16\,\text{fps}$) under \Ours's frozen map, subsequently extending the identical metric to all 48 $60\,\text{s}$ rollouts.
\Cref{tab:extra_composite_drift} and \cref{fig:st_composite_frames} show that post-hoc compositing inherits the untethered rollout's color drift inside moving ($\alpha\le0.1$) and untouched ($\alpha<0.001$) pixels, whereas \Ours shows lower moving-region color drift at similar ($60\,\text{s}$) or higher ($240\,\text{s}$) temporal variation.
These regions derive from the feathered blend map, not from independent material annotations, and stable mean color does not by itself establish physically correct motion.

\begin{table*}[t]
 \centering\scriptsize\setlength{\tabcolsep}{3pt}
 \caption{\textbf{Regional color drift and edge sharpness for the post-hoc composite vs.\ \Ours} (seed 42; 48 scenes at $60\,\text{s}$, 12 pre-selected scenes at $240\,\text{s}$). Drift measures encoded-Lab shift from frames $25\text{--}48$ to the final $20\,\text{s}$ across frozen-map partitions: moving ($\alpha\le0.1$), untouched ($\alpha<0.001$), and anchored ($\alpha\ge0.9$; 47 and 11 scenes). $\Delta$ reports row minus \Ours with 95\% scene-bootstrap intervals and scene win/loss counts; edge excess compares $1\,\text{s}$-mean Sobel magnitude against the source.}
 \label{tab:extra_composite_drift}
 \begin{tabular}{@{}llrlrlrrr@{}}\toprule
 & & \multicolumn{2}{c}{Moving region} & \multicolumn{2}{c}{Pixels the composite leaves untouched} & & \multicolumn{2}{c}{Anchored region}\\\cmidrule(lr){3-4}\cmidrule(lr){5-6}\cmidrule(lr){8-9}
Horizon & Method & Drift $\downarrow$ & $\Delta$ vs.\ ours [95\% CI] (worse/better) & Drift $\downarrow$ & $\Delta$ vs.\ ours [95\% CI] (worse/better) & \shortstack{Moving\\variation} & Drift & \shortstack{Edge\\excess}\\\midrule
60\,s ($n=48$) & \OursLabel & 4.95 & -- & 7.59 & -- & 13.77 & 1.01 & -16.90\\
 & Full coverage, no tether & 9.07 & +4.12 [+2.12, +6.47] (37/11) & 12.52 & +4.93 [+2.25, +8.14] (37/11) & 14.05 & 8.27 & -24.49\\
 & Composite (post-hoc) & 9.03 & +4.09 [+2.09, +6.42] (37/11) & 12.55 & +4.95 [+2.26, +8.17] (37/11) & 13.86 & 0.34 & -1.89\\
\midrule
240\,s ($n=12$) & \OursLabel & 6.27 & -- & 7.68 & -- & 13.76 & 1.06 & -18.51\\
 & Full coverage, no tether & 11.48 & +5.21 [+2.35, +9.27] (10/2) & 12.75 & +5.06 [+1.51, +9.87] (9/3) & 12.20 & 7.59 & -24.07\\
 & Composite (post-hoc) & 11.39 & +5.12 [+2.29, +9.16] (10/2) & 12.74 & +5.06 [+1.51, +9.87] (9/3) & 12.02 & 0.22 & -1.88\\
\bottomrule\end{tabular}

\end{table*}

\paragraph{Sharpness of the anchored region.}
Evaluating strongly anchored pixels ($\alpha\ge0.9$, \cref{tab:st_sharpness}) separates per-frame spatial detail from temporal steadiness.
On individual frames, \Ours retains $0.81\times$ of the source's Sobel gradient magnitude, below the pixel-pasted composite ($0.98\times$), framewise host ($0.92\times$), and untethered run ($0.90\times$), and similar to age decay ($0.84\times$); anchored detail is therefore softer per frame.
After averaging frames over $1\,\text{s}$ windows, \Ours has a smaller gradient deficit than every generative reference at $60\,\text{s}$ ($-16.9$ vs.\ $-20.9$ to $-34.5$) and lower temporal variation ($3.8$ vs.\ $6.5\text{--}8.0$ gray levels). One outlier scene inflates the framewise host's $240\,\text{s}$ edge-ratio mean to $1.29$ (interval $[0.79, 2.25]$).
Because source PSNR, edge sharpness, and temporal variation measure different properties---and age decay has similarly low moving-region mean-color drift ($4.7$ vs.\ $4.9$ at $60\,\text{s}$)---these outcomes should be interpreted jointly rather than as a single realism ranking.

\begin{table*}[t]
 \centering\small\setlength{\tabcolsep}{3pt}
 \caption{\textbf{Anchored-region ($\alpha\ge0.9$) sharpness and temporal variation across methods} (seed 42, 47 benchmark scenes with an anchored region; `--' marks $240\,\text{s}$ untethered and composite runs, which were generated only on the 12 horizon scenes in \cref{tab:extra_composite_drift}). Edge ratio divides late-window per-frame Sobel magnitude by the source ($1.0=$ source sharpness); edge excess measures $1\,\text{s}$ window-mean Sobel difference from the source (\cref{tab:extra_composite_drift}).}
 \label{tab:st_sharpness}
 \begin{tabular}{@{}lcccccccc@{}}\toprule
 & \multicolumn{4}{c}{60\,s (47 sources)} & \multicolumn{4}{c}{240\,s (47 sources)}\\
\cmidrule(lr){2-5}\cmidrule(lr){6-9}
Method & Edge ratio & \shortstack{Edge\\excess} & Variation & \shortstack{Moving\\drift} & Edge ratio & \shortstack{Edge\\excess} & Variation & \shortstack{Moving\\drift}\\\midrule
\OursLabel & 0.81 [0.79, 0.84] & $-$16.9 & 3.8 & 4.9 & 0.81 [0.79, 0.84] & $-$17.0 & 3.8 & 5.1\\
Full coverage, no tether & 0.90 [0.84, 0.96] & $-$24.5 & 8.0 & 9.1 & -- & -- & -- & --\\
Composite (post-hoc) & 0.98 [0.97, 0.99] & $-$1.9 & 0.2 & 9.0 & -- & -- & -- & --\\
Framewise Echo, $k=1$ & 0.92 [0.83, 1.05] & $-$20.9 & 6.5 & 11.2 & 1.29 [0.79, 2.25] & $-$18.7 & 6.1 & 14.0\\
Age-decay read & 0.84 [0.78, 0.92] & $-$25.7 & 7.5 & 4.7 & 0.84 [0.78, 0.92] & $-$25.7 & 7.6 & 4.9\\
Echo-Forcing port, $k=3$ & 0.85 [0.69, 1.07] & $-$34.5 & 6.8 & 7.4 & 0.68 [0.54, 0.86] & $-$45.1 & 6.1 & 11.3\\
\bottomrule\end{tabular}

\end{table*}
\paragraph{Sensitivity to band strength and warm-up duration.}
One-factor changes at $60\,\text{s}$ (\cref{tab:extra_sensitivity}) show higher PSNR when the transition-band strength is increased to $\lambda_b=1$ or warm-up is shortened to three latents, and lower PSNR when $\lambda_b=0$ or warm-up is extended to twelve latents (reported intervals exclude zero). Measured activity differences remain inconclusive; these comparisons do not establish optimality of the locked defaults.
We retain the locked defaults ($\lambda_b=0.5$, six latents) without post-hoc tuning on test scenes.

\paragraph{Sensitivity to ramp width, history decay, and map freezing.}
Single-factor sweeps in \cref{tab:extra_sensitivity_review} trace the remaining design parameters.
Ramp width $\eta$ trades PSNR against color error: $\eta=0.1$ shifts PSNR by \stres{sr_eta010_dpsnr}\,\text{dB} [\stres{sr_eta010_dpsnr_lo}, \stres{sr_eta010_dpsnr_hi}] while raising color error by \stres{sr_eta010_dcol}, whereas $\eta=0.5$ and $1.0$ shift PSNR by \stres{sr_eta050_dpsnr} and \stres{sr_eta100_dpsnr}\,\text{dB} and color error by \stres{sr_eta050_dcol} and \stres{sr_eta100_dcol}; these retrospective sweeps show a trade-off and do not establish that the locked $\eta=0.3$ is optimal for either metric.
Setting older-history decay to $\rho=0.1$ shifts PSNR by \stres{sr_rho010_dpsnr}\,\text{dB} [\stres{sr_rho010_dpsnr_lo}, \stres{sr_rho010_dpsnr_hi}], and $\rho=0.5$ yields \stres{sr_rho050_dpsnr}\,\text{dB} [\stres{sr_rho050_dpsnr_lo}, \stres{sr_rho050_dpsnr_hi}], with all $\eta$ and $\rho$ flow intervals spanning $1.0$.
Replacing the piecewise band with a continuous ramp shifts PSNR by \stres{sr_ramp_dpsnr}\,\text{dB} [\stres{sr_ramp_dpsnr_lo}, \stres{sr_ramp_dpsnr_hi}] while reducing activity to \stres{sr_ramp_ratio}$\times$ [\stres{sr_ramp_rlo}, \stres{sr_ramp_rhi}], and re-estimating the map every $5\,\text{s}$ changes PSNR by \stres{sr_maprefresh_dpsnr}\,\text{dB} [\stres{sr_maprefresh_dpsnr_lo}, \stres{sr_maprefresh_dpsnr_hi}] and raises color error by \stres{sr_maprefresh_dcol}.
\begin{table*}[t]
 \centering\small
 \caption{\textbf{Single-factor sensitivity to transition-band strength $\lambda_b$ and warm-up length at $60\,\text{s}$} (seed 42, 42 appearance and 40 activity scenes; $\Delta\text{PSNR}$ reports arm minus locked \Ours and flow ratio reports arm over locked \Ours, with 95\% scene-bootstrap intervals).}
 \label{tab:extra_sensitivity}
 \begin{tabular}{@{}lrrrrr@{}}\toprule
Arm & PSNR $\uparrow$ & $\Delta$PSNR vs.\ locked [95\% CI] & Color $\downarrow$ & Activity & Activity ratio [95\% CI]\\\midrule
SceneTether ($\lambda_b=0.5$, warm-up 6) & 22.37 & -- & 5.57 & 1.97 & --\\
Band $\lambda_b=0$ & 21.12 & -1.25 [-1.65, -0.86] & 5.86 & 1.93 & 0.98 [0.86, 1.16]\\
Band $\lambda_b=1$ & 23.42 & +1.05 [+0.60, +1.61] & 5.70 & 1.87 & 0.95 [0.84, 1.06]\\
Warm-up 3 latents & 23.20 & +0.83 [+0.39, +1.33] & 5.43 & 1.92 & 0.97 [0.85, 1.15]\\
Warm-up 12 latents & 20.75 & -1.62 [-2.17, -1.09] & 5.55 & 1.96 & 0.99 [0.87, 1.14]\\
\bottomrule\end{tabular}

\end{table*}

\section{Samples on other backbones}
\label{fm:suppl_backbones}
\begin{figure*}[t]
 \centering\raisebox{-224.29100pt}{\includegraphics{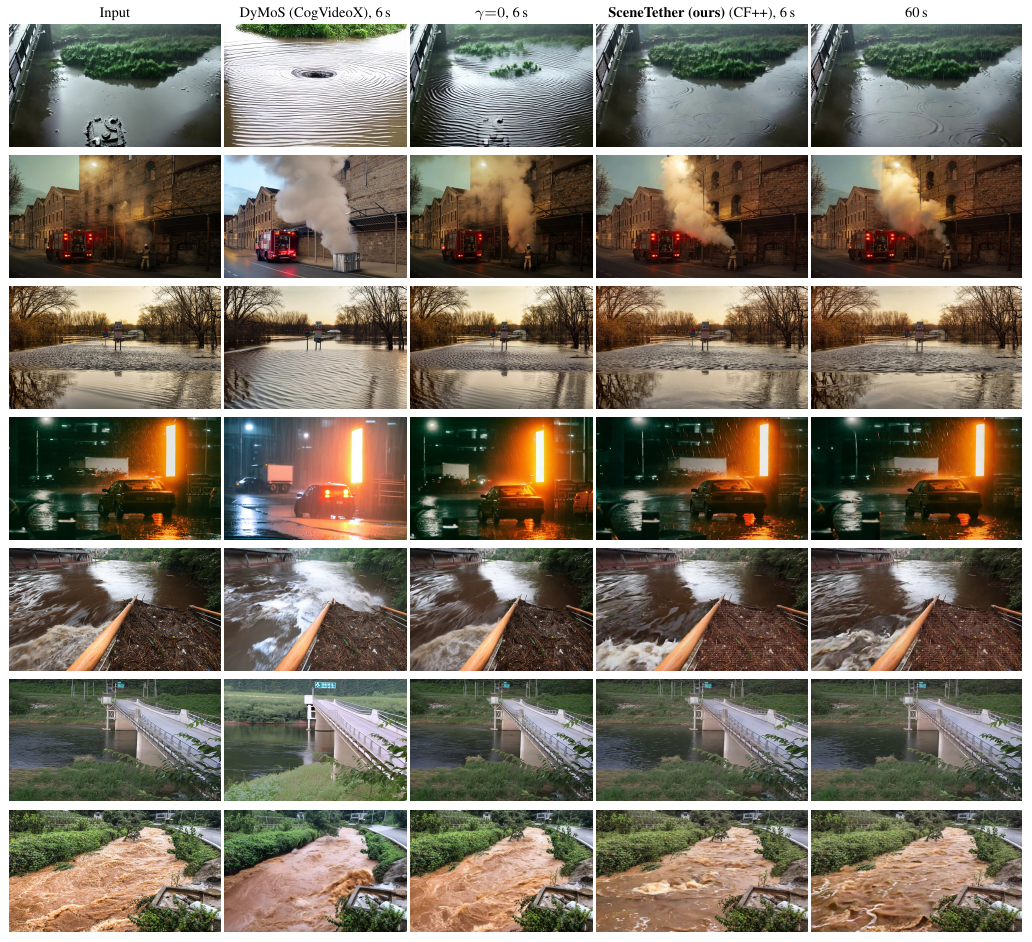}}
 \caption{\textbf{Short native CogVideoX-5B-I2V reference} showing official DyMoS ($\gamma=0.8$ and $\gamma=0$, frame 49 at $6\,\text{s}$) beside \Ours on the CF++ student at $6$ and $60\,\text{s}$ across all seven development scenes (seed 42, unscored).}
 \label{fig:st_cog}
\end{figure*}

\begin{figure*}[t]
 \centering\raisebox{-213.14629pt}{\includegraphics{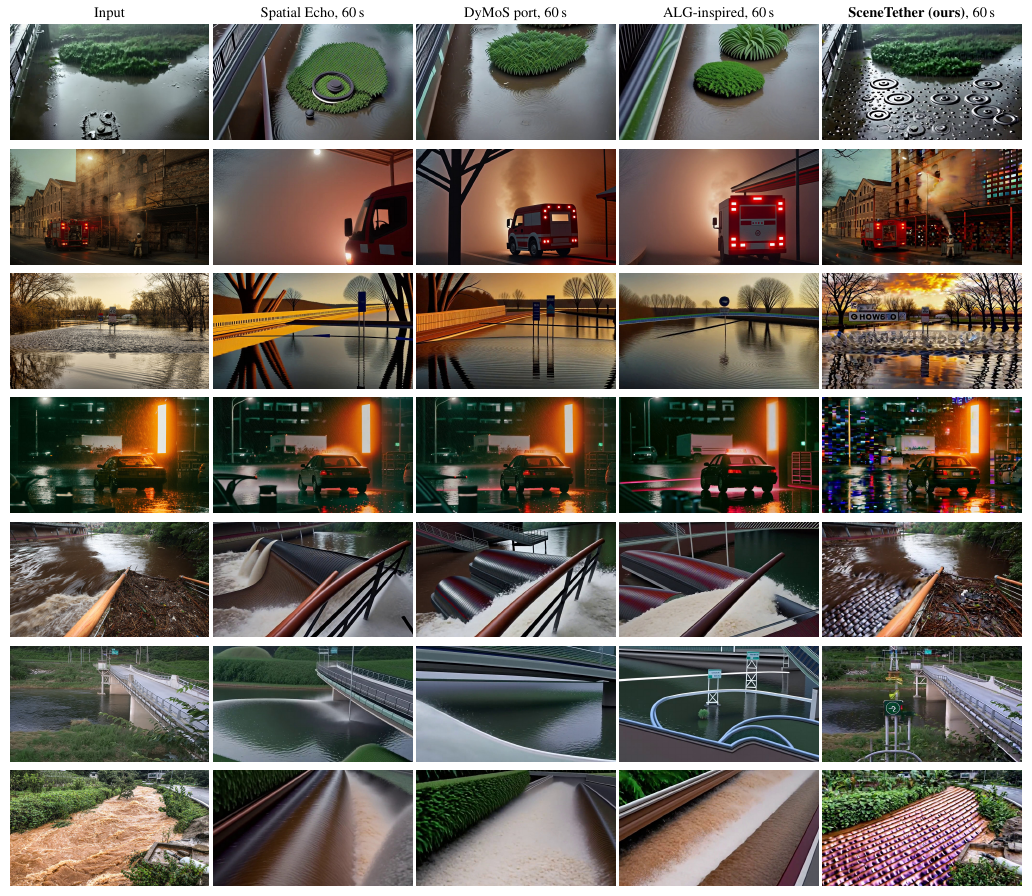}}
 \caption{\textbf{Exploratory transfer to Self-Forcing ($k=3$)} on seven development scenes at $60\,\text{s}$ (seed 42, unchanged CF++-selected settings). \Ours often reduces viewpoint changes seen with the adapted references but introduces texture tiling, block glitches, and object hallucinations.}
 \label{fig:st_sf}
\end{figure*}

\begin{figure*}[t]
 \centering\raisebox{-213.14629pt}{\includegraphics{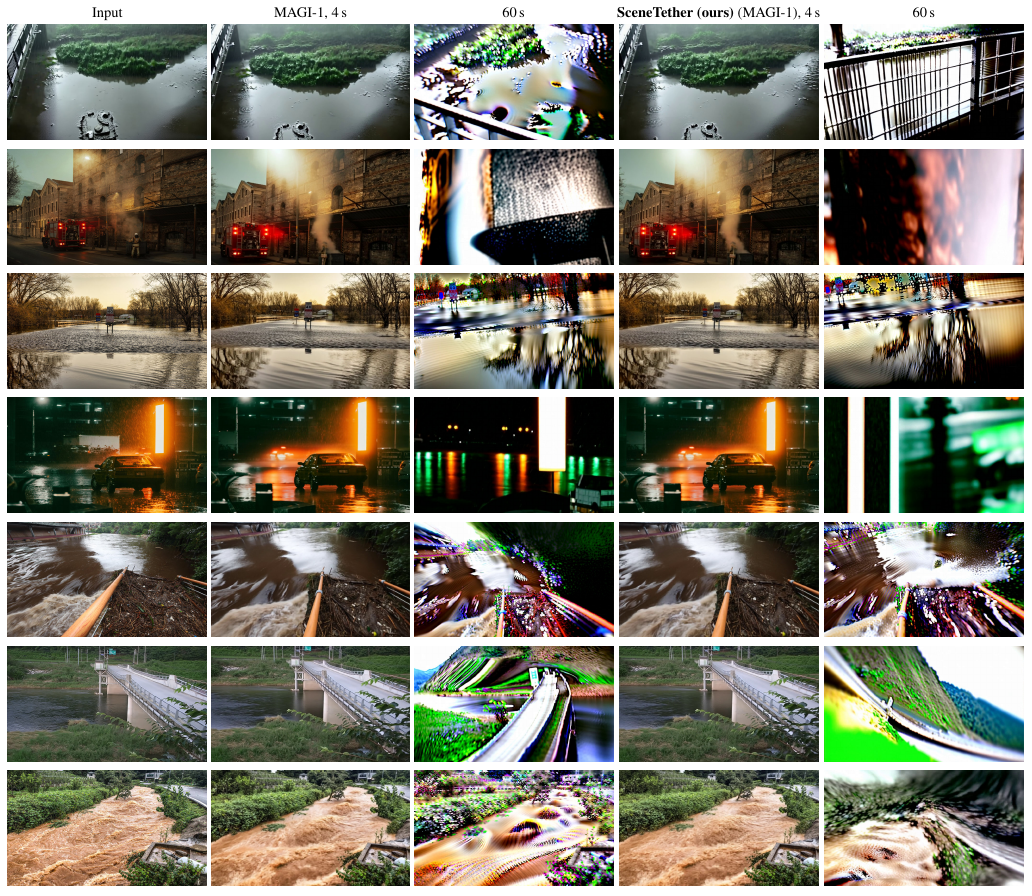}}
 \caption{\textbf{Exploratory port to MAGI-1 4.5B-distill (16 steps)} at $4\,\text{s}$ and $60\,\text{s}$ across all seven development scenes (seed 42, unscored), where neither native caching nor untuned value tethering prevents $60\,\text{s}$ drift.}
 \label{fig:st_magi}
\end{figure*}

\Cref{fig:st_cog,fig:st_sf,fig:st_magi} display every scene of the seven-source development set at seed 42 and native resolution; showing the complete set avoids selecting favorable cases, although these scenes are not an independent transfer test.

\paragraph{Short native I2V reference (CogVideoX-5B-I2V).}
Running official DyMoS on CogVideoX-5B-I2V~\cite{yang2025cogvideox} ($49$ frames at $8\,\text{fps}$, $720\times480$, 50 steps, CFG 6; \cref{fig:st_cog}) changes the camera viewpoint within $6\,\text{s}$ on four of seven scenes at the default $\gamma=0.8$, whereas $\gamma=0$ preserves more of the source layout; \Ours is shown at $6$ and $60\,\text{s}$ on its own backbone for reference.

\paragraph{Second causal student (Self-Forcing).}
We evaluate spatial Echo-Forcing, DyMoS, ALG-inspired ports, and untuned \Ours on Self-Forcing~\cite{huang2025selfforcing} at its native $k=3$ block size, tethering the newest history latent of each block while decaying older latents (\cref{fig:st_sf,tab:st_sf_scored}).
In the seven illustrated development scenes at $60\,\text{s}$, adapted references frequently zoom into altered content, whereas \Ours more often preserves global layout but can introduce tiled water textures, colored window-light glitches, and invented objects; the night-rain example is a clear negative case. These observations are qualitative and limited to these scenes.
Across all 48 benchmark scenes (\cref{tab:st_sf_scored}), \Ours raises PSNR over its Self-Forcing Echo host by \stres{sf_dpsnr}\,\text{dB} [\stres{sf_dpsnr_lo}, \stres{sf_dpsnr_hi}] (winning on \stres{sf_wins}/\stres{sf_n} scenes; the corresponding CF++ gain over its host is \stres{cf_dpsnr}\,\text{dB}) and exceeds the three adapted ports (\stres{sf_port_psnr_min}--\stres{sf_port_psnr_max}\,\text{dB}) by \stres{sf_port_dpsnr_min}--\stres{sf_port_dpsnr_max}\,\text{dB}, but retains only \stres{sf_ratio}$\times$ [\stres{sf_ratio_lo}, \stres{sf_ratio_hi}] of host activity (vs.\ \stres{cf_ratio}$\times$ [\stres{cf_ratio_lo}, \stres{cf_ratio_hi}] on CF++; \stres{sf_port_ratio_min}--\stres{sf_port_ratio_max}$\times$ vs.\ ports) and trails even the untethered CF++ host in absolute PSNR (\stres{sf_psnr_st} vs.\ \stres{cf_psnr_host}\,\text{dB}), showing that the CF++ fidelity--activity operating point does not transfer unchanged; reduced flow alone cannot distinguish suppressed intended motion from removed drift or artifacts.

\paragraph{Image-trained causal model (MAGI-1).}
We also port value tethering to MAGI-1 4.5B-distill~\cite{ai2025magi1autoregressivevideogeneration} ($832\times480$, 16 steps; \cref{fig:st_magi}), which denoises six-latent chunks over a four-chunk sliding KV window without a refresh pass, Echo host, pinned source, or block-relative RoPE.
The port estimates the map from the first six generated latents and interpolates the newest clean latent's values toward cached $z_0$ values, which remain referenced for up to four subsequent chunks after the source leaves the active window. It resembles native MAGI-1 at $4\,\text{s}$ but does not prevent progressive zoom, oversaturation, and texture breakdown by $60\,\text{s}$. This exploratory port lacks the source pinning and refresh of the CF++ setting and does not isolate value tethering.

\begin{table*}[t]
 \centering\small\setlength{\tabcolsep}{3pt}
 \caption{\textbf{Four-minute system comparison with native SVI 2.0 Pro} (Wan2.2-I2V-A14B, 50 steps, CFG 4; 48 chained 81-frame clips with SVI's 5-frame overlap give 3{,}653 frames ($243.5\,\text{s}$ at $15\,\text{fps}$), trimmed at $240\,\text{s}$; the $[0,60]$, $[60,120]$, and $[180,240]\,\text{s}$ windows are resampled to $16\,\text{fps}$ by nearest timestamp) versus CF++ rows (seed 42) across nine category-stratified test sources (eight eligible after snowfall abstention; system-level comparison across distinct backbones, samplers, and compute budgets).}
 \label{tab:st_native_long}
 \begin{tabular}{@{}lccccccccc@{}}\toprule
 & \multicolumn{3}{c}{Source-support PSNR $\uparrow$} & \multicolumn{3}{c}{Mean-Lab color error $\downarrow$} & \multicolumn{3}{c}{MCFF-late (px/pair)}\\
\cmidrule(lr){2-4}\cmidrule(lr){5-7}\cmidrule(lr){8-10}
Window end & 60\,s & 120\,s & 240\,s & 60\,s & 120\,s & 240\,s & 60\,s & 120\,s & 240\,s\\\midrule
SVI 2.0 Pro (Wan2.2-A14B, native) & 15.53 & 12.40 & 11.55 & 28.34 & 33.61 & 35.92 & 1.51 & 0.94 & 0.86\\
Framewise Echo, $k=1$ & 16.74 & 17.99 & 17.59 & 13.98 & 10.59 & 12.05 & 0.93 & 0.68 & 0.78\\
Age-decay read & 17.57 & 17.67 & 17.30 & 6.66 & 6.05 & 6.56 & 0.75 & 0.60 & 0.77\\
Echo-Forcing port, $k=3$ & 16.97 & 16.16 & 15.61 & 14.71 & 18.97 & 20.68 & 0.51 & 0.44 & 0.30\\
\OursLabel & 21.96 & 21.90 & 21.87 & 5.57 & 5.93 & 5.65 & 1.01 & 0.90 & 1.07\\
\bottomrule\end{tabular}

\end{table*}

\begin{table*}[t]
 \centering\small
 \caption{\textbf{Benchmark evaluation on Self-Forcing ($k=3$) vs.\ CF++ ($k=1$) at $60\,\text{s}$} across 48 test scenes (seed 42; untuned CF++ hyperparameters; \stres{sf_n} appearance, \stres{sf_nmot} MCFF-late activity, and 48 VBench-I2V scenes).}
 \label{tab:st_sf_scored}
 \begin{tabular}{@{}llccccc@{}}\toprule
Student & Method & PSNR $\uparrow$ & Color $\downarrow$ & \shortstack{MCFF-late\\(px/pair)} & \shortstack{I2V\\subj. $\uparrow$} & \shortstack{I2V\\bg. $\uparrow$}\\\midrule
CF++ & Framewise Echo, $k=1$ & 16.97 & 15.54 & 1.02 & 0.981 & 0.971\\
 & Echo-Forcing port, $k=3$ & 16.88 & 15.08 & 0.50 & 0.970 & 0.961\\
 & DyMoS port, $k=1$ & 16.89 & 16.95 & 0.97 & 0.979 & 0.969\\
 & ALG-inspired, $k=1$ & 15.87 & 19.96 & 1.07 & 0.977 & 0.969\\
 & \OursLabel & 22.37 & 5.57 & 1.97 & 0.979 & 0.969\\
\midrule
Self-Forcing & Framewise Echo, $k=3$ & 11.68 & 37.51 & 1.33 & 0.947 & 0.943\\
 & Echo-Forcing port, $k=3$ & 11.72 & 35.24 & 1.36 & 0.949 & 0.945\\
 & DyMoS port, $k=3$ & 11.88 & 34.33 & 1.80 & 0.945 & 0.944\\
 & ALG-inspired, $k=3$ & 11.46 & 38.04 & 1.40 & 0.932 & 0.933\\
 & \OursLabel & 16.37 & 12.71 & 0.43 & 0.963 & 0.954\\
\bottomrule\end{tabular}

\end{table*}

\begin{table*}[t]
 \centering\small\setlength{\tabcolsep}{5pt}
 \caption{\textbf{ALG-inspired and DyMoS ports with continuous source pinning at $60\,\text{s}$} (48 test scenes, seed 42; \stres{pp_n} appearance and \stres{pp_nmot} activity scenes; paired scene-bootstrap $\Delta\text{PSNR}$ and flow ratios compare each pinned variant against its standard rotating-anchor counterpart).}
 \label{tab:st_pinned_ports}
 \begin{tabular}{@{}lccccc@{}}\toprule
Method & PSNR $\uparrow$ & \shortstack{$\Delta$PSNR\\{}[95\% CI]} & Color $\downarrow$ & \shortstack{MCFF-late\\(px/pair)} & \shortstack{Activity ratio\\{}[95\% CI]}\\\midrule
ALG-inspired, $k=1$ & 15.87 & -- & 19.96 & 1.07 & --\\
\quad + pinned source & 14.23 & $-$1.65 [$-$2.06, $-$1.25] & 24.70 & 1.88 & 1.76 [1.22, 2.47]\\
\midrule
DyMoS port, $k=1$ & 16.89 & -- & 16.95 & 0.97 & --\\
\quad + pinned source & 14.99 & $-$1.90 [$-$2.59, $-$1.22] & 19.23 & 1.02 & 1.04 [0.69, 1.51]\\
\midrule
\OursLabel & 22.37 & -- & 5.57 & 1.97 & --\\
\bottomrule\end{tabular}

\end{table*}

\section{Source fidelity, realism, and future benchmarks}
\label{fm:suppl_realism}

\begin{figure*}[t]
 \centering\raisebox{-197.02069pt}{\includegraphics{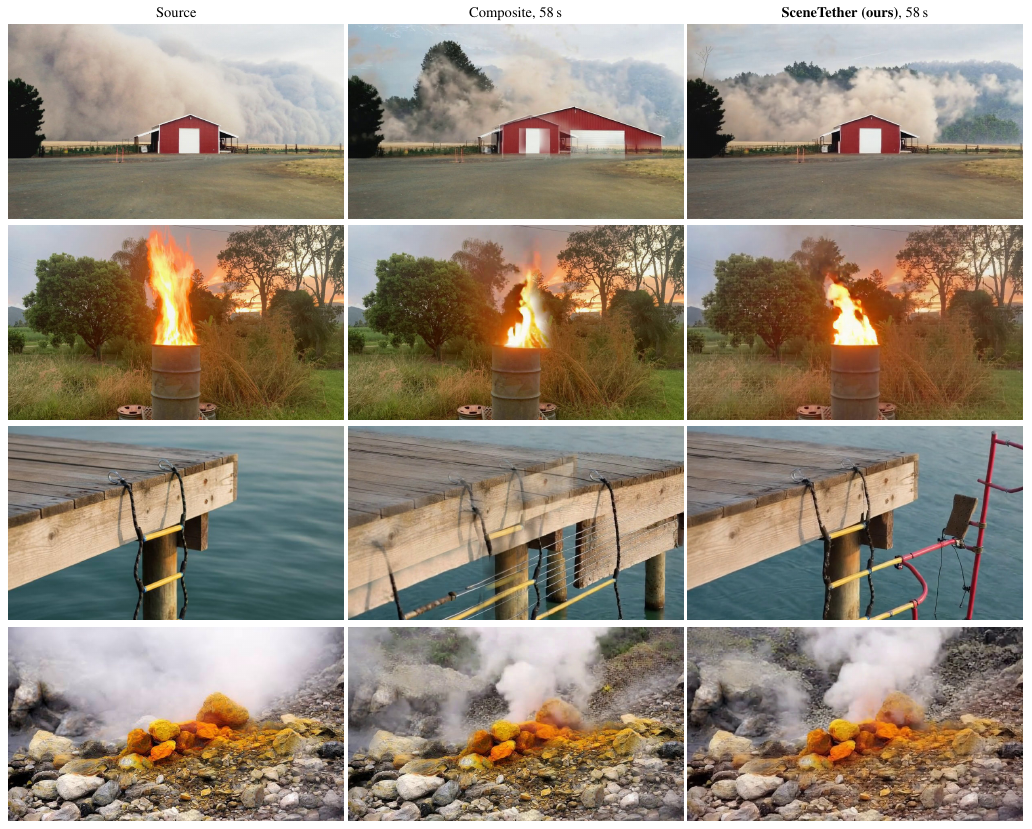}}
 \caption{\textbf{Illustrative compositing-boundary failures at $58\,\text{s}$} (seed 42; four of the six inspected cases, at the benchmark evaluation resolution). Feathering the source against an untethered frame can double structures or clip flames and steam near mask edges. \Ours does not have these particular paste seams, but invents a red ladder in row~3.}
 \label{fig:st_composite_boundary}
\end{figure*}

\paragraph{Static fidelity of the composite versus perceived quality.}
Pixel pasting (\cref{st:completed}) reaches similar or higher static-support PSNR point estimates, while reducing anchored-region temporal variation to \stres{comp_still_flicker} vs.\ \stres{st_still_flicker} gray levels and retaining greater per-frame edge magnitude ($0.98$ vs.\ $0.81$ of source; \cref{tab:st_sharpness}).
Cinemagraph and image-animation methods model motion explicitly~\cite{holynski2021animating,mahapatra2022controllable,mahapatra2023text2cinemagraph}; our post-hoc blend is a diagnostic comparator, not a reproduction of them. Copying the source suits truly rigid masonry or rock but cannot generate the subtle variation of foliage, grass, and calm water~\cite{Li_2024_gen_image_dynamics} in fully copied regions. \Ours retains more measured variation in anchored regions; this alone does not establish physically correct ambient motion.
In the manually inspected boundary cases, feather-blending static source pixels with drifted untethered output can create visible seams (five of six inspected scenes at $58\,\text{s}$; four are shown in \cref{fig:st_composite_boundary}): doubled roofing, clipped flames, ghosted dock pilings, and truncated steam. \Ours avoids these illustrated paste seams but still hallucinates a red dock ladder in row~3; this inspection is qualitative, not an exhaustive incidence estimate.
In our matched-encoding full-minute study (Tab.~3 of the main paper; \stres{hs_m_raters} raters, \stres{hs_m_votes} votes), preference estimates favor \Ours over the composite for both scene preservation (\stres{hs_m_scene}\%) and motion naturalness (\stres{hs_m_motion}\%), a descriptive result given the small rater pool. PSNR, Lab drift, and flow magnitude measure preservation and motion amplitude, not overall visual realism.

\paragraph{Future benchmarks and applications.}
Future long-animation benchmarks should jointly evaluate rigid structural stability, subtle ambient dynamics in nominally static vegetation, and physically plausible progression (versus looping, boiling, or random texture churn) using human-validated masks and real long-take reference videos.
Beyond nature photography, read-time value tethering may suit ambient displays and cinemagraph editing as native image-conditioned causal backbones mature, but our transfer experiments (\cref{fm:suppl_backbones}) show that each architecture needs its own validation. Generated dynamics are not a substitute for simulation or observation in scientific use.

\section{The two components, step by step}
\label{fm:suppl_components}
\begin{figure*}[t]
 \centering\includegraphics{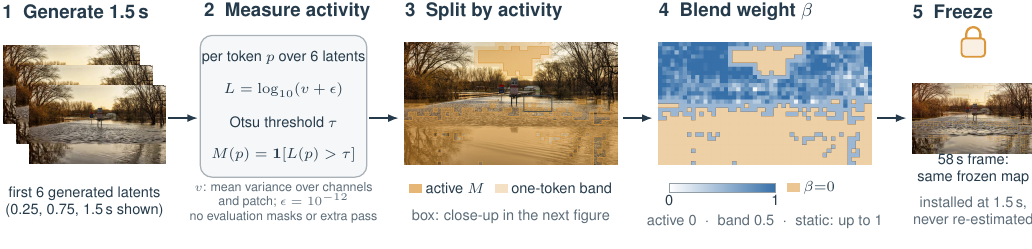}
 \caption{\textbf{Component 1: one-time activity map estimation and freezing.} From the first six generated latents ($1.5\,\text{s}$), \Ours computes $30\times52$ patch log-variances ($16\times16\,\text{px}$ per token), applies Otsu's threshold, and freezes $\beta$ ($\beta=0$ on active tokens, $0.5$ on the one-token boundary ring, and up to $1.0$ in the static interior) through $58\,\text{s}$ (development scene, seed 42).}
 \label{fig:st_comp_map}
\end{figure*}

\begin{figure*}[t]
 \centering
\begingroup
\input{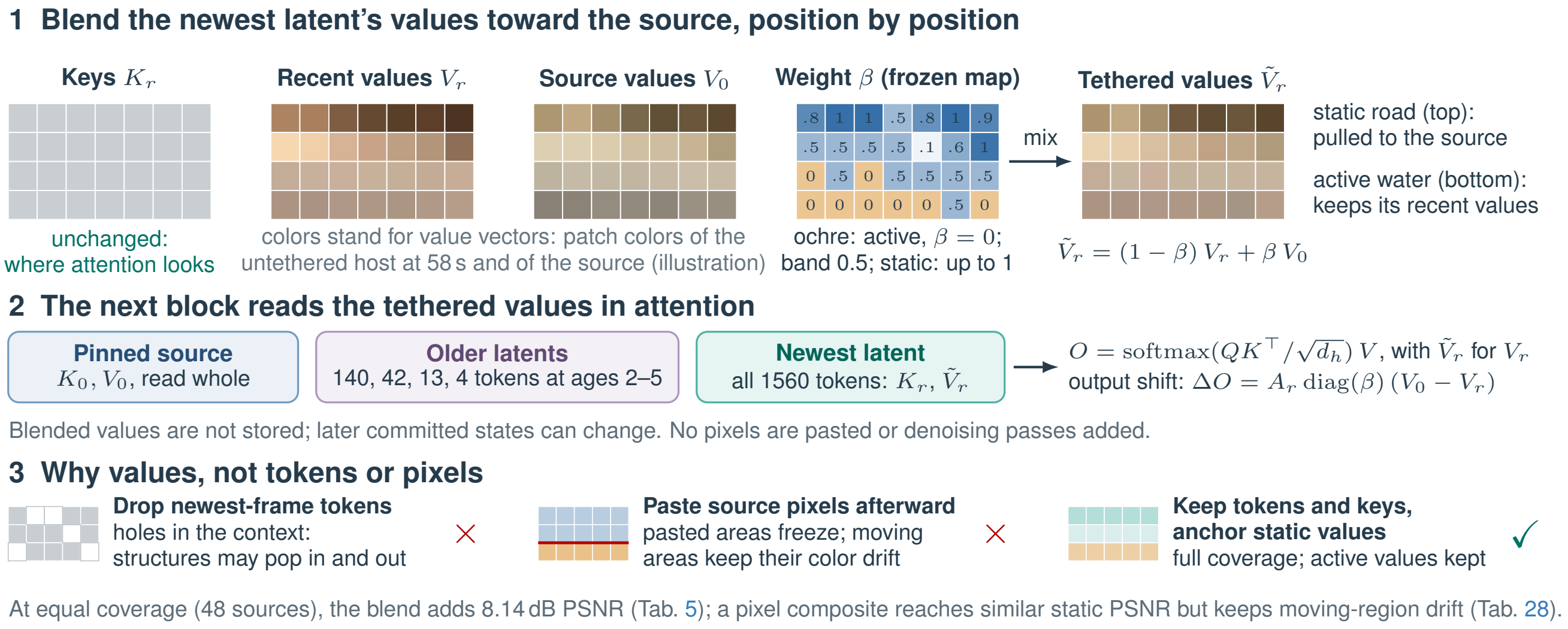}%
\begin{tikzpicture}[x=1cm,y=1cm,font=\sffamily\scriptsize,
 arr/.style={-{Latex[length=1.8mm]},draw=stInk,line width=.8pt},
 stp/.style={font=\sffamily\bfseries\small,text=stInk,anchor=west,inner sep=0pt},
 note/.style={font=\sffamily\fontsize{7}{8.2}\selectfont,text=stInk,align=center,inner sep=0pt},
 lab/.style={font=\sffamily\bfseries\fontsize{7.3}{8.5}\selectfont,text=stInk,align=center,inner sep=0pt}]
\def\c{.32}
\path[use as bounding box] (0,-3.05) rectangle (17.4,3.95);
\node[stp] at (0,3.75) {1\;\;Blend the newest latent's values toward the source, position by position};
\node[lab] at (1.12,3.12) {Keys $K_r$};
\foreach \i in {0,...,3}{\foreach \j in {0,...,6}{\filldraw[fill=stInk!25,draw=white,line width=.35pt] ({\j*\c},{2.85-\i*\c}) rectangle ++(\c,-\c);}}
\node[note,text=stTeal!80!black] at (1.12,1.22) {unchanged:\\where attention looks};
\node[lab] at (4.02,3.12) {Recent values $V_r$};
\foreach \i/\j/\b/\m in \tpcells {\filldraw[fill=tpr\i\j,draw=white,line width=.35pt] ({2.9+\j*\c},{2.85-\i*\c}) rectangle ++(\c,-\c);}
\node[lab] at (6.92,3.12) {Source values $V_0$};
\foreach \i/\j/\b/\m in \tpcells {\filldraw[fill=tps\i\j,draw=white,line width=.35pt] ({5.8+\j*\c},{2.85-\i*\c}) rectangle ++(\c,-\c);}
\node[note,text=stInk!70] at (5.47,1.22) {colors stand for value vectors: patch colors of the\\untethered host at 58\,s and of the source (illustration)};
\node[lab] at (9.82,3.12) {Weight $\beta$ (frozen map)};
\foreach \i/\j/\b/\m in \tpcells {
 \pgfmathsetmacro{\pc}{100*\b}
 \ifnum\m=1 \filldraw[fill=stOchre!55,draw=white,line width=.35pt] ({8.7+\j*\c},{2.85-\i*\c}) rectangle ++(\c,-\c);
 \else \filldraw[fill=stBlue!\pc!white,draw=white,line width=.35pt] ({8.7+\j*\c},{2.85-\i*\c}) rectangle ++(\c,-\c); \fi
 \pgfmathsetmacro{\bt}{round(10*\b)/10}
 \node[font=\sffamily\fontsize{4.8}{5}\selectfont,text=stInk] at ({8.7+\j*\c+\c/2},{2.85-\i*\c-\c/2}) {\pgfmathprintnumber[fixed,precision=1,skip 0.]{\bt}};}
\node[note] at (9.82,1.22) {ochre: active, $\beta=0$;\\band 0.5; static: up to 1};
\draw[arr] (11.05,2.21)--(11.75,2.21);
\node[note] at (11.4,2.48) {mix};
\node[lab] at (12.97,3.12) {Tethered values $\tilde V_r$};
\foreach \i/\j/\b/\m in \tpcells {\filldraw[fill=tpt\i\j,draw=white,line width=.35pt] ({11.85+\j*\c},{2.85-\i*\c}) rectangle ++(\c,-\c);}
\node[note] at (12.97,1.22) {$\tilde V_r=(1-\beta)\,V_r+\beta\,V_0$};
\node[note,align=left,anchor=west] at (14.4,2.6) {static road (top):\\pulled to the source};
\node[note,align=left,anchor=west] at (14.4,1.85) {active water (bottom):\\keeps its recent values};
\node[stp] at (0,.62) {2\;\;The next block reads the tethered values in attention};
\foreach \x/\w/\col/\t/\s in {0/3.2/stBlue/Pinned source/{$K_0,V_0$, read whole},3.4/4.0/stPurple/Older latents/{140, 42, 13, 4 tokens at ages 2--5},7.6/3.4/stTeal/Newest latent/{all 1560 tokens: $K_r$, $\tilde V_r$}}{
 \filldraw[rounded corners=1.2mm,fill=\col!10,draw=\col!70,line width=.6pt] (\x,-.45) rectangle ++(\w,.78);
 \node[lab,text=\col!70!black] at ({\x+\w/2},.1) {\t};
 \node[note] at ({\x+\w/2},-.2) {\s};}
\draw[arr] (11.1,-.06)--(11.6,-.06);
\node[note,align=left,anchor=west] at (11.7,.12) {$O=\operatorname{softmax}(QK^\top/\sqrt{d_h})\,V$, with $\tilde V_r$ for $V_r$};
\node[note,align=left,anchor=west] at (11.7,-.24) {output shift: $\Delta O=A_r\,\mathrm{diag}(\beta)\,(V_0-V_r)$};
\node[note,align=left,anchor=west,text=stInk!75] at (0,-.78) {Blended values are not stored; later committed states can change. No pixels are pasted or denoising passes added.};
\node[stp] at (0,-1.25) {3\;\;Why values, not tokens or pixels};
\newcommand{\minigrid}[2]{\foreach \i in {0,...,2}{\foreach \j in {0,...,4}{\filldraw[fill=#2,draw=white,line width=.3pt] ($(#1)+({\j*.2},{-\i*.2})$) rectangle ++(.2,-.2);}}}
\minigrid{0,-1.6}{stInk!25}
\foreach \p in {(.2,-1.6),(.6,-1.8),(0,-2.0),(.8,-2.0),(.4,-1.6)}{\filldraw[fill=white,draw=stInk!30,line width=.3pt] \p rectangle ++(.2,-.2);}
\node[note,align=left,anchor=west] at (1.15,-1.9) {\textbf{Drop newest-frame tokens}\\holes in the context:\\structures may pop in and out};
\node[text=red!70!black,font=\bfseries\large] at (5.05,-1.9) {$\times$};
\minigrid{5.85,-1.6}{stBlue!35}
\foreach \j in {0,...,4}{\filldraw[fill=stOchre!60,draw=white,line width=.3pt] ({5.85+\j*.2},-2.0) rectangle ++(.2,-.2);}
\draw[red!70!black,line width=.9pt] (5.85,-2.0)--(6.85,-2.0);
\node[note,align=left,anchor=west] at (7.0,-1.9) {\textbf{Paste source pixels afterward}\\pasted areas freeze; moving\\areas keep their color drift};
\node[text=red!70!black,font=\bfseries\large] at (10.9,-1.9) {$\times$};
\minigrid{11.7,-1.6}{stTeal!30}
\foreach \j in {0,...,4}{\filldraw[fill=stTeal!15,draw=white,line width=.3pt] ({11.7+\j*.2},-1.8) rectangle ++(.2,-.2);}
\foreach \j in {0,...,4}{\filldraw[fill=stOchre!45,draw=white,line width=.3pt] ({11.7+\j*.2},-2.0) rectangle ++(.2,-.2);}
\node[note,align=left,anchor=west] at (12.85,-1.9) {\textbf{Keep tokens and keys,}\\\textbf{anchor static values}\\full coverage; active values kept};
\node[text=stTeal!80!black,font=\bfseries\large] at (16.75,-1.9) {$\checkmark$};
\node[note,align=left,anchor=west,text=stInk!75] at (0,-2.75) {At equal coverage (48 sources), the blend adds 8.14\,dB PSNR (Tab.~\ref{tab:st_paired_completed}); a pixel composite reaches similar static PSNR but keeps moving-region drift (Tab.~\ref{tab:extra_composite_drift}).};
\end{tikzpicture}
\endgroup
 \caption{\textbf{Component 2: read-time value tethering} within the $4\times7$-token road--water crop of \cref{fig:st_comp_map}. Keys $K_r$ remain fixed as newest-frame value vectors are interpolated toward co-located source vectors $V_0(p)$ with weights $\beta(p)$. Patch colors are illustrative summaries at $58\,\text{s}$; the actual blend takes place in attention-value space, and untethered high-activity entries do not imply unchanged dynamic outputs. The schematic omits the current block, which attention also reads in full.}
 \label{fig:st_comp_tether}
\end{figure*}

\Cref{fig:st_comp_map,fig:st_comp_tether} visualize \Ours's two core operations step by step on a single development rollout.
First, \cref{fig:st_comp_map} computes tokenwise log-variance across the first six generated latents ($1.5\,\text{s}$), partitions high-activity tokens with Otsu's threshold, and freezes $\beta(p)$ for the rollout. An early lighting change marks a sky patch as high activity, illustrating that the map measures variation rather than semantic motion (the frozen-map limitation in Sec.~6 of the main paper).
Second, \cref{fig:st_comp_tether} zooms into a $4\times7$ road--water boundary patch at $58\,\text{s}$: the read keeps $K_r$, interpolates low-activity road values toward $V_0$, and leaves high-activity water values $V_r$ untethered. Patch colors summarize the operation; they are not literal value channels and do not imply that keys encode only geometry and values only appearance.

\begin{table*}[t]
 \centering\scriptsize\setlength{\tabcolsep}{4pt}
 \caption{\textbf{Single-factor sensitivity to ramp width $\eta$, history decay $\rho$, ramp continuity, and periodic map refresh at $60\,\text{s}$} (seed 42, 42 appearance and 40 activity scenes; differences report arm minus locked \Ours and flow ratios report arm over locked \Ours, with 95\% scene-bootstrap intervals).}
 \label{tab:extra_sensitivity_review}
 \begin{tabular}{@{}lrrrrrr@{}}\toprule
Arm & PSNR $\uparrow$ & $\Delta$PSNR [95\% CI] & Color $\downarrow$ & $\Delta$Color [95\% CI] & Activity & Activity ratio [95\% CI]\\\midrule
SceneTether ($\eta=0.3$, $\rho=0.3$, band 0.5, frozen map) & 22.37 & -- & 5.57 & -- & 1.97 & --\\
Taper $\eta=0.1$ & 22.62 & +0.25 [$-$0.00, +0.53] & 6.10 & +0.53 [+0.27, +0.84] & 1.94 & 0.99 [0.88, 1.11]\\
Taper $\eta=0.5$ & 21.66 & $-$0.71 [$-$1.08, $-$0.36] & 5.07 & $-$0.50 [$-$0.88, $-$0.12] & 1.88 & 0.95 [0.86, 1.06]\\
Taper $\eta=1.0$ & 19.64 & $-$2.73 [$-$3.44, $-$2.04] & 4.86 & $-$0.71 [$-$1.52, +0.14] & 1.97 & 1.00 [0.89, 1.15]\\
History budget $\rho=0.1$ & 22.14 & $-$0.23 [$-$0.39, $-$0.08] & 5.77 & +0.21 [+0.01, +0.43] & 2.02 & 1.02 [0.92, 1.14]\\
History budget $\rho=0.5$ & 22.52 & +0.15 [$-$0.06, +0.37] & 5.54 & $-$0.03 [$-$0.30, +0.29] & 1.90 & 0.96 [0.82, 1.11]\\
One continuous ramp, no band rule & 22.50 & +0.13 [$-$0.11, +0.42] & 5.58 & +0.01 [$-$0.37, +0.37] & 1.74 & 0.88 [0.81, 0.98]\\
Map re-estimated every 5\,s & 22.65 & +0.28 [$-$0.40, +0.97] & 6.26 & +0.70 [+0.03, +1.36] & 2.24 & 1.13 [0.94, 1.38]\\
\bottomrule\end{tabular}

\end{table*}

\IfFileExists{tables/tether_diag_nov06.tex}{%
\section{Local attention-output diagnostic}
\label{fm:suppl_tether_diag}
Across six benchmark scenes (reproducing scored rollouts byte-for-byte), we log the per-head pre-projection attention shift $\Delta O$ between tethered reads and otherwise identical untethered newest-frame reads (\cref{tab:st_tether_diag}).
Excluding the six untethered warm-up latents in minute~1, we compute the median across blocks, layers, and denoising steps per scene and minute before averaging across scenes, partitioning queries via the frozen activity map (characterizing single-call shifts rather than closed-form rollout bounds).
\begin{table*}[t]
 \centering\small
 \caption{\textbf{Local pre-projection attention shift across four-minute rollouts} (six test scenes, seed 42). $\Delta O$ denotes tethered minus untethered attention output relative to untethered norm $O$ and value shift $\Delta V$ (means of scene medians, plus mean and range of per-scene minute-4-to-minute-1 ratios).}
 \label{tab:st_tether_diag}
 \input{tables/tether_diag_nov06}
\end{table*}
}{}

\end{document}